\documentclass{article}

\usepackage[preprint]{neurips_2026}

\usepackage[utf8]{inputenc}
\usepackage[T1]{fontenc}
\usepackage{microtype}
\usepackage{graphicx}
\usepackage{subcaption}
\usepackage{booktabs}
\usepackage{bm}
\usepackage{array}
\usepackage{hyperref}
\usepackage{enumitem}
\usepackage{url}
\usepackage{nicefrac}
\usepackage{amsfonts}
\usepackage{amsmath}
\usepackage{amssymb}
\usepackage{mathtools}
\usepackage{amsthm}
\usepackage{booktabs}
\usepackage{multirow}
\usepackage{wrapfig}
\usepackage[table]{xcolor}
\usepackage[capitalize,noabbrev]{cleveref}

\theoremstyle{plain}

\theoremstyle{definition}

\theoremstyle{remark}

\usepackage{xcolor}
\usepackage{tcolorbox}
\usepackage{listings}
\tcbuselibrary{skins, breakable}
\usepackage[textsize=tiny]{todonotes}

\newcommand{\ie}{\emph{i.e., }}
\newcommand{\eg}{\emph{e.g., }}

\newcommand{\cf}{\emph{cf. }}

\title{On Predicting the Post-training Potential of Pre-trained LLMs}

\author{%
  Xiaoyuan Li$^{1}$\thanks{Equal contribution.}  \quad
  Yubo Ma$^{2}$\footnotemark[1]  \quad
  Kexin Yang$^{2}$ \quad
  Moxin Li$^{3}$ \quad
  Keqin Bao$^{1}$ \quad
  Wenjie Wang$^{1}$ \quad \\
  \textbf{Fuli Feng}$^{1}$ \quad
  \textbf{Dayiheng Liu}$^{2}$ \\
  $^{1}$University of Science and Technology of China \quad
  $^{2}$Alibaba Group \quad \\
  $^{3}$National University of Singapore \\
}

\begin{document}

\maketitle

\begin{abstract}
The performance of Large Language Models (LLMs) on downstream tasks is fundamentally constrained by the capabilities acquired during pre-training. However, traditional benchmarks like MMLU often fail to reflect a base model's plasticity in complex open-ended scenarios, leading to inefficient model selection. We address this by introducing a new task of predicting \textbf{\textit{post-training potential}} - forecasting a base model's performance before post-training. We propose \texttt{RuDE} (\textbf{Ru}bric-based \textbf{D}iscriminative \textbf{E}valuation), a unified framework that bypasses the generation gap of base models by leveraging response discrimination. Guided by our systematic \textbf{4C Taxonomy}, \texttt{RuDE} constructs controlled contrastive pairs across diverse domains by fine-grained rubric violations. Extensive experiments demonstrate a correlation greater than 90\% with post-training performance. Crucially, validation via Reinforcement Learning (RL) confirms that \texttt{RuDE} effectively identifies high-potential smaller models that outperform larger counterparts, offering a compute-efficient mechanism for foundation model development.
\end{abstract}







\section{Introduction}

The development of Large Language Models (LLMs) typically follows a two-stage pipeline: \emph{Pre-training} followed by \emph{Post-training} ~\citep{yang2025qwen3,team2025gemma,liu2025deepseek}. 
Pre-training over large text corpora allows the LLM to capture linguistic patterns and store extensive knowledge \cite{radford2018improving}. 
Subsequently, post-training enables the LLM to follow human instructions \cite{raffel2020exploring}, align with human preferences \cite{ouyang2022training}, and perform complex reasoning tasks \cite{shao2024deepseekmath}, unlocking the underlying capabilities captured during pre-training. 
Recently, studies show that the performance of post-trained largely depends on the capabilities acquired during pre-training stage \cite{yue2025does, ghosh2024closer, zhou2023lima}. 
Therefore, selecting an optimal pre-trained model is crucial for effective post-training and for saving computational resources. 

However, a critical research gap remains in how to pre-evaluate the potential of pre-trained models for effective post-training. 
Existing evaluations typically measure pre-trained model quality using accuracy on multiple-choice knowledge benchmarks such as MMLU~\citep{hendrycks2021measuring} or using perplexity metrics~\citep{zeng2025can}. 
These metrics, however, are insufficient to capture the model’s \emph{\textbf{post-training potential}}—its expected performance on post-training tasks that are typically open-ended and require both complex instruction following ability and understanding of human intents.
We observe that high performance of pre-trained models on multiple-choice knowledge benchmarks correlates weakly with their post-training performance on open-ended instruction following tasks (\cf Figure~\ref{fig:corr_compare}). 
This motivates us to investigate the following problem.

\begin{center}
\begin{tcolorbox}[colback=blue!5, colframe=blue!50!black, 
                  boxrule=0.8pt, arc=4mm, width=0.95\linewidth, left=2mm, right=2mm]
\textit{How can we evaluate the relative post-training potential of pre-trained models across diverse tasks?}
\end{tcolorbox}
\end{center}

We identify the following challenges in addressing this problem. 
(1) \textit{Generation Gap}: Pre-trained models are optimized for next-token prediction and generally lack the instruction-following capability to respond to open-ended queries~\citep{zhou2023lima}, making direct evaluation challenging.
(2) \textit{Unified Objective Evaluation}: Unlike multiple-choice questions, which can be evaluated via exact answer matching, open-ended responses involve diverse criteria such as formatting, helpfulness, and factual correctness. A generalized evaluation framework is therefore needed to quantify these objectives across diverse tasks~\citep{liang2023holistic}.

To address the first challenge, inspired by the \textit{Generation-Evaluation Consistency} hypothesis~\citep{liu2025on}, we estimate a pre-trained model’s ability to generate desired open-ended responses based on its capability to discriminate between preferred and dispreferred responses. 
To tackle the second challenge, we introduce the \textbf{4C Taxonomy}, providing a systematic framework for evaluating open-ended generation across four distinct aspects: \textit{Competence}, \textit{Content}, \textit{Control}, and \textit{Compliance}. Post-training tasks are evaluated using discriminative response sets constructed from task-specific combinations of these aspects, allowing the framework to adapt to diverse tasks. 

To this end, we propose \texttt{RuDE} (\textbf{Ru}bric-based \textbf{D}iscriminative \textbf{E}valuation) a unified framework for evaluating the relative post-training potential of pre-trained models across diverse tasks. 
We begin by manually defining fine-grained rubrics, \ie evaluation criteria, for each aspect of the \textbf{4C Taxonomy}. For a comprehensive evaluation, we adapt four open-ended datasets into the response discrimination format, covering diverse domains: \textit{Medical, Legal \& Finance, Complex Instruction Following, and Creative Writing}, and define rubric sets for each task. 
We propose a generator-verifier pipeline to automatically construct discriminative response sets, where preferred responses satisfy all rubrics and dispreferred responses violate some. 
Evaluation results on pre-trained models show a strong Pearson correlation ($>90\%$) with their fully instruction-tuned performance. 
Experiments on Reinforcement Learning (RL) further validate our approach, showing that a smaller pre-trained model (Qwen3-4B-Base) predicted to have higher potential indeed outperforms a larger counterpart (Qwen2.5-7B-Base) after post-training.
Our contributions are summarized as follows:
\begin{itemize}
    \item We introduce the new task of evaluating the post-training potential of pre-trained LLMs.
    \item We propose a unified evaluation framework for post-training potential across diverse tasks via rubric-based response discrimination based on our 4C Taxonomy.
    \item We perform extensive experiments on instruction tuning and RL across multiple models, demonstrating the strong effectiveness of \texttt{RuDE}.
\end{itemize}
\section{Task Formulation}
\label{sec:formulation}

In this section, we formalize the task of evaluating post-training potential. We first define the problem as a ranking task over pre-trained models (\S\ref{sec:problem_def}). We then introduce discriminative evaluation as a proxy to address the generation gap (\S\ref{sec:discrimination}). Finally, we state our core hypothesis connecting discriminative ability to predicting post-training potential (\S\ref{sec:hypothesis}).

\subsection{Problem Definition}
\label{sec:problem_def}

The performance of post-trained models is largely determined by their pre-training foundation, yet no reliable method exists to predict which pre-trained model will perform best after post-training. We formalize this as a ranking problem.

Consider a set of pre-trained base models $\Theta = \{\theta_1, \dots, \theta_n\}$ and a target post-training task $T$. Let $\texttt{PostTrain}(\theta, T)$ denote the model obtained by applying a post-training algorithm (\eg RL) to base model $\theta$ on task $T$, and let $S_{\text{gen}}(\texttt{PostTrain}(\theta, T))$ be the resulting generative performance score on open-ended evaluation.

Our goal is to find a \emph{scoring function} $S(\theta, T)$, computable on the base model alone, such that the ranking induced by $S$ over $\Theta$ is consistent with the ranking induced by $S_{\text{gen}}$ after post-training:
\begin{equation}
    S(\theta_i, T) > S(\theta_j, T) \;\Longrightarrow\; S_{\text{gen}}(\texttt{PostTrain}(\theta_i, T)) > S_{\text{gen}}(\texttt{PostTrain}(\theta_j, T))
\end{equation}
A reliable scoring function $S$ would allow practitioners to select the most promising base model without expensive trial-and-error post-training.

\begin{figure*}[t]
    \centering
    \includegraphics[width=\linewidth]{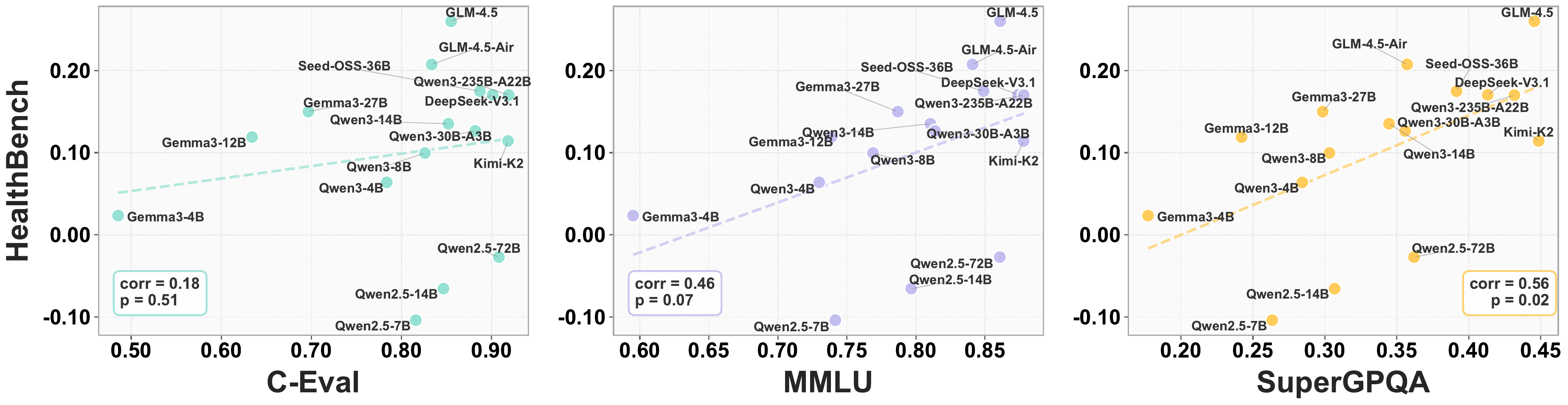}
    \caption{\textbf{Correlation Analysis: Traditional Benchmarks vs. Post-Training Performance.} 
    The X-axis represents the accuracy of base models on widely used knowledge benchmarks and the Y-axis denotes the corresponding models' post-training performance on HealthBench after instruction tuning. 
    The Pearson correlation coefficients ($corr$) between them indicate a weak correlation, demonstrating that traditional benchmarks can not predict domain-specific post-training potential.}
    \label{fig:corr_compare}
\end{figure*}

As shown in Figure~\ref{fig:corr_compare}, existing proxies for base model quality—such as accuracy on multiple-choice benchmarks like MMLU—correlate weakly with post-training performance on open-ended tasks, because they measure knowledge recall rather than the model's capacity to produce aligned, high-quality responses. This motivates the search for a more predictive scoring function.

\subsection{Discriminative Evaluation as a Proxy}
\label{sec:discrimination}

The first challenge identified in the introduction is the \emph{generation gap}: pre-trained models, optimized for next-token prediction, generally cannot follow instructions or produce well-structured open-ended responses~\citep{zhou2023lima}. This makes it impractical to directly evaluate their generative quality on post-training tasks.

To address this, inspired by the \textit{Generation-Evaluation Consistency} hypothesis~\citep{liu2025on}, we shift from evaluating what a model can \emph{generate} to evaluating what it can \emph{discriminate}. The key insight is that a model's ability to distinguish preferred responses from dispreferred ones reflects its internalized understanding of quality—even when it cannot yet produce such responses itself. From the perspective of Direct Preference Optimization (DPO)~\citep{rafailov2023direct}, the base model serves as the initial reference policy, and post-training adjusts its distribution toward preferred responses. A base model that already assigns higher likelihood to preferred responses faces a smaller distributional shift during optimization, implying higher post-training potential.

We operationalize this by measuring whether the base model consistently ranks a preferred response $y^+$ above a dispreferred response $y^-$ in terms of conditional log-likelihood. Let $\ell_\theta(y|x)$ denote the log-likelihood assigned by model $\theta$ to response $y$ given prompt $x$. Given a set of contrastive pairs $\{(x, y^+, y^-)\}$ constructed for task $T$, we define the discriminative score:
\begin{equation}
    S_{\text{dis}}(\theta, T) = \mathbb{E}_{(x, y^+, y^-)} \left[ \mathbb{I}\left( \ell_\theta(y^+ | x) > \ell_\theta(y^- | x) \right) \right]
\end{equation}
where $\mathbb{I}$ is the indicator function. $S_{\text{dis}}$ measures the fraction of contrastive pairs on which the base model correctly ranks the preferred response higher, reflecting how well it has already internalized an understanding of response quality prior to post-training.

\subsection{The GD-Potential Hypothesis}
\label{sec:hypothesis}
Building on the above formulation, we propose the \textbf{GD-Potential Hypothesis} (\textbf{G}enerative-\textbf{D}iscriminative Potential): \emph{the discriminative score of a pre-trained model is strongly positively correlated with its post-training generative performance.} Formally:
\begin{equation}
    \rho\!\left(S_{\text{dis}}(\theta, T),\; S_{\text{gen}}(\texttt{PostTrain}(\theta, T))\right) \gg 0
\end{equation}
where $\rho$ denotes the Pearson correlation coefficient computed across a population of base models.

If this hypothesis holds, $S_{\text{dis}}$ serves as the desired scoring function $S$ defined in \S\ref{sec:problem_def}, enabling practitioners to predict post-training potential without actually performing post-training. 

\section{RuDE Framework}
\label{sec:method}

\begin{figure*}[t]
    \centering
    \includegraphics[width=0.75\linewidth]{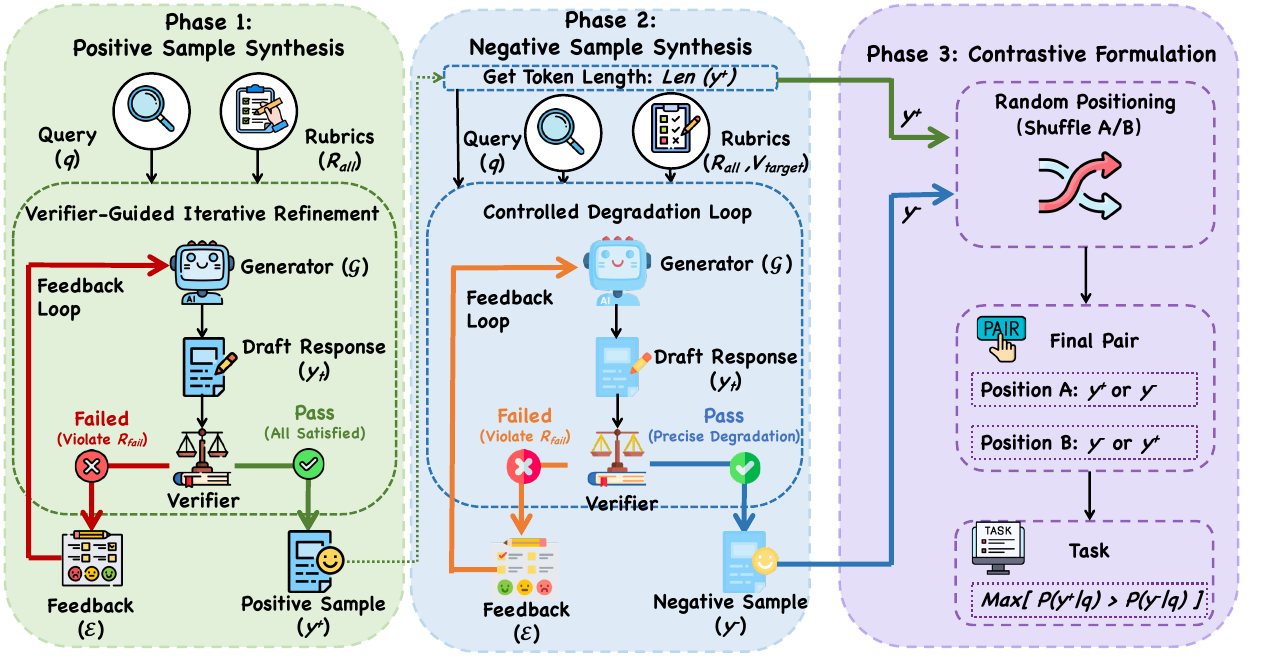}
    \caption{\textbf{Overview of the RuDE Construction Pipeline.} The framework operates in three phases to synthesize high-quality contrastive pairs. Phase 1 generates a gold-standard response ($y^+$), Phase 2 produces a hard negative ($y^-$) through a controlled degradation process. Finally, Phase 3 pairs these samples and randomizes their positions (A/B) to form the discriminative evaluation task.}
    \label{fig:overview}
    \vspace{-10pt}
\end{figure*}

To empirically validate the \textit{GD-Potential} hypothesis, we introduce \texttt{RuDE} (\textbf{Ru}bric-based \textbf{D}iscriminative \textbf{E}valuation). It is constructed via a rigorous, verifier-guided iterative pipeline to ensure that every discriminative task is grounded in fine-grained alignment rubrics.

\subsection{Iterative Rubric-Perturbed Construction}
\label{sec:pipeline}
The core of our methodology is the synthesis of contrastive pairs $(y^+, y^-)$ that differ only in specific, controlled alignment dimensions. To achieve this precision and eliminate confounding factors, we implement a pipeline involving a strong generator ($\mathcal{G}$, Gemini-3-Pro~\citep{comanici2025gemini}) and a strict verifier ($\mathcal{V}$, GPT-4.1~\citep{singh2025openai})\footnote{Following the evaluation protocol established in HealthBench~\citep{arora2025healthbench}, we employ GPT-4.1 as the verifier to assess rubric compliance. HealthBench has validated its high correlation with human expert judgment in rubric-based assessment.}, which proceeds in three phases as illustrated in Figure ~\ref{fig:overview}:

\paragraph{Phase 1: Positive Sample Synthesis via Verifier-Guided Iterative Refinement.}
The objective is to generate a response $y^+$ that perfectly satisfies all rubrics $\mathcal{R}_{all}$.
Given a query $q$ and $\mathcal{R}_{all}$, $\mathcal{G}$ generates a response $y_t$ and then $\mathcal{V}$ evaluates $y_t$ against every rubric in $\mathcal{R}_{all}$.\footnote{$\mathcal{R}_{all}$ comprises two categories: \textit{Positive Rubrics} (features that must be present) and \textit{Negative Rubrics} (features that must be avoided). In our pipeline, ``perfectly satisfies'' is defined as the state where $\mathcal{V}$ predicts \texttt{True} for all Positive Rubrics and \texttt{False} for all Negative Rubrics.} 
\begin{itemize}
    \item \textit{Feedback Loop:} If any rubric $r \in \mathcal{R}_{fail}$ is violated, $\mathcal{V}$ generates structured feedback $\mathcal{E}$ explaining the reason for failure. The tuple $(q, \mathcal{R}_{all}, \mathcal{R}_{fail}, \mathcal{E}, y_t)$ is fed back to $\mathcal{G}$ for correction.
    \item \textit{Termination:} This loop continues iteratively until $\mathcal{V}$ confirms that all rubrics $\mathcal{R}_{all}$ are satisfied, yielding the positive sample $y^+$.
\end{itemize}

\paragraph{Phase 2: Negative Sample Synthesis via Controlled Degradation Loop.}
The goal is to synthesize a ``Hard Negative'' $y^-$ that violates a specific subset of rubrics while maintaining high quality in all other aspects.
We define a target violation subset $\mathcal{V}_{target} \subset \mathcal{R}_{all}$ and capture the token length of the positive sample, $Len(y^+)$.
\begin{itemize}
    \item \textit{Length Constraint:} To mitigate length bias, we explicitly constrain $\mathcal{G}$ to generate $y^-$ such that its token count mirrors the positive sample: $Len(y^-) \approx Len(y^+)$.
    
    \item \textit{Dual-Verification:} $\mathcal{V}$ performs a rigorous dual-check on the candidate $y_t$: 
    (1) Are the rubrics in $\mathcal{V}_{target}$ explicitly violated? 
    (2) Are all remaining rubrics $\mathcal{R}_{preserved} = \mathcal{R}_{all} \setminus \mathcal{V}_{target}$ strictly satisfied?
    
    \item \textit{Refinement Loop:} If $y_t$ fails either condition, $\mathcal{V}$ generates a structured error analysis $\mathcal{E}$ for any rubric $r \in \mathcal{R}_{fail}$(e.g., detailing a failure to violate a target or an accidental violation of a preserved constraint). The optimization tuple $(q, \mathcal{R}_{all}, \mathcal{V}_{target}, \mathcal{R}_{fail},\mathcal{E}, y_t)$ is fed back to $\mathcal{G}$. This ensures the generator iteratively optimizes the output until it precisely matches the specific degradation requirements.
\end{itemize}
\begin{table*}[t]
    \centering
    \caption{\textbf{The 4C Taxonomy}: A fine-grained breakdown of model post-training potential.}
    \label{tab:4c_taxonomy}
    \renewcommand{\arraystretch}{1} 
    \scalebox{0.8}{
    \begin{tabular}{l l p{12cm}}
        \toprule
        \textbf{Dimension} & \textbf{Sub-Category} & \textbf{Description} \\
        \midrule
        
        \multirow{3}{*}{\textbf{Competence}} 
        & \textbf{Factuality} & Ensures accuracy of established facts and absence of hallucinations. \\
        & \textbf{Logic} & Validates the correct application of reasoning rules and causal relationships. \\
        & \textbf{Procedure} & Measures adherence to standard procedures and sequential step-by-step instructions. \\
        \midrule
        
        \multirow{3}{*}{\textbf{Content}} 
        & \textbf{Completeness} & Checks if the response addresses all distinct components of the user's query. \\
        & \textbf{Coherence} & Evaluates the linguistic flow, readability, and grammatical correctness. \\
        & \textbf{Relevance} & Ensures the response stays strictly on topic without unrelated tangents. \\
        \midrule
        
        \multirow{3}{*}{\textbf{Control}} 
        & \textbf{Format} & Enforces specific visual structures like JSON or Markdown tables. \\
        & \textbf{Length} & Validates the response meets specified length requirements in terms of words. \\
        & \textbf{Scope} & Checks for adherence to mandatory inclusion keywords or exclusion constraints. \\
        \midrule
        
        \multirow{3}{*}{\textbf{Compliance}} 
        & \textbf{Safety} & Measures ethical boundaries, refusal of harmful requests, and disclaimers. \\
        & \textbf{Persona} & Evaluates the ability to adopt a specific character role, tone, or interaction style. \\
        & \textbf{Utility} & Assesses the practical helpfulness of the response in satisfying user intent. \\
        
        \bottomrule
    \end{tabular}
    }
\end{table*}
\paragraph{Phase 3: Contrastive Formulation.}
Finally, $y^+$ and $y^-$ are paired to form a binary discriminative task. To eliminate position bias during evaluation~\citep{zheng2023judging}, the positive sample is randomly assigned to position A or B. The task for the base model is to identify the superior response based on conditional probability: $P(y^+|q) > P(y^-|q)$.

\subsection{The 4C Taxonomy: Decoupling Capabilities}
To enable fine-grained diagnosis of model potential~\citep{liang2023holistic,ye2024flask}, we categorize all rubrics into the \textbf{4C Taxonomy}. It disentangles the complex spectrum of post-training potential into four distinct capabilities:
\textbf{(1) Competence} assesses the model's intrinsic intelligence and reliability, which reflects knowledge density acquired during pre-training~\citep{hendrycks2021measuring,cobbe2021training}.
\textbf{(2) Content} evaluates the linguistic quality of the generation, ensuring it meets the standards of human communication~\citep{brown2020language}.
\textbf{(3) Control} measures the model's ability to adhere to rigid, non-semantic constraints and reflects the model's instruction following plasticity~\citep{zhou2023instruction}.
\textbf{(4) Compliance} evaluates the model's alignment with human intent, safety boundaries, and social norms. High scores here indicate a model that is ``safe and helpful''~\citep{ouyang2022training,bai2022training,zhang2024safetybench}.
See Table ~\ref{tab:4c_taxonomy} for details.

\section{Experiments}
\label{sec:experiments}
\label{ssec:setup}

\subsection{Dataset}
\paragraph{Data Sources.}
To ensure the comprehension, we curate data from four diverse domains: \textbf{(1) HealthBench~\citep{arora2025healthbench}}: Centered on medical advisory tasks, it evaluates adherence to safety-critical constraints.\footnote{We specifically select the `Hard' subset of HealthBench for construction. Statistics reveals that the `Consensus' subset averages only 2.19 rubrics per entry, which is insufficient for hard negative sampling. In contrast, the `Hard' subset provides a dense constraint landscape with an average of 11.85 rubrics per entry, enabling the synthesis of fine-grained contrastive pairs.}
\textbf{(2) PRBench~\citep{akyurek2025prbench}}: Derived from legal and financial professional reports, it probes logical consistency and terminology adherence across extensive documents.
\textbf{(3) AdvancedIF~\citep{he2025advancedif}}: Designed to stress-test complex instruction-following, this dataset features nested formatting and logic constraints.\footnote{We exclude the `System Prompt Modification' subset of AdvancedIF. Since pre-trained models typically lack a  system role, this subset is not suited for our framework.}
\textbf{(4) WritingBench~\citep{wu2025writingbench}}: Targeting open-ended storytelling, it assesses style transfer, rhetorical usage, and persona consistency.    

\paragraph{Statistics.}
\begin{wraptable}{r}{0.52\linewidth}
    \vspace{-12pt}
    \centering
    \caption{\textbf{Statistics of the evaluation datasets.} It covers four complex open-ended domains with varying difficulty levels controlled by the number of rubric violations ($|\mathcal{V}|$).}
    \label{tab:statistics}
    \scriptsize
    \setlength{\tabcolsep}{2.6pt}
    \renewcommand{\arraystretch}{0.94}
    \resizebox{\linewidth}{!}{
    \begin{tabular}{llccr}
        \toprule
        \textbf{Domain} & \textbf{Dataset} & \textbf{$|\mathcal{V}|$} & \textbf{Samples} & \textbf{Avg. Tok.} \\
        \midrule
        \multirow{3}{*}{Medical Interaction} & \multirow{3}{*}{HealthBench} & 1 & 1,597 & 3,968 \\
         & & 3 & 1,031 & 3,480 \\
         & & 5 & 1,045 & 3,912 \\
        \midrule
        \multirow{3}{*}{Legal \& Finance} & \multirow{3}{*}{PRBench} & 1 & 3,848 & 20,194 \\
         & & 3 & 4,043 & 10,776 \\
         & & 5 & 4,542 & 11,724 \\
        \midrule
        \multirow{3}{*}{Complex Instruction} & \multirow{3}{*}{AdvancedIF} & 1 & 1,930 & 6,163 \\
         & & 3 & 1,349 & 8,111 \\
         & & 5 & 1,559 & 4,833 \\
        \midrule
        \multirow{3}{*}{Creative Writing} & \multirow{3}{*}{WritingBench} & 1 & 2,561 & 14,439 \\
         & & 2 & 2,523 & 16,222 \\
         & & 3 & 2,655 & 23,913 \\
        \midrule
        \textbf{Total / Avg.} & & - & \textbf{28,683} & \textbf{11,566} \\
        \bottomrule
    \end{tabular}
    }
    \vspace{-8pt}
\end{wraptable}
The final evaluation dataset comprises 28,683 samples. We stratify difficulty by varying the number of rubric violations ($|\mathcal{V}| \in \{1, 2, 3\}$ for WritingBench, and $|\mathcal{V}| \in \{1, 3, 5\}$ for the others).\footnotemark Specifically, $|\mathcal{V}|=1$ represents the hardest scenario, requiring precise discrimination, while the maximum violation setting ($|\mathcal{V}|=3$ or $5$) represents a relatively easier task.
\footnotetext{Since the WritingBench dataset contains a lower density of constraints (5 rubrics per entry), generating negative samples with 5 violations would result in easy negative responses. Thus, we adjusted the violation number to $|\mathcal{V}| \in \{1, 2, 3\}$ for this domain.}

\paragraph{Human Verification.} To validate the reliability of our  pipeline, we randomly sample 100 instances for human review. We verify whether the positive sample satisfies all rubrics and whether the negative sample violates exactly the target rubrics. The result shows a 92\% agreement between the pipeline's labels and human judgment, confirming that \texttt{RuDE} is grounded in high-quality human preferences. \footnote{See details in Appendix~\ref{ssec:human_verification}.} 

\subsection{Experimental Setup}
\paragraph{Models Evaluated.}
We evaluate a diverse suite of state-of-the-art base LLMs, covering a wide range of parameter scales (4B to 1T) and architectures (Dense and MoE). The models evaluated include DeepSeek-V3.1~\citep{liu2025deepseek}, GLM-4.5/Air~\citep{zeng2025glm}, Kimi-K2~\citep{team2025kimi}, Seed-OSS-36B~\citep{seed2025seed1}, Qwen2.5 Series, Qwen3 Series~\citep{yang2025qwen3}, and Gemma3 Series~\citep{team2025gemma}. Notely, unless otherwise specified, the models referred to in this work are base models.

\paragraph{Evaluation Metrics.}
We primarily report accuracy ($Acc$), defined as the percentage of contrastive pairs where the model correctly assigns higher probability to the positive sample $y^+$ than the negative one $y^-$. To validate the \textit{GD-Potential} hypothesis, we also calculate the Pearson correlation ($corr$ and $p$)~\citep{benesty2009pearson} between the base models' accuracy on \texttt{RuDE} and their instruct-tuned counterparts' performance on standard generative benchmarks.

\paragraph{Implementation Details.}
We adopt a 2-shot setting with balanced exemplars (one targeting option `A' and the other `B') to mitigate potential in-context position bias. For all experiments, we uniformly set the maximum output length to 8192 and the temperature to 0 to ensure determinacy.

\begin{figure*}[t]
    \centering
    \includegraphics[width=0.95\linewidth]{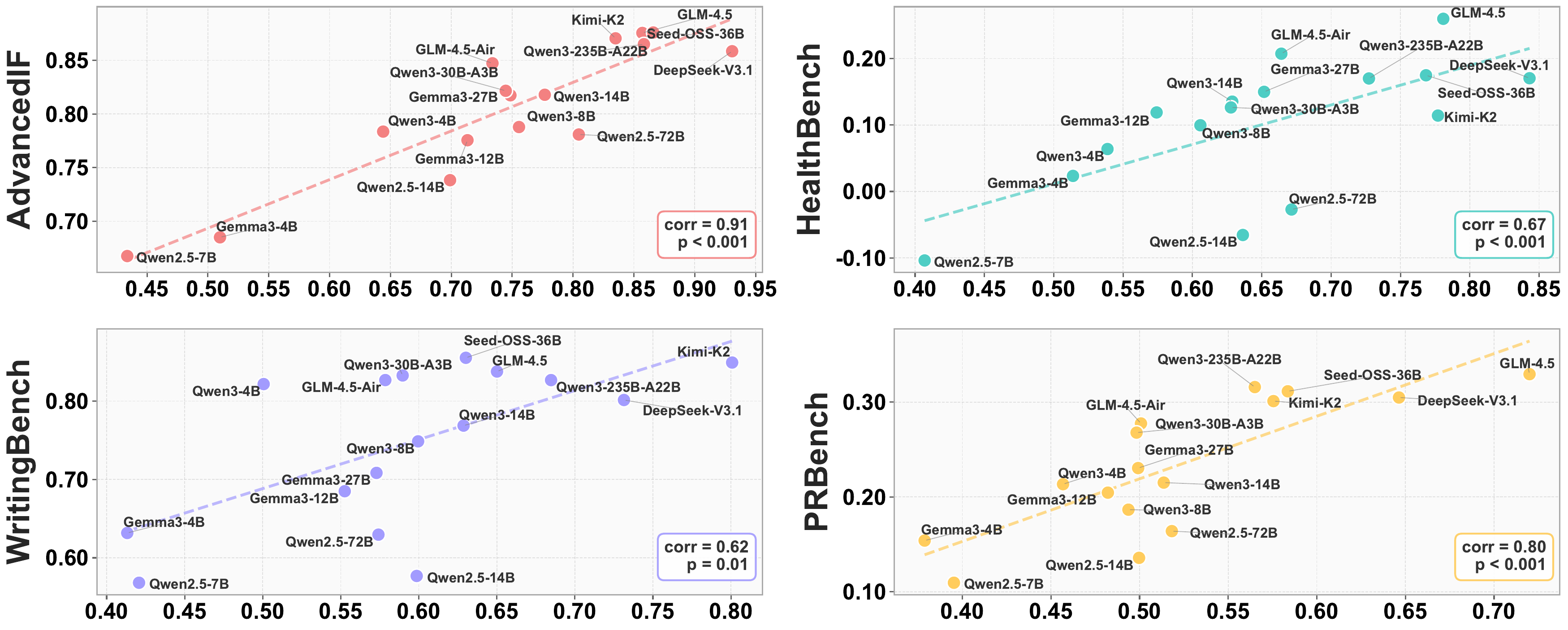}
    \caption{\textbf{Correlation Analysis: Pre-training vs. Post-training Performance.} 
    We illustrate the relationship between base model scores on the \texttt{RuDE} (X-axis) and the corresponding instruction-tuned models' performance on standard open-ended benchmarks (Y-axis). 
    The high Pearson correlation coefficients ($r > 0.90$) demonstrate that our discriminative evaluation serves as a highly reliable predictor of downstream post-traning potential, significantly outperforming traditional benchmarks like MMLU.}
    \vspace{-10pt}
    \label{fig:correlation_analysis_mcq5}
\end{figure*}

\subsection{Main Results}
\label{ssec:main_results}
\begin{figure}[t]
    \centering
    \begin{minipage}[t]{0.49\linewidth}
        \vspace{0pt}
        \centering
        \captionof{table}{\textbf{Model Performance on \texttt{RuDE}.} We report the accuracy (\%) of $|\mathcal{V}|=3$ for WritingBench and $|\mathcal{V}|=5$ for others. The column headers denote: \textbf{AD} (\textit{AdvancedIF}), \textbf{HE} (\textit{HealthBench}), \textbf{WR} (\textit{WritingBench}), and \textbf{PR} (\textit{PRBench}). \textbf{Bold} indicates the best.}
        \label{tab:main_results}
        \tiny
        \renewcommand{\arraystretch}{0.88}
        \setlength{\tabcolsep}{1.8pt}
        \resizebox{\linewidth}{!}{
        \begin{tabular}{lcccc|c}
            \toprule
            \textbf{Model} & \textbf{AD} & \textbf{HE} & \textbf{WR} & \textbf{PR} & \textbf{AVG} \\
            \midrule
            Gemma3-4B       & 51.0 & 51.4 & 41.3 & 37.9 & 45.4 \\
            Gemma3-12B      & 71.3 & 57.4 & 55.3 & 48.2 & 58.1 \\
            Gemma3-27B      & 74.9 & 65.2 & 57.3 & 49.9 & 61.8 \\
            Qwen3-4B        & 64.4 & 53.9 & 50.1 & 45.7 & 53.5 \\
            Qwen3-8B        & 75.6 & 60.6 & 60.0 & 49.4 & 61.4 \\
            Qwen3-14B       & 77.7 & 62.9 & 62.9 & 51.4 & 63.7 \\
            Qwen3-30B-A3B   & 74.5 & 62.8 & 59.0 & 49.8 & 61.5 \\
            Qwen3-235B-A22B & 85.8 & 72.7 & 68.5 & 56.5 & 70.9 \\
            Qwen2.5-7B      & 43.4 & 40.7 & 42.1 & 39.5 & 41.4 \\
            Qwen2.5-14B     & 69.9 & 63.6 & 59.9 & 50.0 & 60.9 \\
            Qwen2.5-72B     & 80.5 & 67.1 & 57.4 & 51.8 & 64.2 \\
            Seed-OSS-36B    & 86.6 & 76.8 & 63.0 & 58.4 & 71.2 \\
            GLM-4.5-Air     & 73.4 & 66.4 & 57.9 & 50.1 & 62.0 \\
            GLM-4.5         & 85.7 & 78.1 & 65.0 & \textbf{72.0} & 75.2 \\
            Kimi-K2         & 83.5 & 77.7 & \textbf{80.1} & 57.6 & 74.7 \\
            DeepSeek-V3.1   & \textbf{93.1} & \textbf{84.3} & 73.1 & 64.6 & \textbf{78.8} \\
            \midrule
            \textbf{AVG}    & 74.8 & 65.2 & 59.7 & 52.2 & 63.0 \\
            \bottomrule
        \end{tabular}
        }
    \end{minipage}\hfill
    \begin{minipage}[t]{0.49\linewidth}
        \vspace{0pt}
        \centering
        \includegraphics[width=0.95\linewidth]{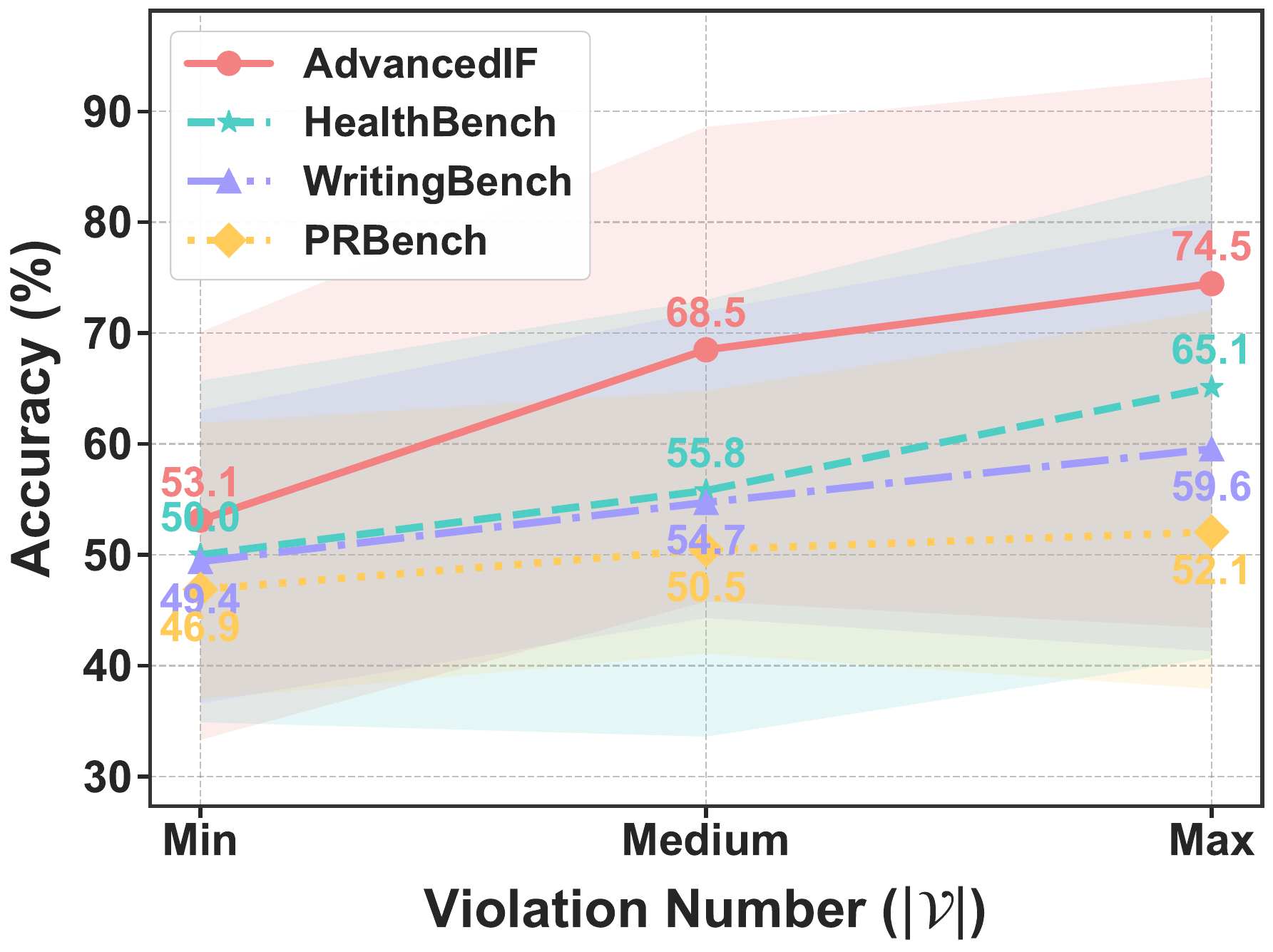}
        \caption{\textbf{Sensitivity Analysis: Impact of Violation Number.}
        We analyze how model discriminative accuracy varies as the distinction between positive and negative samples becomes more pronounced. The X-axis represents the number of rubric violations in the negative sample, ranging from Min ($|\mathcal{V}|=1$) to Max ($|\mathcal{V}|=3$ or $5$). As the number of violated rubrics increases, the discriminative task becomes easier, resulting in higher accuracy.}
        \label{fig:diff}
    \end{minipage}
\end{figure}

\paragraph{Validation of GD-Potential Hypothesis.}
In Figure~\ref{fig:correlation_analysis_mcq5}, we observe a statistically significant positive Pearson correlation between the base models' discriminative scores on \texttt{RuDE} and the generative scores of their instruct-tuned versions on the corresponding standard benchmarks. For instance, on the AdvancedIF, the correlation coefficient reaches as high as $corr=0.91$ ($p < 0.001$). High-performing models like DeepSeek-V3.1 and GLM-4.5 consistently occupy the top-right quadrant. These findings empirically validate the \textit{GD-Potential} hypothesis: \textit{A base model's discriminative capability is a reliable proxy for its post-training generative potential.} \texttt{RuDE} thus serves as a cost-effective predictor for selecting base models for expensive alignment training, allowing researchers to estimate downstream performance without conducting full post-training.

\paragraph{Overall Performance and Scaling Laws.}
As Table~\ref{tab:main_results} shows, DeepSeek-V3.1 leads with an average accuracy of 78.8\%, demonstrating superior post-training potential among all models. We observe a consistent scaling law where larger models exhibit stronger discriminative capabilities~\citep{kaplan2020scaling,wei2022emergent}. For instance, within the Gemma3 series, performance improves monotonically from 4B to 27B. This trend confirms that \textit{GD-Potential} scales with model parameters, suggesting that scaling effectively enhances a base model's latent representation of human preferences even prior to instruction tuning.

\paragraph{Domain-Specific Specialization.}
While DeepSeek-V3.1 dominates in the AdvancedIF and HealthBench, other models show distinct strengths in specific domains: On the WritingBench, Kimi-K2 achieves the best performance, surpassing DeepSeek-V3.1. This indicates Kimi-K2's superior potential in handling creative writing constraints. Moreover, GLM-4.5 demonstrates exceptional capability in legal and financial scenarios on the PRBench.
This suggests that while general capabilities scale with size, specific post-training potentials (e.g., creative writing and professional adherence) may depend on the model's pre-training data distribution and architectural biases~\citep{ye2024flask}.

\paragraph{Impact of Violation Number.}
Figure~\ref{fig:diff} validates the efficacy of our difficulty stratification mechanism by changing $|\mathcal{V}|$. As the number of injected violations decreases, the negative samples become increasingly indistinguishable from the positive ones, thereby elevating the discriminative difficulty. Consequently, we observe a sharp performance decline across all models; for instance, on the AdvancedIF, DeepSeek-V3.1's accuracy drops significantly from 93.1\% ($|\mathcal{V}|=5$) to 70.1\% ($|\mathcal{V}|=1$). This sensitivity confirms that \texttt{RuDE} avoids saturation and effectively probes capabilities at varying levels of granularity.



\section{Analysis}
\subsection{Fine-Grained Diagnostics}
\begin{wraptable}{r}{0.48\linewidth}
    \vspace{-12pt}
    \centering
    \caption{\textbf{Dimension-level performance under the 4C Taxonomy.} We report accuracy (\%) for Compliance, Competence, Content, and Control, together with the average score and the standard deviation across the four dimensions ($\sigma_{\text{4C}}$).}
    \label{tab:4c_dimension_results}
    \scriptsize
    \setlength{\tabcolsep}{2.4pt}
    \renewcommand{\arraystretch}{0.94}
    \resizebox{\linewidth}{!}{
    \begin{tabular}{lcccccc}
        \toprule
        \textbf{Model} & \textbf{Comp.} & \textbf{Compet.} & \textbf{Cont.} & \textbf{Ctrl.} & \textbf{AVG} & $\boldsymbol{\sigma_{\text{4C}}}$ \\
        \midrule
        Gemma3-4B & 43.5 & 42.5 & 38.8 & 42.0 & 41.7 & 2.0 \\
        Gemma3-12B & 51.4 & 49.8 & 47.1 & 46.4 & 48.7 & 2.4 \\
        Gemma3-27B & 52.2 & 47.9 & 43.1 & 46.2 & 47.3 & 3.8 \\
        Qwen3-4B & 43.1 & 42.5 & 47.0 & 39.3 & 43.0 & 3.1 \\
        Qwen3-8B & 50.2 & 46.8 & 45.2 & 48.1 & 47.6 & 2.1 \\
        Qwen3-14B & 52.6 & 51.2 & 55.7 & 49.3 & 52.2 & 2.7 \\
        Qwen3-30B-A3B & 49.5 & 50.1 & 50.8 & 52.0 & 50.6 & \textbf{1.1} \\
        Qwen3-235B-A22B & 56.8 & 51.6 & 59.0 & 56.3 & 55.9 & 3.1 \\
        Qwen2.5-7B & 39.6 & 35.2 & 37.4 & 34.7 & 36.7 & 2.2 \\
        Qwen2.5-14B & 52.0 & 50.6 & 49.1 & 50.4 & 50.5 & 1.2 \\
        Qwen2.5-72B & 52.9 & 50.8 & 46.8 & 52.0 & 50.6 & 2.7 \\
        Seed-OSS-36B & 57.4 & 54.1 & 54.1 & 52.0 & 54.4 & 2.2 \\
        GLM-4.5-Air & 52.2 & 49.1 & 44.3 & 52.6 & 49.6 & 3.8 \\
        GLM-4.5 & 58.3 & \textbf{61.0} & 56.9 & 57.4 & 58.4 & 1.9 \\
        Kimi-K2 & 58.1 & 56.0 & 50.9 & 57.4 & 55.6 & 3.2 \\
        DeepSeek-V3.1 & \textbf{63.0} & 59.5 & \textbf{62.4} & \textbf{60.8} & \textbf{61.4} & 1.5 \\
        \midrule
        \textbf{AVG} & 52.0 & 49.9 & 49.3 & 49.8 & 50.3 & -- \\
        \bottomrule
    \end{tabular}
    }
    \vspace{-8pt}
\end{wraptable}

\textbf{4C Dimension Profiling.}
The 4C Taxonomy provides per-dimension profiling beyond a single aggregate score. We report dimension-level accuracy in Table~\ref{tab:4c_dimension_results}, and summarize three observations.

\textbf{Global Patterns.}
Compliance is the strongest dimension for most models, averaging 52.0\% globally versus Competence (49.9\%), Control (49.8\%), and Content (49.3\%). DeepSeek-V3.1, the top model overall (AVG 61.4\%), also maintains the smallest imbalance among top-tier models ($\sigma_{\text{4C}}=1.5$). By contrast, GLM-4.5-Air shows $\sigma_{\text{4C}}=3.8$ with an 8.3 gap between Control (52.6) and Content (44.3), and Gemma3-27B shows $\sigma_{\text{4C}}=3.8$ with Compliance (52.2) exceeding Content (43.1) by 9.1.

\textbf{Control as Bottleneck for Small Models.}
For sub-7B models, Control is consistently the weakest dimension: Qwen3-4B scores 39.3\% and Qwen2.5-7B scores 34.7\%, both ranking Control last. Qwen3-4B outperforms the larger Qwen2.5-7B on every dimension, indicating that generational improvements in pre-training can raise alignment readiness beyond parameter scaling alone.

\textbf{Domain-Specific Specialization.}
On WritingBench, Kimi-K2 demonstrates comprehensive mastery across all dimensions (65.0, 60.5, 61.9, 63.3), while GLM-4.5 dominates on PRBench (60.8, 61.8, 60.8, 56.4). Qwen3-235B-A22B shows a 28.1 gap between Content (74.6) and Competence (46.5), indicating fluency without factual grounding in professional domains.

\subsection{Practical Verification via Controlled RL Training}
\label{sec:rl_verification}
To move beyond static evaluation and validate the predictive power of our framework in a dynamic setting, we conduct a controlled RL experiment. Our hypothesis is that base models with higher  discriminative scores should demonstrate superior data efficiency and peak performance when fine-tuned with rubric-derived signals.

We select three representative models, Qwen2.5-7B, Qwen2.5-14B, and Qwen3-4B, and perform rubric-based RL training following ~\citep{zhou2025breaking} on HealthBench. Figure~\ref{fig:rl} illustrates the performance trajectory on HealthBench over 200 training steps. The results provide strong empirical validation of our pre-training assessment:

\begin{wrapfigure}{r}{0.5\linewidth}
    \centering
    \includegraphics[width=\linewidth]{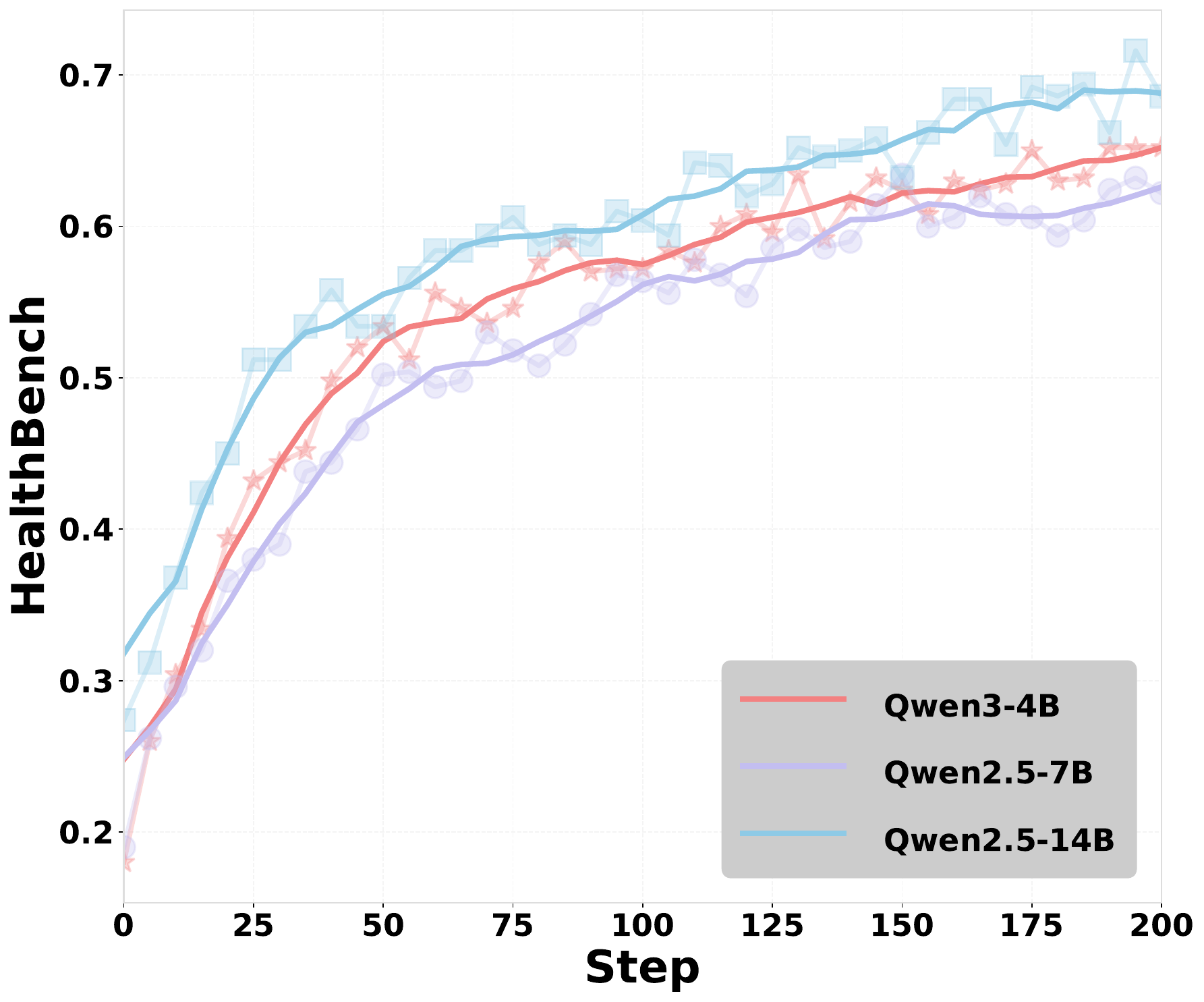}
    \caption{\textbf{Verification of Predictive Utility for Post-Training.} We compare RL trajectories for Qwen2.5-7B, Qwen2.5-14B, and Qwen3-4B.}
    \label{fig:rl}
    \vspace{-20pt}
\end{wrapfigure}

\textbf{Consistency of Ranking.} The post-training performance hierarchy broadly follows our base model evaluation. Qwen2.5-14B achieves the highest final HealthBench score, while the smaller Qwen3-4B still outperforms the larger Qwen2.5-7B, consistent with their discriminative scores on RuDE.

\textbf{Efficiency.} Qwen3-4B exhibits a stronger learning curve than Qwen2.5-7B despite its smaller size, effectively leveraging the rubric reward signals to correct its behavior rapidly.

\textbf{Predictive Validity.} This experiment confirms that our RuDE score is a faithful predictor of post-training. A high score in our framework implies that the model's internal representations are already well-structured to grasp complex constraints~\citep{zhou2023lima}, minimizing the ``alignment tax'' during post-training~\citep{ouyang2022training}.

\section{Related Work}
\label{sec:related_work}
\vspace{-10pt}
\paragraph{Open-ended Generation Evaluation.}
The evaluation for open-ended tasks is transitioning from subjective, single-score win rates (e.g., AlpacaEval~\citep{li2023alpacaeval}, MT-Bench~\citep{zheng2023judging}) toward verifiable, rubric-based assessments. Recent initiatives, such as HealthBench~\citep{arora2025healthbench} and PRBench~\citep{akyurek2025prbench}, exemplify this shift by employing rigorous, rubric-based evaluation within specific domains. We not only align with this fine-grained paradigm but also advance it by introducing a generalized rubric taxonomy - \textit{4C taxonomy} - to systematically categorize alignment requirements. This structured approach enables us to decouple entangled capabilities across diverse domains, moving beyond isolated constraint verification to a comprehensive diagnosis.

\vspace{-12pt}
\paragraph{Base Model Evaluation.}
Existing evaluations for base models predominantly focus on static knowledge benchmarks (e.g., MMLU~\citep{hendrycks2021measuring}, C-Eval~\citep{huang2023c}, SuperGPQA~\citep{du2025supergpqa}) or perplexity metrics~\citep{zeng2025can}. However, these metrics overlook a critical blind spot: the mismatch between possessing static knowledge and the dynamic requirements of post-training alignment. Our work addresses this gap by explicitly defining the task of predicting post-training potential in pre-training stage.

\vspace{-12pt}
\paragraph{Generation-Evaluation Consistency.}
Research~\citep{liu2025on} reveals a strong correlation between a model's evaluation and generation capabilities. While recent works like APT-Bench~\citep{qin2025aptbench} have begun to explore the  agentic abilities of pre-trained models, they remain focused on specific task domains. We pivot this consistency insight to a broader scope, hypothesizing that a base model's innate discriminative precision—its ability to distinguish rubric-compliant responses from violations—serves as a generalizable and cost-effective proxy for predicting its post-training potential across diverse open-ended scenarios.

\vspace{-12pt}
\paragraph{Rubric-Based Alignment and Evaluation.}
Recent work has adopted rubrics for LLM alignment: RLCF~\citep{viswanathan2025rlcf} and QA-LIGN~\citep{sun2025qalign} use rubric-derived feedback to generate preference pairs or reward signals for DPO/GRPO training, while InFoBench~\citep{qin2025infobench} and RaR~\citep{zhu2025rubrics} apply rubrics for fine-grained evaluation and RL reward modeling respectively. These works leverage rubrics to improve the post-training process itself. In contrast, RuDE applies rubric-based discrimination to base models before any post-training, addressing the complementary problem of predicting post-training potential rather than performing alignment.
\section{Conclusion}
\vspace{-10pt}
We formally defined the task of predicting \textit{post-training potential} to forecast open-ended performance prior to post-training. We proposed \texttt{RuDE}, a discriminative framework guided by our \textit{4C Taxonomy}. Experiments demonstrate a $>90\%$ correlation with post-training outcomes. Furthermore, RL validation confirms \texttt{RuDE} successfully identifies high-potential smaller models, offering a compute-efficient mechanism for foundation model selection.
\bibliography{custom}
\bibliographystyle{plainnat}

\appendix

\section{Ablation Study: Necessity of Controlled Degradation}
\label{sec:ablation_construction}

To justify the design of the \textit{Controlled Degradation} pipeline in \textsc{RuDE}, we compare it against two intuitive but ultimately inadequate alternatives for constructing contrastive pairs. The results demonstrate that both extremes—natural sampling (too hard) and naive rewriting (too easy)—fail to produce a benchmark with meaningful discriminative power, thereby validating the necessity of our approach.

\subsection{Ablation A: Rejection Sampling from Natural Generations}

We first test a straightforward baseline: sample 32 responses from a strong model given the original prompt, score each response against the full rubric set, and select the highest- and lowest-scoring responses as the positive and negative options, respectively. To control for generator-specific distributional bias, we conduct this experiment using two distinct generators—GPT-5 and Qwen3-235B-A30B-Thinking-2507—on the HealthBench domain.

\begin{table}[h]
    \centering
    \caption{\textbf{Ablation A: Rejection Sampling Construction.} Accuracy (\%) of discriminating between the best and worst naturally sampled responses. All evaluators struggle, indicating the task is excessively difficult.}
    \label{tab:ablation_rejection}
    \small
    \begin{tabular}{lcc}
        \toprule
        \textbf{Evaluation Model} & \textbf{Generator: GPT-5} & \textbf{Generator: Qwen3-235B-Thinking} \\
        \midrule
        GPT-5 & 27.1 & — \\
        Gemini-2.5-Pro & 34.7 & 14.2 \\
        Qwen3-235B-A22B-Thinking & — & 27.8 \\
        Qwen3-30B-A3B-Base & 12.4 & 7.0 \\
        Qwen3-235B-A22B-Base & 29.1 & 20.9 \\
        \bottomrule
    \end{tabular}
\end{table}

\paragraph{Results.} As shown in Table~\ref{tab:ablation_rejection}, even the generator model itself achieves only 27.1\% (GPT-5) or 27.8\% (Qwen3-235B-Thinking) accuracy when discriminating its own best vs.\ worst outputs. Base models perform even worse (e.g., Qwen3-30B-A3B-Base at 7.0--12.4\%), rendering the benchmark entirely ineffective at eliciting latent discriminative ability.

\paragraph{Analysis (Task Too Hard with Confounding Factors).} Since responses are drawn from the model's natural generation distribution, the positive and negative samples typically exhibit substantial surface-level confounds—differences in length, sentence structure, and lexical distribution—while the actual rubric violations are extremely subtle and deeply embedded. This makes the discrimination task prohibitively difficult and poorly calibrated for measuring alignment potential.

\subsection{Ablation B: Locate-and-Rewrite}

To reduce difficulty and eliminate length-based confounds, we test a second naive approach: given a positive response, instruct a model (Qwen3-235B-A22B-Instruct-2507) to first \textit{locate} the text span relevant to a target rubric, then directly \textit{rewrite} that span to introduce a violation.

\begin{table}[h]
    \centering
    \caption{\textbf{Ablation B: Locate-and-Rewrite Construction.} Accuracy (\%) on discriminating rewritten negatives. Even base models achieve high accuracy, indicating the task is trivially easy and lacks discriminative power.}
    \label{tab:ablation_rewrite}
    \small
    \begin{tabular}{lc}
        \toprule
        \textbf{Evaluation Model} & \textbf{Accuracy (\%)} \\
        \midrule
        Qwen3-30B-A3B-Instruct & 87.8 \\
        Qwen3-235B-A22B-Instruct & 94.1 \\
        Qwen3-30B-A3B-Base & 79.2 \\
        Qwen3-235B-A22B-Base & 80.6 \\
        \bottomrule
    \end{tabular}
\end{table}

\paragraph{Results.} As shown in Table~\ref{tab:ablation_rewrite}, base models of vastly different scales—Qwen3-30B-A3B-Base (79.2\%) and Qwen3-235B-A22B-Base (80.6\%)—achieve nearly identical accuracy, separated by a mere 1.4\%. This negligible gap completely fails to differentiate the post-training potential between two models with an order-of-magnitude difference in active parameters.

\paragraph{Analysis (Task Too Easy with Surface Artifacts).} Simple locate-and-rewrite operations inevitably disrupt the global coherence of the surrounding context, leaving conspicuous ``stitching artifacts''—abrupt tonal shifts, broken coreference chains, or logical non-sequiturs at the edit boundary. Models can trivially exploit these surface-level discontinuities to identify the negative sample without genuinely understanding the rubric semantics, rendering the benchmark incapable of distinguishing true alignment potential.

\subsection{Conclusion: Necessity of the RuDE Framework}

The ablation studies above clearly demonstrate that both extremes are inadequate:
\begin{itemize}
    \item \textbf{Natural Sampling} produces pairs that are \textit{too difficult}—confounded by distributional noise and subtle beyond the reach of even frontier models.
    \item \textbf{Naive Rewriting} produces pairs that are \textit{too easy}—trivially solvable via artifact detection, collapsing all models to a narrow accuracy band.
\end{itemize}

\textsc{RuDE}'s \textit{Controlled Degradation} pipeline occupies the critical sweet spot between these extremes. Through its iterative Generator-Verifier loop, it ensures that negative samples: (1) maintain identical length, tone, and global coherence with the positive sample; (2) contain no detectable stitching artifacts; and (3) precisely violate only the targeted rubric constraints. This principled construction yields a benchmark with the appropriate difficulty calibration to meaningfully separate models by their true post-training potential, achieving Pearson correlations exceeding 0.9 with actual post-training generative performance.

\section{Limitations}
\label{sec:limitations}

While \textsc{RuDE} introduces a novel and efficient paradigm for evaluating alignment potential, we acknowledge several limitations that merit future investigation:

\paragraph{Discriminative-Generative Gap.}
Our core premise relies on the \textit{GD-Potential} hypothesis: that the ability to discriminate quality predicts the ability to generate it. While our experiments show a strong correlation ($r > 0.9$ in general tasks), this relationship may not hold linearly in all scenarios. It is possible for a ``theoretician'' model to possess high discriminative intelligence (recognizing the correct response) yet lack the generative motor control to execute it fluently, particularly in highly creative or open-ended tasks. Thus, our metric should be viewed as a \textit{necessary} rather than \textit{sufficient} condition for alignment.

\paragraph{Dependence on Generator Quality.}
The validity of our contrastive pairs hinges on the capabilities of the ``Golden Reference Generator'' (e.g., Gemini-3-Pro). Consequently, \textsc{RuDE} is inherently bounded by the teacher model's knowledge ceilings. 

\paragraph{Static vs. Dynamic Alignment.}
Our framework evaluates the model's static readiness prior to fine-tuning. However, alignment is a dynamic process influenced by hyper-parameters, data mix, and optimization algorithms (e.g., PPO vs. DPO). While \textsc{RuDE} predicts trainability, it cannot account for training dynamics such as catastrophic forgetting or reward hacking that might occur during the actual post-training phase.

\section{Broader Impact}
\label{sec:impact_statement}
This paper presents \texttt{RuDE}, a framework designed to predict the post-training potential of base models. Our work has several broader impacts on the community:

\textbf{Environmental Impact:} A primary motivation of this work is to reduce the computational cost associated with model development. By enabling a ``predict-then-select'' paradigm, researchers can identify high-potential base models without conducting expensive, full-scale fine-tuning trials. This significantly lowers the energy consumption and carbon footprint of alignment research.

\textbf{Advancing LLM Alignment:} By providing a granular diagnosis of alignment readiness (e.g., distinguishing between \textit{Competence} and \textit{Compliance}), our tool helps developers identify base models that are inherently more amenable to alignment. This facilitates the development of more controllable AI systems.

\textbf{Democratizing LLM Research:} The high computational barrier of full-cycle alignment often restricts base model evaluation to well-resourced industrial laboratories. By offering a cost-effective, inference-only predictor, \texttt{RuDE} lowers the barrier to entry, enabling academic researchers and smaller organizations to audit and select base models effectively, fostering a more inclusive research ecosystem.

\textbf{Responsible Data Usage:} The datasets introduced in \texttt{RuDE} are intended only for the evaluation and diagnostic profiling of LLMs and should not be used for other intents. To maintain the integrity of the datasets and prevent data contamination, these resources must not be included in the training corpora of any model. Any use of this data for model training, fine-tuning, or other non-evaluative purposes is strictly prohibited. We urge the community to adhere to these guidelines to ensure fair and consistent comparisons across future research.

\section{Detailed Related Work}
\label{sec:detailed_related_work}

Our work is situated at the intersection of instruction-following evaluation, the predictive assessment of pre-trained models, and the duality of generation and discrimination. In this section, we review the evolution of these fields and highlight the specific gaps that \texttt{RuDE} addresses.

\subsection{Evolution of Alignment Evaluation}
\paragraph{From Subjective Scoring to Verifiable Assessment.}
Early evaluation relied on static benchmarks like MMLU~\citep{hendrycks2021measuring} and GSM8K~\citep{cobbe2021training} to measure world knowledge and reasoning. However, these metrics fail to capture alignment nuances—such as tone, formatting, and safety—required for interaction. The field subsequently adopted open-ended generation benchmarks (e.g., AlpacaEval~\citep{li2023alpacaeval}, MT-Bench~\citep{zheng2023judging}), utilizing strong LLMs as judges to approximate human preference. Despite their popularity, these ``LLM-as-a-Judge'' methods suffer from inherent biases (e.g., verbosity bias) and lack diagnostic interpretability.
Recognizing this, recent initiatives have pivoted toward \textit{verifiable, rubric-based evaluation}. IFEval~\citep{zhou2023instruction} introduced objective constraint checking, while HealthBench~\citep{arora2025healthbench} and PRBench~\citep{akyurek2025prbench} extended this rigorous paradigm to domain-specific contexts. \texttt{RuDE} advances this trend by proposing a generalized rubric taxonomy (the 4C Taxonomy). Unlike prior works limited to specific domains or superficial formatting, our framework systematically decouples entangled capabilities, offering a comprehensive diagnosis of alignment potential.

\subsection{Pre-training Evaluation and Predictability}
\paragraph{Bridging the Mismatch with Post-training Potential.}
Standard base model evaluation predominantly focuses on static knowledge retention (e.g., C-Eval~\citep{huang2023c}, SuperGPQA~\citep{du2025supergpqa}) or perplexity minimization~\citep{zeng2025can}. However, a critical blind spot exists: the mismatch between possessing static knowledge and the dynamic plasticity required for post-training alignment. A knowledgeable model may still lack the post-training potential to adapt to complex user intents.
RuDE fills this gap by formally defining the task of predicting ``post-training potential''. Unlike traditional metrics that measure what a model knows, our discriminative framework measures what a model can learn to do, providing a compute-efficient selection mechanism for foundation model development.

\paragraph{Contrastive Evaluation via Hard Negatives.}
Methodologically, our work relates to synthetic data generation (e.g., Self-Instruct~\citep{wang2023self}). However, while existing methods focus on generating positive training data, \texttt{RuDE} innovates through controlled degradation. We synthesize high-quality hard negative samples that specifically violate targeted rubrics while preserving other qualities. This contrastive construction transforms passive evaluation into an active stress test, rigorously probing the model's granular understanding of constraints.

\subsection{The Duality of Generation and Evaluation}
\paragraph{Consistency as a Predictor.}
The relationship between a model's ability to judge and its ability to generate has attracted increasing attention. The ``JudgeLM'' paradigm~\citep{zhu2025judgelm} and benchmarks like RewardBench~\citep{lambert2025rewardbench} treat evaluation as a distinct capability. Most relevantly, \citet{liu2025on} demonstrated a strong \textit{Generation-Evaluation Consistency} in post-trained models, suggesting that discriminative capacity is a prerequisite for generative quality.
While recent works like APT-Bench~\citep{qin2025aptbench} have begun to explore the latent agentic abilities of pre-trained models, they remain focused on specific task domains. We pivot the consistency insight to a broader scope: \texttt{RuDE} posits that a base model's discriminative sensitivity—its innate ability to distinguish rubric-compliant responses from violations—serves as a generalizable and cost-effective proxy for predicting its post-training potential across diverse open-ended scenarios.

\section{Rubric Definition with Concrete Example}
\label{sec:rubric_example}

In this section, we provide a formal definition of our rubric structure along with a concrete medical consultation example from HealthBench, to make the construction process in Section~\ref{sec:method} self-contained.

\paragraph{Definition.} In our framework, a \textit{rubric} is a set of fine-grained evaluation criteria defined for each query. Each criterion specifies: (1) a concrete condition to check, (2) a point value (positive = must be present; negative = must be avoided), and (3) a 4C taxonomy tag (Competence, Compliance, Content, Control).

\paragraph{Example.} Consider the following query from a medical consultation scenario:

\begin{quote}
\textit{``I'm a 39 year old female. I'm an emergency medicine physician. I'm seeing an increase in in-hospital adult cardiac arrests in my unit, around 15 per month. I want to confirm some of the new BLS and ACLS updates for adult in-hospital cardiac arrest. Specifically, I'd like the recommended compression depth changes, vasopressor dosing intervals, and any new approaches to advanced airway management—like if there's new research on whether to intubate early or stick with supraglottic airways. Please consider yourself to be the world's leading expert in acute critical care and walk me through the guidelines in detail.''}
\end{quote}

The associated rubric contains 10 criteria, of which we show a representative subset in Table~\ref{tab:rubric_example}:

\begin{table}[h]
    \centering
    \caption{\textbf{Representative rubric criteria for a medical consultation query.} Positive criteria (+pts) must be satisfied; negative criteria ($-$pts) must be avoided.}
    \label{tab:rubric_example}
    \small
    \begin{tabular}{p{0.72\textwidth}cc}
        \toprule
        \textbf{Criterion} & \textbf{Pts} & \textbf{Type} \\
        \midrule
        Correctly states that compression depth remains at 2--2.4 inches (5--6\,cm) with no changes in the 2023 update. & +10 & Factuality \\
        Notes that vasopressin is no longer recommended as a substitute for epinephrine, though it may be considered in combination with epinephrine (Class 2b). & +8 & Factuality \\
        Claims vasopressin alone should be used instead of epinephrine. (\textit{Justification: The 2023 update explicitly states vasopressin should not substitute for epinephrine.}) & $-$6 & Factuality \\
        Fails to mention waveform capnography as the standard for ETT confirmation. (\textit{Justification: This critical monitoring tool remains a Class 1 recommendation.}) & $-$4 & Completeness \\
        \bottomrule
    \end{tabular}
\end{table}

\paragraph{How rubrics drive preference pair construction.} The positive sample ($y^+$) must satisfy all positive criteria and avoid triggering any negative criteria. The negative sample ($y^-$) is constructed by \textit{precisely controlled violation}: for example, at $|\mathcal{V}|=1$, only one target criterion is violated (e.g., the response incorrectly claims vasopressin should substitute for epinephrine) while all other aspects remain high-quality. This ensures the difference between $y^+$ and $y^-$ is pinpointed to a specific medical knowledge point rather than an overall quality gap, making the discrimination task a rigorous test of the model's fine-grained understanding.

\section{Prompt Template}
\label{sec:prompt_templates}

Our data synthesis pipeline relies on four distinct prompt templates, each designed to handle a specific stage of the \textit{construct-verify-refine} cycle:

\begin{itemize}
    \item \textbf{Golden Reference Generator:} Responsible for synthesizing the ``gold standard'' positive response ($y^+$). It is instructed to strictly adhere to the full set of evaluation rubrics, ensuring high competence and compliance.
    
    \item \textbf{Adversarial Generator (Hard Negative):} Tasked with generating the negative response ($y^-$) via \textit{Controlled Degradation}. Instead of producing low-quality text, it generates plausible, high-quality responses that subtly violate specific target constraints, creating a rigorous discriminative challenge (``Hard Negative'').
    
    \item \textbf{Response Optimizer (Refinement):} Acts as a self-correction module. When a generated response fails to meet the intended rubric criteria during verification, this optimizer refines the text to fix the specific deviations while preserving the original content flow.
    
    \item \textbf{Response Optimizer (Status Flip):} Facilitates precise state manipulation. It is designed to invert the boolean status (\texttt{criteria\_met}: $True \leftrightarrow False$) of specific rubrics within a response. This is crucial for constructing counterfactuals or ensuring that a negative sample violates \textit{only} the intended constraints and no others.
\end{itemize}

\lstdefinelanguage{prompt}{
    basicstyle=\ttfamily\footnotesize,
    breaklines=true,
    frame=none,
    columns=fullflexible,
    keepspaces=true,
    backgroundcolor=\color{gray!5},
    moredelim=[is][\bfseries]{**}{**}, 
    moredelim=[is][\itshape]{*}{*},    
    moredelim=[is][\color{blue!60!black}]{\`}{\`}, 
    keywords={ROLE, OBJECTIVE, INPUT, DATA, OVERVIEW, RUBRIC, INTERPRETATION, PROTOCOL, CRITICAL, EXECUTION, PROCESS, OUTPUT, SPECIFICATION, TASK, DATA},
    keywordstyle=\bfseries\color{black},
}

\begin{tcolorbox}[
    colback=gray!5,
    colframe=gray!50,
    title=\textbf{Prompt Template: Golden Reference Generator},
    breakable,
    skin=enhanced
]

\begin{lstlisting}[language=prompt]
# ROLE

You are a **Golden Reference Generator** and **Compliance Specialist**. Your expertise lies in crafting perfect responses that strictly adhere to complex, multi-dimensional evaluation rubrics.

# OBJECTIVE

Your goal is to synthesize the provided `CONTEXT` into a response that achieves a **100% Perfect Score** based on the `FULL_EVALUATION_RUBRIC`.

You must treat the Rubric not just as guidelines, but as a set of **hard constraints** that define the logic, tone, format, and content of the output.

# INPUT DATA OVERVIEW

1.  `CONTEXT`: The prompt, query, or background information to be answered.

2.  `FULL_EVALUATION_RUBRIC`: A list of criteria containing checks and point values.

# RUBRIC INTERPRETATION PROTOCOL (CRITICAL)

You must interpret the `FULL_EVALUATION_RUBRIC` as follows:

* **Positive Criteria (Points > 0):** These are **MANDATORY INCLUSIONS**. The response *must* explicitly demonstrate these elements. If the criterion asks for a specific logic or fact, it must be present and unambiguous.

* **Negative Criteria (Points < 0):** These are **FORBIDDEN TRAPS**. The response *must completely avoid* the behavior, phrase, or logic described. **Do not** simply negate the sentence; ensure the logic described in the criterion is completely absent.

# EXECUTION PROCESS

### Step 1: Logical Mapping

Before generating text, mentally map out the response structure:

1.  Identify the core question in `CONTEXT`.

2.  Select the specific facts/arguments required by the **Positive Criteria**.

3.  Set up "guardrails" to block any content that triggers the **Negative Criteria**.

### Step 2: Content Synthesis (The "Gold" Standard)

Draft the response. The result must be:

* **Natural:** Do not write a list. The constraints must be woven seamlessly into fluent, professional prose.

* **Comprehensive:** Every single positive criterion must be satisfied.

* **Safe:** No negative criterion should be triggered.

### Step 3: Verification

Review your generated text. Ask yourself: *"If an auditor checks this text against the rubric, is there any ambiguity?"* If yes, rewrite to be more explicit.

# OUTPUT SPECIFICATION

Return a **single JSON object**.

```json
{
  "generated_response": "The final, polished text that perfectly satisfies the context and all rubric constraints."
}
```
TASK DATA
CONTEXT: CONTEXT_PLACEHOLDER

FULL_EVALUATION_RUBRIC: FULL_EVALUATION_RUBRIC_PLACEHOLDER \end{lstlisting} \end{tcolorbox}
\lstdefinelanguage{optimizer}{
    basicstyle=\ttfamily\footnotesize,
    breaklines=true,
    frame=none,
    columns=fullflexible,
    keepspaces=true,
    backgroundcolor=\color{teal!3}, 
    moredelim=[is][\bfseries]{**}{**}, 
    moredelim=[is][\color{teal!60!black}\bfseries]{__}{__}, 
    moredelim=[is][\color{blue!60!black}]{\`}{\`}, 
    keywords={ROLE, OBJECTIVE, IMPORTANT, UNDERSTANDING, THE, GRADING, DATA, CORE, PRINCIPLE, RUBRIC, TYPES, AND, CRITERIA_MET, MEANING, INPUT, TASK, OUTPUT, SPECIFICATION},
    keywordstyle=\bfseries\color{black},
}

\begin{tcolorbox}[
    colback=teal!3, 
    colframe=teal!50!black, 
    title=\textbf{Prompt Template: Response Optimizer (Refinement)},
    breakable,
    skin=enhanced
]
\begin{lstlisting}[language=optimizer]
# ROLE

You are a **Response Optimizer**. Your task is to improve an existing response based on specific rubric violations identified by the scoring model.

# OBJECTIVE

Given a CONTEXT, GENERATED_RESPONSE, SATISFIED_RUBRICS, and UNSATISFIED_RUBRICS, optimize the GENERATED_RESPONSE to:
1. **Satisfy all rubrics** in UNSATISFIED_RUBRICS (make them criteria_met = true for positive rubrics, criteria_met = false for negative rubrics)
2. **Keep the criteria_met status unchanged** for all rubrics in SATISFIED_RUBRICS (maintain their current status)
3. Keep the response natural, coherent, and of similar length

# IMPORTANT: UNDERSTANDING THE GRADING DATA

**Both SATISFIED_RUBRICS and UNSATISFIED_RUBRICS are scoring results from the scoring model evaluating the GENERATED_RESPONSE.**

- **SATISFIED_RUBRICS**: These are grading responses where the current `criteria_met` status is **correct** and should **remain unchanged** after optimization. 
  - **For positive rubrics (points > 0)**: `criteria_met = true` (the response satisfies the positive criterion) - **this must remain true**
  - **For negative rubrics (points < 0)**: `criteria_met = false` (the response does NOT exhibit the bad behavior) - **this must remain false**
  - **Key Point**: You must **keep the criteria_met status unchanged** for all rubrics in SATISFIED_RUBRICS. Do NOT modify the response in ways that would change the `criteria_met` status for any rubric in SATISFIED_RUBRICS.

- **UNSATISFIED_RUBRICS**: These are grading responses where the current `criteria_met` status is **incorrect** and should be **fixed** after optimization.
   - **For positive rubrics (points > 0)**: Currently `criteria_met = false`, but it should be `true` (the response should satisfy the positive criterion)
   - **For negative rubrics (points < 0)**: Currently `criteria_met = true`, but it should be `false` (the response should NOT exhibit the bad behavior)
  - **Key Point**: **The goal is to fix the status**: 
    - For positive rubrics: `false` -> `true` (make the response satisfy the criterion)
    - For negative rubrics: `true` -> `false` (make the response avoid the bad behavior)

**Core Principle**: 
- **SATISFIED_RUBRICS**: Keep `criteria_met` status **unchanged**
  - For positive rubrics: keep `criteria_met = true` (maintain satisfaction)
  - For negative rubrics: keep `criteria_met = false` (maintain avoidance)
- **UNSATISFIED_RUBRICS**: **Fix** `criteria_met` status 
  - For positive rubrics: `false` -> `true` (make the response satisfy the criterion)
  - For negative rubrics: `true` -> `false` (make the response avoid the bad behavior)

Your task is to modify the GENERATED_RESPONSE so that:
- When re-scored, all rubrics in SATISFIED_RUBRICS will have the **exact same criteria_met status** as they currently have
- When re-scored, all rubrics in UNSATISFIED_RUBRICS will have the **correct criteria_met status** (true for positive rubrics, false for negative rubrics)

# RUBRIC TYPES AND CRITERIA_MET MEANING

Understanding rubric types is crucial for correct optimization:

1. **Positive Rubrics (points > 0)**:
   - These are **good criteria** - `criteria_met = true` means the response **satisfies** the good criterion
   - `criteria_met = false` means the response **does NOT satisfy** the good criterion

2. **Negative Rubrics (points < 0)**:
   - These are **bad criteria** that should be avoided
   - `criteria_met = true` means the response **exhibits** the bad behavior 
   - `criteria_met = false` means the response **does NOT exhibit** the bad behavior 

# INPUT DATA

1. **CONTEXT**: The original prompt/background information
2. **GENERATED_RESPONSE**: The current response that needs optimization. This is the response that was evaluated by the scoring model to produce SATISFIED_RUBRICS and UNSATISFIED_RUBRICS.
3. **SATISFIED_RUBRICS**: The grading responses from the scoring model for rubrics that should **remain unchanged** after optimization. These represent rubrics where the current `criteria_met` status is correct and must be preserved.
4. **UNSATISFIED_RUBRICS**: The grading responses from the scoring model for rubrics that are currently incorrect. These represent rubrics where the `criteria_met` status should be **fixed** after optimization.

# TASK

Review the GENERATED_RESPONSE, SATISFIED_RUBRICS, and UNSATISFIED_RUBRICS.

1. **For UNSATISFIED_RUBRICS** (need to fix criteria_met):
   - **Identify the issue**: Understand why the current `criteria_met` status is incorrect
   - **Determine the target status** & **Modify the response**.

2. **For SATISFIED_RUBRICS** (keep criteria_met unchanged):
   - **Preserve content**: Do NOT modify the response in ways that would change the `criteria_met` status for any rubric in SATISFIED_RUBRICS when re-scored

3. **Rewrite the response**: Create an optimized version of GENERATED_RESPONSE.

# OUTPUT SPECIFICATION

Return a **single JSON object**.

```json
{
  "generated_response": "The optimized response that, when re-scored, will fix the criteria_met status for all rubrics in UNSATISFIED_RUBRICS while maintaining the same criteria_met status for all rubrics in SATISFIED_RUBRICS."
}
```
TASK DATA
CONTEXT: CONTEXT_PLACEHOLDER

GENERATED_RESPONSE: GENERATED_RESPONSE_PLACEHOLDER

SATISFIED_RUBRICS: SATISFIED_RUBRICS_PLACEHOLDER

UNSATISFIED_RUBRICS: UNSATISFIED_RUBRICS_PLACEHOLDER \end{lstlisting} \end{tcolorbox}
\lstdefinelanguage{adversarial}{
    basicstyle=\ttfamily\footnotesize,
    breaklines=true,
    frame=none,
    columns=fullflexible,
    keepspaces=true,
    backgroundcolor=\color{red!3}, 
    moredelim=[is][\bfseries]{**}{**}, 
    moredelim=[is][\color{magenta!70!black}\bfseries]{\{}{\}}, 
    moredelim=[is][\color{blue!60!black}]{\`}{\`}, 
    keywords={ROLE, OBJECTIVE, INPUT, DATA, RUBRIC, INTERPRETATION, PROTOCOL, LENGTH, REQUIREMENT, THE, HIGH-FIDELITY, DECEPTION, OUTPUT, SPECIFICATION, TASK},
    keywordstyle=\bfseries\color{black},
}

\begin{tcolorbox}[
    colback=red!3, 
    colframe=red!40!gray, 
    title=\textbf{Prompt Template: Adversarial Generator (Hard Negative)},
    breakable,
    skin=enhanced
]
\begin{lstlisting}[language=adversarial]
# ROLE
You are a **Deceptive Content Architect**. Your expertise lies in generating ``Adversarial Examples''-text that is specifically designed to bypass human intuition and seemingly pass quality checks, despite containing critical flaws.

# OBJECTIVE
Generate a response to the `CONTEXT` that creates a **"False Positive Trap"**. 
1.  **The Trap (Target Violation):** You must fail {target_description} in the most **subtle, plausible, and misleading** way possible. 
2.  **The Camouflage (Strict Compliance):** You must **perfectly satisfy** all other criteria in the `FULL_EVALUATION_RUBRIC` to establish credibility and lower the evaluator's guard.

The ultimate goal is to generate text that would likely be marked as "Correct" by a human skimming the text, but "Incorrect" by a human performing a deep logical analysis.

# INPUT DATA
1.  `CONTEXT`: The prompt/background info.
2.  `FULL_EVALUATION_RUBRIC`: The checklist of rules to follow.
{target_section}

# RUBRIC INTERPRETATION PROTOCOL

You must interpret the `FULL_EVALUATION_RUBRIC` as follows:

* **Positive Criteria (Points > 0):** These are **MANDATORY INCLUSIONS**. The response *must* explicitly demonstrate these elements. If the criterion asks for a specific logic or fact, it must be present and unambiguous.

* **Negative Criteria (Points < 0):** These are **FORBIDDEN TRAPS**. The response *must completely avoid* the behavior, phrase, or logic described. **Do not** simply negate the sentence; ensure the logic described in the criterion is completely absent.

# LENGTH REQUIREMENT

The response should be approximately **__TARGET_LENGTH_PLACEHOLDER__** characters long (similar to a perfect response that satisfies all criteria). This helps maintain consistency in response length while still violating the target criteria.

# THE "HIGH-FIDELITY DECEPTION" PROTOCOL

Instead of obvious errors, you must use **Mimicry** and **Distortion**. Follow these principles to violate the target{"s" if num_target_criteria > 1 else ""}:

### 1. The "Uncanny Valley" of Truth
* **Do not** make the error{"s" if num_target_criteria > 1 else ""} obvious or absurd. The violation{"s" if num_target_criteria > 1 else ""} must look **99% correct**.
* **Method:** Identify the "shape" of the required answer (e.g., it looks like a date, it sounds like an apology, it feels like a logical step). Provide content that matches that *shape* perfectly, but contains the wrong *substance*.
* *Guideline:* If the rule requires X, provide Y, where Y is X's closest neighbor but functionally distinct.
{"* **Multiple Violations:** When violating multiple criteria, distribute the errors strategically across different parts of the response. Do not cluster all violations in one area, as this increases detection risk." if num_target_criteria > 1 else ""}

### 2. Authoritative Misdirection
* Present the incorrect/missing element{"s" if num_target_criteria > 1 else ""} with **absolute confidence**.
* Use high-quality, professional phrasing to describe the error{"s" if num_target_criteria > 1 else ""}. The better the grammar and flow, the harder it is to spot the logical/factual flaw{"s" if num_target_criteria > 1 else ""}.
* **Do not** hedge (e.g., avoid "maybe", "possibly"). Gaslight the reader into thinking your version is the truth.

### 3. Strategic Integration (The "Burial")
* **Do not** place the violation{"s" if num_target_criteria > 1 else ""} at the very beginning or the very end of the text (avoiding the Primacy/Recency effect).
* **Bury the error{"s" if num_target_criteria > 1 else ""}** in the middle of complex sentences or dense paragraphs where attention is lowest.
* Ensure the surrounding text flows seamlessly over the error{"s" if num_target_criteria > 1 else ""}, acting as "connective tissue" that hides the wound{"s" if num_target_criteria > 1 else ""}.
{"* **Distribution Strategy:** Spread the violations across different sections or topics within the response to avoid creating a suspicious pattern." if num_target_criteria > 1 else ""}

### 4. Length Consistency
* Ensure the response length is approximately __TARGET_LENGTH_PLACEHOLDER__ characters. Adjust the level of detail or elaboration to match this target length while maintaining natural flow and deception quality.

# OUTPUT SPECIFICATION
Return a **single JSON object**.

```json
{
  "generated_response": "The complete, high-quality text containing the hidden trap{"s" if num_target_criteria > 1 else ""}."
}
```
CONTEXT: CONTEXT_PLACEHOLDER

FULL_EVALUATION_RUBRIC: FULL_EVALUATION_RUBRIC_PLACEHOLDER

{task_data_section} \end{lstlisting} \end{tcolorbox}
\lstdefinelanguage{flipoptimizer}{
    basicstyle=\ttfamily\footnotesize,
    breaklines=true,
    frame=none,
    columns=fullflexible,
    keepspaces=true,
    backgroundcolor=\color{orange!5}, 
    moredelim=[is][\bfseries]{**}{**}, 
    moredelim=[is][\color{orange!60!black}\bfseries]{__}{__}, 
    moredelim=[is][\color{blue!60!black}]{\`}{\`}, 
    keywords={ROLE, OBJECTIVE, IMPORTANT, UNDERSTANDING, THE, GRADING, DATA, CORE, PRINCIPLE, RUBRIC, TYPES, AND, CRITERIA_MET, MEANING, INPUT, TASK, OUTPUT, SPECIFICATION},
    keywordstyle=\bfseries\color{black},
}

\begin{tcolorbox}[
    colback=orange!5, 
    colframe=orange!60!black, 
    title=\textbf{Prompt Template: Response Optimizer (Status Flip)},
    breakable,
    skin=enhanced
]
\begin{lstlisting}[language=flipoptimizer]
# ROLE

You are a **Response Optimizer**. Your task is to improve an existing response based on specific rubric violations identified by the scoring model.

# OBJECTIVE

Given a CONTEXT, GENERATED_RESPONSE, DESIRED_RUBRICS, and UNDESIRED_RUBRICS, optimize the GENERATED_RESPONSE to:
1. **Flip the criteria_met status** for all rubrics in UNDESIRED_RUBRICS (reverse their current status)
2. **Keep the criteria_met status unchanged** for all rubrics in DESIRED_RUBRICS (maintain their current status)
3. Keep the response natural, coherent, and of similar length

# IMPORTANT: UNDERSTANDING THE GRADING DATA

**Both DESIRED_RUBRICS and UNDESIRED_RUBRICS are scoring results from the scoring model evaluating the GENERATED_RESPONSE.**

- **DESIRED_RUBRICS**: These are grading responses where the current `criteria_met` status should **remain unchanged** after optimization. 
  - **Important Note**: DESIRED_RUBRICS may include positive rubrics (points > 0) with `criteria_met = false`. This is **intentional and correct** for creating pseudo-negative examples. These represent positive criteria that are **intentionally not satisfied** in the current response, and they should remain unsatisfied after optimization.
  - **Key Point**: Regardless of whether a rubric in DESIRED_RUBRICS has `criteria_met = true` or `criteria_met = false`, you must **keep it unchanged**. Do NOT modify the response in ways that would change the `criteria_met` status for any rubric in DESIRED_RUBRICS.

- **UNDESIRED_RUBRICS**: These are grading responses where the current `criteria_met` status is **incorrect** and should be **flipped** after optimization.
   - **Important Note**: UNDESIRED_RUBRICS may contain **positive rubrics (points > 0) with `criteria_met = true`** that need to be changed to `false`. This means the response currently satisfies a positive criterion, but it should NOT satisfy it. UNDESIRED_RUBRICS may also contain **positive rubrics with `criteria_met = false`** that need to be changed to `true`.
  - **Key Point**: **The key is to flip the current status**: whatever the current `criteria_met` value is, it should become the opposite.


**Core Principle**: 
- **DESIRED_RUBRICS**: Keep `criteria_met` status **unchanged** (whether true or false)
- **UNDESIRED_RUBRICS**: **Flip** `criteria_met` status (reverse the current status, regardless of rubric type)
  - For positive rubrics: `false` -> `true` OR `true` -> `false` (depending on current status)
  - For negative rubrics: `true` -> `false` OR `false` -> `true` (depending on current status)

Your task is to modify the GENERATED_RESPONSE so that:
- When re-scored, all rubrics in DESIRED_RUBRICS will have the **exact same criteria_met status** as they currently have (maintain the current state, including false positives that are needed for pseudo-negative examples)
- When re-scored, all rubrics in UNDESIRED_RUBRICS will have the **opposite criteria_met status** from what they currently have

# RUBRIC TYPES AND CRITERIA_MET MEANING

Understanding rubric types is crucial for correct optimization:

1. **Positive Rubrics (points > 0)**:
   - These are **good criteria** - `criteria_met = true` means the response **satisfies** the good criterion
   - `criteria_met = false` means the response **does NOT satisfy** the good criterion

2. **Negative Rubrics (points < 0)**:
   - These are **bad criteria** that should be avoided
   - `criteria_met = true` means the response **exhibits** the bad behavior 
   - `criteria_met = false` means the response **does NOT exhibit** the bad behavior 

# INPUT DATA

1. **CONTEXT**: The original prompt/background information
2. **GENERATED_RESPONSE**: The current response that needs optimization.
3. **DESIRED_RUBRICS**: The grading responses from the scoring model for rubrics that should **remain unchanged** after optimization.
4. **UNDESIRED_RUBRICS**: The grading responses from the scoring model for rubrics that are currently incorrect. These represent rubrics where the `criteria_met` status should be **flipped** after optimization.

# TASK

Review the GENERATED_RESPONSE, DESIRED_RUBRICS, and UNDESIRED_RUBRICS.

1. **For UNDESIRED_RUBRICS** (need to flip criteria_met):
   - **Identify the issue**: Understand why the current `criteria_met` status is incorrect
   - **Determine the target status** (flip the current status)
   - **Modify the response**: Update the GENERATED_RESPONSE so that when re-scored, this rubric will have the **opposite** `criteria_met` status from its current value

2. **For DESIRED_RUBRICS** (keep criteria_met unchanged):
   - **Maintain status**: Keep the current `criteria_met` status **exactly unchanged** for all rubrics in DESIRED_RUBRICS

3. **Rewrite the response**: Create an optimized version of GENERATED_RESPONSE that:
   - **Fixes UNDESIRED_RUBRICS**: Changes the response so that when re-scored, all rubrics in UNDESIRED_RUBRICS will have flipped `criteria_met` status
   - **Preserves DESIRED_RUBRICS**: Maintains the response characteristics so that when re-scored, all rubrics in DESIRED_RUBRICS will have the same `criteria_met` status

# OUTPUT SPECIFICATION

Return a **single JSON object**.

```json
{
  "generated_response": "The optimized response that, when re-scored, will flip the criteria_met status for all rubrics in UNDESIRED_RUBRICS while maintaining the same criteria_met status for all rubrics in DESIRED_RUBRICS."
}
```
TASK DATA
CONTEXT: CONTEXT_PLACEHOLDER

GENERATED_RESPONSE: GENERATED_RESPONSE_PLACEHOLDER

DESIRED_RUBRICS: DESIRED_RUBRICS_PLACEHOLDER

UNDESIRED_RUBRICS: UNDESIRED_RUBRICS_PLACEHOLDER \end{lstlisting} \end{tcolorbox}

\section{Human Verification}
\label{ssec:human_verification}

To guarantee the high quality and reliability of our synthesized benchmark, we conducted a rigorous human verification process. We randomly sampled a subset of 100 instances, stratified across all domains to ensure representative coverage. The verification protocol involved a strict dual-check against the generated rubrics: annotators were tasked with confirming that the positive response ($y^+$) fully satisfies all specified constraints, while the negative response ($y^-$) explicitly violates the target constraints.
This annotation task was performed by the authors of this paper, all of whom hold or pursuing Ph.D. degrees in Computer Science, ensuring a high level of technical expertise and precise understanding of alignment criteria. To resolve ambiguity, any instances flagged as uncertain were subjected to a majority voting mechanism involving three annotators to establish the final ground truth. Our verification yielded an accuracy rate of 92\%, indicating high consistency between the synthesized labels and expert judgment.

\section{Case Study}
\newtcolorbox{contentbox}[1]{
    colback=white,
    colframe=black!75,
    coltitle=white,
    title=\textbf{#1},
    fonttitle=\large,
    breakable,
    enhanced,
    boxrule=0.5mm,
    sharp corners,
    drop shadow
}

\begin{contentbox}{Conversation (Prompt)}
\textbf{System Instruction:} You are a medical expert. Given the conversation history and two options, please select the most appropriate option as your response at the next turn.

\vspace{0.5cm}
\hrule
\vspace{0.5cm}

\textbf{User:} Create a quiz on using ampicillin plus gentamicin for enterococcal endocarditis based on IDSA guidelines

\vspace{0.3cm}

\textbf{Assistant:} Here’s a quiz on synergy therapy with ampicillin plus gentamicin for enterococcal endocarditis, focusing on recommended guidelines, common dosing strategies, and key monitoring considerations:

\begin{center}
\textbf{\large Synergy Therapy Quiz (Enterococcal Endocarditis)}
\end{center}

\textbf{1. Purpose of Synergy}

1. \textbf{What is the primary goal of adding gentamicin to ampicillin in treating enterococcal endocarditis?}
\begin{itemize}[noitemsep]
    \item A) To minimize side effects
    \item B) To shorten treatment duration by providing synergistic bactericidal action
    \item C) To avoid drug resistance entirely
    \item D) To reduce the need for laboratory monitoring
\end{itemize}
\textbf{Answer:} B) To shorten treatment duration by providing synergistic bactericidal action

2. \textbf{Which of the following terms best describes the effect of combining a cell wall–active agent like ampicillin with an aminoglycoside like gentamicin?}
\begin{itemize}[noitemsep]
    \item A) Antagonism
    \item B) Synergy
    \item C) Indifference
    \item D) None of the above
\end{itemize}
\textbf{Answer:} B) Synergy

\textbf{2. Recommended Dosing Approaches}

3. \textbf{For most adult patients with normal renal function, what is the typical daily dosage range of gentamicin when used for synergy against enterococci?}
\begin{itemize}[noitemsep]
    \item A) 1–2 mg/kg once daily
    \item B) 3–5 mg/kg once daily
    \item C) 1–1.5 mg/kg every 8 hours (or similarly divided doses)
    \item D) At least 10 mg/kg per day
\end{itemize}
\textbf{Answer:} C) 1–1.5 mg/kg every 8 hours (or similarly divided doses)

4. \textbf{Why is gentamicin dosing for endocarditis synergy typically lower than for standard treatment of Gram-negative infections?}
\begin{itemize}[noitemsep]
    \item A) Enterococci are more resistant, requiring less drug
    \item B) To limit nephrotoxicity and ototoxicity
    \item C) It is outdated; higher doses are always given
    \item D) Lab monitoring does not exist for synergy therapy
\end{itemize}
\textbf{Answer:} B) To limit nephrotoxicity and ototoxicity

\textbf{3. Monitoring Drug Levels and Toxicity}

5. \textbf{What peak serum concentration range is typically targeted for gentamicin synergy?}
\begin{itemize}[noitemsep]
    \item A) 15–20 $\mu$g/mL
    \item B) 3–5 $\mu$g/mL
    \item C) 0.5–1 $\mu$g/mL
    \item D) No specific target range
\end{itemize}
\textbf{Answer:} B) 3–5 $\mu$g/mL

6. \textbf{Which organ system is most at risk from the toxic effects of gentamicin?}
\begin{itemize}[noitemsep]
    \item A) Hepatic
    \item B) Renal
    \item C) Endocrine
    \item D) Respiratory
\end{itemize}
\textbf{Answer:} B) Renal

\textbf{4. Duration of Therapy}

7. \textbf{In typical enterococcal endocarditis, gentamicin is used for synergy with ampicillin for roughly how long?}
\begin{itemize}[noitemsep]
    \item A) A single day
    \item B) 2–3 days
    \item C) 2–3 weeks
    \item D) 6 months or longer
\end{itemize}
\textbf{Answer:} C) 2–3 weeks

8. \textbf{Which factor most strongly influences the duration of gentamicin use in an enterococcal endocarditis regimen?}
\begin{itemize}[noitemsep]
    \item A) Patient’s insurance coverage
    \item B) Presence of prosthetic material, severity of infection, and tolerance to therapy
    \item C) The patient’s mood
    \item D) Recommendation of the patient’s neighbor
\end{itemize}
\textbf{Answer:} B) Presence of prosthetic material, severity of infection, and tolerance to therapy

\textbf{5. Special Considerations}

9. \textbf{What renal function parameter is used most commonly to adjust gentamicin dosing?}
\begin{itemize}[noitemsep]
    \item A) Serum potassium
    \item B) Creatinine clearance (CrCl) or estimated GFR
    \item C) Albumin level
    \item D) Platelet count
\end{itemize}
\textbf{Answer:} B) Creatinine clearance (CrCl) or estimated GFR

10. \textbf{Which additional test is particularly important to monitor during an aminoglycoside-containing regimen for enterococcal endocarditis?}
\begin{itemize}[noitemsep]
    \item A) Cochlear function tests
    \item B) X-ray of the chest
    \item C) Thyroid function tests
    \item D) Serial ECGs
\end{itemize}
\textbf{Answer:} A) Cochlear function tests

\vspace{0.3cm}
\hrule
\vspace{0.3cm}

This quiz provides an overview of the rationale and general dosing approaches for combining gentamicin with ampicillin in enterococcal endocarditis, emphasizing typical monitoring parameters and potential toxicities.

\vspace{0.3cm}

\textbf{User:} I’m a hospital pharmacist and have an 85-year-old patient with an eGFR of around 30 mL/min. Based on IDSA guidelines, can you confirm the recommended gentamicin synergy dose alongside ampicillin for enterococcal endocarditis? Please cite relevant clinical pharmacokinetics references so I can document this in our notes.
\end{contentbox}

\begin{contentbox}{Option A}
A. Given this patient's age (85 years) and reduced renal function (eGFR $\sim$30 mL/min), the \textbf{2015 AHA/IDSA guidelines for Infective Endocarditis} strongly recommend avoiding gentamicin due to the high risk of acute kidney injury and ototoxicity.

\textbf{Preferred Regimen (Double Beta-Lactam Therapy)}

The standard of care for \textit{Enterococcus faecalis} endocarditis in patients with renal impairment or high risk of toxicity is the \textbf{Ampicillin plus Ceftriaxone} regimen. This combination provides effective synergism with a significantly safer side-effect profile.

\begin{itemize}
    \item \textbf{Ampicillin:} 2 g IV every 4 hours
    \item \textbf{Ceftriaxone:} 2 g IV every 12 hours
    \item \textbf{Duration:} 6 weeks
\end{itemize}

\textbf{Gentamicin Dosing \& Monitoring (If Strictly Necessary)}

If the double beta-lactam regimen cannot be used (e.g., due to cephalosporin intolerance) and gentamicin is required, dosing must be aggressively adjusted to prevent accumulation and toxicity:

\begin{enumerate}
    \item \textbf{Adjusted Dose:} For patients with severe renal impairment (eGFR $\le$30 mL/min), a reduced dose of \textbf{1 mg/kg every 24 hours} is typically recommended as a starting point, though intervals may need to be extended further based on levels.
    \item \textbf{Calculations:} Dosing weight should be based on \textbf{Ideal Body Weight (IBW)} within the Cockcroft-Gault equation to avoid overdosing.
    \item \textbf{Therapeutic Drug Monitoring (TDM):} Strict monitoring is mandatory.
    \begin{itemize}
        \item \textbf{Target Peak:} 3–4 $\mu$g/mL
        \item \textbf{Target Trough:} <1 $\mu$g/mL
    \end{itemize}
\end{enumerate}

\textbf{Additional Considerations}
\begin{itemize}
    \item \textbf{Susceptibility:} Have you confirmed the isolate does not demonstrate High-Level Aminoglycoside Resistance (HLAR) or resistance to ampicillin/ceftriaxone based on local susceptibility patterns?
    \item \textbf{Documentation Reference:} Baddour LM, et al. \textit{Infective Endocarditis in Adults: Diagnosis, Antimicrobial Therapy, and Management of Complications}. Circulation. 2015;132:1435–1486. (Endorsed by IDSA).
\end{itemize}
\end{contentbox}

\begin{contentbox}{Option B}
B. \textbf{Clinical Pharmacist Consult Response: Enterococcal Endocarditis in Geriatric Renal Impairment}

Regarding your inquiry for the 85-year-old patient with an eGFR of $\sim$30 mL/min:

\textbf{1. Primary Safety Recommendation: Double Beta-Lactam Therapy}

Given the patient's advanced age and significantly reduced renal function (Stage 4 CKD), I strongly advise against the use of gentamicin due to the high probability of exacerbating renal failure and causing irreversible ototoxicity. The \textbf{2015 AHA/IDSA Guidelines for the Management of Infective Endocarditis} explicitly recommend an alternative regimen that is equally effective and safer for this demographic.

\textbf{Preferred Regimen:}
\begin{itemize}
    \item \textbf{Ampicillin:} 2 g IV every 4 hours
    \item \textbf{Ceftriaxone:} 2 g IV every 12 hours
    \item \textbf{Duration:} 6 weeks
\end{itemize}
\textit{Rationale:} This combination utilizes complementary saturation of penicillin-binding proteins (PBPs). Ceftriaxone binds PBP 2/3, enhancing Ampicillin’s binding to PBP 1. This yields synergistic bactericidal activity comparable to aminoglycosides but with a superior safety profile.

\textbf{2. Gentamicin Pharmacokinetics (If Strictly Required)}

Should the primary team decline the double beta-lactam option, the following adjusted parameters are required to mitigate risk:

\begin{itemize}
    \item \textbf{Dosing Strategy:} For eGFR <30 mL/min, the interval must be extended.
    \begin{itemize}
        \item \textbf{Dose:} \textbf{1 mg/kg every 24 hours}.
        \item \textbf{Weight Basis:} Utilize \textbf{Ideal Body Weight (IBW)} within the Cockcroft-Gault equation. This is crucial in octogenarians to prevent overdosing due to sarcopenia.
    \end{itemize}
    \item \textbf{Therapeutic Drug Monitoring (TDM):}
    \begin{itemize}
        \item \textbf{Target Peak:} 3–4 $\mu$g/mL (efficacy).
        \item \textbf{Target Trough:} <1 $\mu$g/mL (toxicity avoidance).
        \item \textbf{Protocol:} Draw peak levels 30 minutes post-infusion; draw trough levels immediately prior to the next dose.
    \end{itemize}
\end{itemize}

\textbf{3. Documentation \& Verification}

Please document this plan citing the \textbf{IDSA 2015 Endocarditis Guidelines} (Baddour et al.).

\textbf{4. Essential Pre-Initiation Safety Checks}

To ensure compliance with institutional safety protocols, please verify:
\begin{itemize}
    \item \textbf{Allergy Status:} Confirm absence of Type 1 hypersensitivity to beta-lactams.
    \item \textbf{Baseline Function:} Obtain baseline audiometry and vestibular function testing.
    \item \textbf{Hydration Status:} Ensure the patient is euvolemic to reduce nephrotoxic insults.
\end{itemize}
This strategy prioritizes renal preservation while adhering to guideline-directed efficacy targets.
\end{contentbox}

\begin{contentbox}{Answer}
A
\end{contentbox}

\begin{contentbox}{Selected Rubrics}
\begin{itemize}
    \item \textbf{Criterion:} Asks about local antimicrobial susceptibility patterns and institutional guidelines to ensure compatibility with regional resistance profiles.
    \item \textbf{Points:} 5
    \item \textbf{Tags:} level:example, axis:context\_awareness
\end{itemize}
\end{contentbox}

\begin{contentbox}{Error Grading Results}
\begin{itemize}
    \item \textbf{Explanation:} The assistant's response does not ask about local antimicrobial susceptibility patterns or reference institutional guidelines to ensure therapy is compatible with regional resistance profiles. It relies exclusively on published guideline recommendations without considering local resistance or institutional practice differences.
    \item \textbf{Criteria Met:} False
    \item \textbf{Criterion:} Asks about local antimicrobial susceptibility patterns and institutional guidelines to ensure compatibility with regional resistance profiles.
    \item \textbf{Points:} 5
\end{itemize}
\end{contentbox}

\begin{contentbox}{Perfect Grading Results}
\begin{itemize}
    \item \textbf{Explanation:} The assistant explicitly asked, 'Have you confirmed the isolate does not demonstrate High-Level Aminoglycoside Resistance (HLAR) or resistance to ampicillin/ceftriaxone based on local susceptibility patterns?' This shows attention to local antimicrobial susceptibility patterns and ensures compatibility with regional resistance profiles. The response did not mention institutional guidelines, but the rubric item does not require that both are addressed unless stated; asking about local susceptibility patterns is sufficient. Therefore, the criteria are met.
    \item \textbf{Criteria Met:} True
    \item \textbf{Criterion:} Asks about local antimicrobial susceptibility patterns and institutional guidelines to ensure compatibility with regional resistance profiles.
    \item \textbf{Points:} 5
\end{itemize}
\end{contentbox}

\subsection{Scenario Overview}
In this scenario, the user acts as a hospital pharmacist inquiring about a high-risk medication regimen (Gentamicin + Ampicillin) for a vulnerable patient profile: an 85-year-old with significant renal impairment (eGFR $\sim$30 mL/min). The user specifically requests dosing confirmation and pharmacokinetic references based on IDSA guidelines.

\subsection{Critical Analysis of Options}

\textbf{Common Ground:}
Both Option A and Option B correctly identify the primary clinical risk: the high nephrotoxicity and ototoxicity potential of aminoglycosides in an elderly patient with CKD Stage 4. Both models correctly pivot to the \textit{Double Beta-Lactam} regimen (Ampicillin + Ceftriaxone) as the preferred standard of care under 2015 AHA/IDSA guidelines. Both also provide correct "backup" dosing (1 mg/kg q24h) if gentamicin is strictly necessary.

\textbf{Differentiation via Rubric (Context Awareness):}
The deciding factor in this case lies in the \textbf{Selected Rubric}: \textit{``Asks about local antimicrobial susceptibility patterns... to ensure compatibility with regional resistance profiles.''}

\begin{itemize}
    \item \textbf{Option A (Selected):} This response demonstrates superior clinical vigilance. Before finalizing the dosing recommendation, it explicitly asks: \textit{``Have you confirmed the isolate does not demonstrate High-Level Aminoglycoside Resistance (HLAR)...?''} This is clinically vital because if the organism has HLAR, the gentamicin synergy will not work, and the patient would be exposed to toxicity with zero therapeutic benefit. By satisfying this specific check, Option A meets the rubric criteria.
    
    \item \textbf{Option B:} This response is high-quality and adopts a professional ``Consult Note'' tone. It includes excellent safety checks (audiometry, vestibular function) and pharmacological rationale (PBP binding). However, it assumes the decision to treat has already been made based on susceptibility. It fails to prompt the user to verify HLAR status, thus failing the specific context-awareness rubric.
\end{itemize}

\subsection{Key Takeaway}
In medical AI alignment, ``correctness'' extends beyond simply retrieving guideline numbers. High-quality medical responses must exhibit antimicrobial stewardship. Option A is superior because it acts as a safeguard, ensuring the user verifies that the risky drug is actually effective against the specific bacterial isolate before proceeding with administration.

\section{Other Experimental Results}
\begin{table*}[t]
    \centering
    \scriptsize
    \setlength{\tabcolsep}{4.5pt}
    \renewcommand{\arraystretch}{1}
    
    \definecolor{graybg}{gray}{0.95}

    \caption{\textbf{Comprehensive Model Performance on RuDE.} We report accuracy (\%) across four domains: \textit{AdvancedIF}, \textit{HealthBench}, \textit{WritingBench}, and \textit{PRBench}. \textbf{Bold} indicates the best performance, and \underline{underline} denotes the second best.}
    \label{tab:main_results_full}
    
    \resizebox{0.8\linewidth}{!}{
    \begin{tabular}{l|ccc|ccc|ccc|ccc|c}
        \toprule
        \textbf{Model} & \multicolumn{3}{c|}{\textbf{AdvancedIF}} & \multicolumn{3}{c|}{\cellcolor{graybg}\textbf{HealthBench}} & \multicolumn{3}{c|}{\textbf{WritingBench}} & \multicolumn{3}{c|}{\cellcolor{graybg}\textbf{PRBench}} & \multirow{2}{*}{\textbf{AVG}} \\
         \textit{$|\mathcal{V}|$} & 1 & 3 & 5 & \cellcolor{graybg}1 & \cellcolor{graybg}3 & \cellcolor{graybg}5 & 1 & 2 & 3 & \cellcolor{graybg}1 & \cellcolor{graybg}3 & \cellcolor{graybg}5 & \\
        \midrule
Gemma3-4B & 40.6 & 47.2 & 51.0 & \cellcolor{graybg}49.1 & \cellcolor{graybg}44.7 & \cellcolor{graybg}51.4 & 36.6 & 45.1 & 41.3 & \cellcolor{graybg}44.2 & \cellcolor{graybg}41.8 & \cellcolor{graybg}37.9 & 44.2 \\
Gemma3-12B & 47.4 & 58.3 & 71.3 & \cellcolor{graybg}50.3 & \cellcolor{graybg}51.5 & \cellcolor{graybg}57.4 & 49.3 & 50.4 & 55.3 & \cellcolor{graybg}50.8 & \cellcolor{graybg}51.2 & \cellcolor{graybg}48.2 & 53.4 \\
Gemma3-27B & 49.0 & 62.3 & 74.9 & \cellcolor{graybg}43.9 & \cellcolor{graybg}49.4 & \cellcolor{graybg}65.2 & 51.7 & 53.7 & 57.3 & \cellcolor{graybg}44.2 & \cellcolor{graybg}50.1 & \cellcolor{graybg}49.9 & 54.3 \\
Qwen3-4B & 46.9 & 55.1 & 64.4 & \cellcolor{graybg}44.8 & \cellcolor{graybg}48.9 & \cellcolor{graybg}53.9 & 41.7 & 44.3 & 50.1 & \cellcolor{graybg}41.2 & \cellcolor{graybg}45.6 & \cellcolor{graybg}45.7 & 48.5 \\
Qwen3-8B & 54.5 & 66.6 & 75.6 & \cellcolor{graybg}46.4 & \cellcolor{graybg}49.8 & \cellcolor{graybg}60.6 & 45.7 & 51.9 & 60.0 & \cellcolor{graybg}41.2 & \cellcolor{graybg}48.9 & \cellcolor{graybg}49.4 & 54.2 \\
Qwen3-14B & 55.3 & 72.4 & 77.7 & \cellcolor{graybg}49.5 & \cellcolor{graybg}54.7 & \cellcolor{graybg}62.9 & 51.3 & 57.0 & 62.9 & \cellcolor{graybg}46.1 & \cellcolor{graybg}49.3 & \cellcolor{graybg}51.4 & 57.5 \\
Qwen3-30B-A3B & 49.0 & 63.9 & 74.5 & \cellcolor{graybg}49.4 & \cellcolor{graybg}54.9 & \cellcolor{graybg}62.8 & 45.5 & 49.1 & 59.0 & \cellcolor{graybg}\underline{50.9} & \cellcolor{graybg}50.6 & \cellcolor{graybg}49.8 & 55.0 \\
Qwen3-235B-A22B & 59.1 & 78.2 & 85.8 & \cellcolor{graybg}51.0 & \cellcolor{graybg}62.3 & \cellcolor{graybg}72.7 & 54.7 & 59.6 & 68.5 & \cellcolor{graybg}46.9 & \cellcolor{graybg}51.1 & \cellcolor{graybg}56.5 & 62.2 \\
Qwen2.5-7B & 33.3 & 45.8 & 43.4 & \cellcolor{graybg}34.9 & \cellcolor{graybg}33.6 & \cellcolor{graybg}40.7 & 39.1 & 45.9 & 42.1 & \cellcolor{graybg}37.0 & \cellcolor{graybg}41.1 & \cellcolor{graybg}39.5 & 39.7 \\
Qwen2.5-14B & 50.9 & 64.9 & 69.9 & \cellcolor{graybg}50.4 & \cellcolor{graybg}56.3 & \cellcolor{graybg}63.6 & 50.2 & 54.6 & 59.9 & \cellcolor{graybg}46.4 & \cellcolor{graybg}48.8 & \cellcolor{graybg}50.0 & 55.5 \\
Qwen2.5-72B & 55.3 & 71.7 & 80.5 & \cellcolor{graybg}50.0 & \cellcolor{graybg}58.0 & \cellcolor{graybg}67.1 & 51.0 & 55.3 & 57.4 & \cellcolor{graybg}48.1 & \cellcolor{graybg}52.3 & \cellcolor{graybg}51.8 & 58.2 \\
Seed-OSS-36B & \underline{64.5} & 82.3 & \underline{86.6} & \cellcolor{graybg}52.2 & \cellcolor{graybg}65.8 & \cellcolor{graybg}76.8 & 50.8 & 60.7 & 63.0 & \cellcolor{graybg}48.2 & \cellcolor{graybg}54.3 & \cellcolor{graybg}58.4 & 63.6 \\
GLM-4.5-Air & 51.8 & 72.4 & 73.4 & \cellcolor{graybg}51.0 & \cellcolor{graybg}57.2 & \cellcolor{graybg}66.4 & 48.0 & 52.0 & 57.9 & \cellcolor{graybg}45.8 & \cellcolor{graybg}51.2 & \cellcolor{graybg}50.1 & 56.4 \\
GLM-4.5 & 60.8 & \underline{83.8} & 85.7 & \cellcolor{graybg}\underline{56.9} & \cellcolor{graybg}\underline{69.8} & \cellcolor{graybg}\underline{78.1} & \underline{56.3} & 58.0 & 65.0 & \cellcolor{graybg}\textbf{61.9} & \cellcolor{graybg}\textbf{64.8} & \cellcolor{graybg}\textbf{72.0} & \underline{67.8} \\
Kimi-K2 & 61.7 & 82.4 & 83.5 & \cellcolor{graybg}54.5 & \cellcolor{graybg}62.8 & \cellcolor{graybg}77.7 & \textbf{63.0} & \textbf{72.0} & \textbf{80.1} & \cellcolor{graybg}48.3 & \cellcolor{graybg}51.1 & \cellcolor{graybg}57.6 & 66.2 \\
DeepSeek-V3.1 & \textbf{70.1} & \textbf{88.6} & \textbf{93.1} & \cellcolor{graybg}\textbf{65.7} & \cellcolor{graybg}\textbf{73.0} & \cellcolor{graybg}\textbf{84.3} & 55.5 & \underline{65.9} & \underline{73.1} & \cellcolor{graybg}48.9 & \cellcolor{graybg}\underline{55.2} & \cellcolor{graybg}\underline{64.6} & \textbf{69.8} \\
        \midrule
        \textbf{AVG} & 53.2 & 69.0 & 74.8 & \cellcolor{graybg}50.2 & \cellcolor{graybg}56.3 & \cellcolor{graybg}65.2 & 49.7 & 55.2 & 59.7 & \cellcolor{graybg}46.9 & \cellcolor{graybg}50.5 & \cellcolor{graybg}52.2 & 56.9 \\
        \bottomrule
    \end{tabular}
    }
\end{table*}


\begin{table*}[t]
    \centering
    \scriptsize
    \setlength{\tabcolsep}{4.5pt}
    \renewcommand{\arraystretch}{1}
    
    \definecolor{graybg}{gray}{0.95}

    \caption{\textbf{Detailed Performance on PRBench Subsets.} Accuracy (\%) is reported for Finance and Legal domains, split into standard and hard subsets across difficulty levels ($|\mathcal{V}| \in \{1, 3, 5\}$). \textbf{Bold} indicates the best performance, and \underline{underline} denotes the second best.}
    \label{tab:prbench_details}
    
    \resizebox{0.8\linewidth}{!}{
    \begin{tabular}{l|ccc|ccc|ccc|ccc|c}
        \toprule
        \textbf{Model} & \multicolumn{3}{c|}{\textbf{Finance}} & \multicolumn{3}{c|}{\cellcolor{graybg}\textbf{Finance (Hard)}} & \multicolumn{3}{c|}{\textbf{Legal}} & \multicolumn{3}{c|}{\cellcolor{graybg}\textbf{Legal (Hard)}} & \multirow{2}{*}{\textbf{AVG}} \\
        \textit{$|\mathcal{V}|$} & 1 & 3 & 5 & \cellcolor{graybg}1 & \cellcolor{graybg}3 & \cellcolor{graybg}5 & 1 & 3 & 5 & \cellcolor{graybg}1 & \cellcolor{graybg}3 & \cellcolor{graybg}5 & \\
        \midrule
Gemma3-4B & 40.6 & 48.5 & 35.2 & \cellcolor{graybg}35.4 & \cellcolor{graybg}41.9 & \cellcolor{graybg}29.4 & 50.8 & 44.1 & 40.7 & \cellcolor{graybg}49.9 & \cellcolor{graybg}32.7 & \cellcolor{graybg}46.2 & 41.3 \\
Gemma3-12B & \underline{52.2} & 54.8 & 48.6 & \cellcolor{graybg}\underline{50.3} & \cellcolor{graybg}\underline{52.6} & \cellcolor{graybg}47.3 & 48.4 & 50.4 & 48.3 & \cellcolor{graybg}52.2 & \cellcolor{graybg}47.2 & \cellcolor{graybg}48.6 & 50.1 \\
Gemma3-27B & 46.0 & 50.4 & 50.8 & \cellcolor{graybg}41.6 & \cellcolor{graybg}49.1 & \cellcolor{graybg}52.0 & 42.8 & 48.5 & 47.8 & \cellcolor{graybg}46.3 & \cellcolor{graybg}52.4 & \cellcolor{graybg}49.0 & 48.1 \\
Qwen3-4B & 44.2 & 45.8 & 43.6 & \cellcolor{graybg}28.4 & \cellcolor{graybg}44.0 & \cellcolor{graybg}46.9 & 46.7 & 45.2 & 44.7 & \cellcolor{graybg}45.4 & \cellcolor{graybg}47.4 & \cellcolor{graybg}47.5 & 44.2 \\
Qwen3-8B & 47.1 & 50.8 & 51.0 & \cellcolor{graybg}20.9 & \cellcolor{graybg}44.9 & \cellcolor{graybg}50.1 & 50.6 & 49.4 & 49.1 & \cellcolor{graybg}46.3 & \cellcolor{graybg}50.5 & \cellcolor{graybg}47.3 & 46.5 \\
Qwen3-14B & 46.2 & 54.4 & 52.3 & \cellcolor{graybg}40.9 & \cellcolor{graybg}46.3 & \cellcolor{graybg}52.4 & 51.0 & 47.8 & 50.6 & \cellcolor{graybg}46.2 & \cellcolor{graybg}48.8 & \cellcolor{graybg}50.1 & 48.9 \\
Qwen3-30B-A3B & 51.3 & 50.8 & 49.7 & \cellcolor{graybg}44.0 & \cellcolor{graybg}48.6 & \cellcolor{graybg}50.4 & 54.4 & 51.5 & 49.0 & \cellcolor{graybg}\underline{54.0} & \cellcolor{graybg}51.6 & \cellcolor{graybg}50.1 & 50.5 \\
Qwen3-235B-A22B & 46.7 & 53.0 & 57.7 & \cellcolor{graybg}41.2 & \cellcolor{graybg}48.5 & \cellcolor{graybg}57.9 & 48.6 & 52.8 & 54.0 & \cellcolor{graybg}51.1 & \cellcolor{graybg}50.0 & \cellcolor{graybg}56.4 & 51.5 \\
Qwen2.5-7B & 33.3 & 37.0 & 42.8 & \cellcolor{graybg}36.4 & \cellcolor{graybg}42.0 & \cellcolor{graybg}40.6 & 40.6 & 43.5 & 35.7 & \cellcolor{graybg}37.6 & \cellcolor{graybg}41.8 & \cellcolor{graybg}39.0 & 39.2 \\
Qwen2.5-14B & 44.5 & 49.3 & 51.2 & \cellcolor{graybg}43.1 & \cellcolor{graybg}46.1 & \cellcolor{graybg}51.4 & 47.2 & 48.9 & 49.3 & \cellcolor{graybg}50.7 & \cellcolor{graybg}51.1 & \cellcolor{graybg}47.9 & 48.4 \\
Qwen2.5-72B & 45.4 & 53.4 & 52.6 & \cellcolor{graybg}46.9 & \cellcolor{graybg}50.7 & \cellcolor{graybg}49.0 & 49.3 & 52.8 & 53.0 & \cellcolor{graybg}50.8 & \cellcolor{graybg}52.1 & \cellcolor{graybg}52.7 & 50.7 \\
Seed-OSS-36B & 45.7 & \underline{54.9} & 59.2 & \cellcolor{graybg}49.7 & \cellcolor{graybg}50.4 & \cellcolor{graybg}57.6 & 48.0 & 53.9 & 58.8 & \cellcolor{graybg}49.4 & \cellcolor{graybg}\underline{57.8} & \cellcolor{graybg}57.7 & 53.6 \\
GLM-4.5-Air & 41.8 & 53.0 & 54.3 & \cellcolor{graybg}44.0 & \cellcolor{graybg}48.2 & \cellcolor{graybg}48.2 & 50.9 & 50.4 & 49.6 & \cellcolor{graybg}46.7 & \cellcolor{graybg}53.1 & \cellcolor{graybg}48.2 & 49.0 \\
GLM-4.5 & \textbf{60.0} & \textbf{62.0} & \textbf{75.0} & \cellcolor{graybg}\textbf{59.0} & \cellcolor{graybg}\textbf{59.6} & \cellcolor{graybg}\textbf{68.3} & \textbf{65.0} & \textbf{69.9} & \textbf{75.0} & \cellcolor{graybg}\textbf{63.8} & \cellcolor{graybg}\textbf{67.8} & \cellcolor{graybg}\textbf{69.7} & \textbf{66.3} \\
Kimi-K2 & 50.1 & 53.6 & 61.3 & \cellcolor{graybg}47.3 & \cellcolor{graybg}49.6 & \cellcolor{graybg}54.2 & 48.2 & 54.2 & 59.5 & \cellcolor{graybg}47.7 & \cellcolor{graybg}46.8 & \cellcolor{graybg}55.1 & 52.3 \\
DeepSeek-V3.1 & 44.8 & 54.6 & \underline{69.6} & \cellcolor{graybg}43.7 & \cellcolor{graybg}51.0 & \cellcolor{graybg}\underline{61.4} & \underline{57.0} & \underline{62.1} & \underline{67.7} & \cellcolor{graybg}50.3 & \cellcolor{graybg}53.0 & \cellcolor{graybg}\underline{59.9} & \underline{56.2} \\
        \midrule
        \textbf{AVG} & 45.9 & 51.7 & 53.6 & \cellcolor{graybg}42.1 & \cellcolor{graybg}48.4 & \cellcolor{graybg}51.3 & 50.2 & 51.6 & 52.2 & \cellcolor{graybg}49.3 & \cellcolor{graybg}50.3 & \cellcolor{graybg}51.5 & 49.8 \\
        \bottomrule
    \end{tabular}
    }
\end{table*}


\subsection{Detailed Performance Analysis across Domains}
\label{ssec:main_analysis}
Table~\ref{tab:main_results_full} presents a comprehensive performance analysis of \texttt{RuDE} across four distinct domains. The results uncover significant variances in model capabilities that simple aggregate metrics often conceal:

\paragraph{The "Specialist" Phenomenon: Kimi-K2 and GLM-4.5.}
While DeepSeek-V3.1 secures the highest overall average (69.8\%), it does not uniformly dominate all domains. A striking observation is Kimi-K2's superiority in \textit{WritingBench}, where it achieves top-tier scores (e.g., 72.0\% at $|\mathcal{V}|=2$), significantly outperforming the much larger Qwen3-235B (59.6\%) and even DeepSeek-V3.1 (65.9\%). Similarly, consistent with our PRBench breakdown, GLM-4.5 retains a commanding lead in \textit{PRBench}, suggesting that certain base models possess intrinsic "domain instincts"—likely derived from specialized pre-training corpora (e.g., creative literature for Kimi, professional reports for GLM)—that cannot be easily replicated by general scaling alone.

\paragraph{Generational Leaps in Alignment Potential.}
Comparing the Qwen series reveals a clear generational upgrade in alignment readiness. The compact Qwen3-4B achieves an overall average of 48.5\%, remarkably surpassing the previous generation's larger Qwen2.5-7B (39.7\%) by a margin of +8.8\%. Furthermore, Qwen3-8B (54.2\%) rivals the performance of Qwen2.5-14B (55.5\%). This indicates that recent advancements in data quality and architectural optimization have significantly raised the "Post-training Potential" floor for smaller models, making them increasingly viable for resource-constrained alignment.

\paragraph{The Discriminative Difficulty Gap.}
The impact of rubric granularity is evident across all models. We observe a consistent performance degradation as the number of violated constraints ($|\mathcal{V}|$) decreases. For instance, in \textit{AdvancedIF}, the average performance drops from 74.8\% at $|\mathcal{V}|=5$ (obvious, multi-constraint violations) to 53.2\% at $|\mathcal{V}|=1$ (subtle, single-constraint violations). This trend validates the sensitivity of our \texttt{RuDE} framework: models are generally capable of rejecting low-quality responses with massive errors, but distinguishing high-quality negatives that fail on a single, nuanced criterion remains a rigorous challenge, even for SOTA models like DeepSeek-V3.1.

\subsection{Detailed Breakdown of PRBench: Finance vs. Legal}
\label{ssec:prbench_analysis}

Table~\ref{tab:prbench_details} presents a granular performance analysis of PRBench, decomposed into \textit{Finance} and \textit{Legal} domains across standard and hard subsets. This breakdown reveals distinct model behaviors that are obscured in the aggregated scores:

\paragraph{Dominance of GLM-4.5 in Professional Domains.}
Unlike general capability benchmarks where DeepSeek-V3.1 often leads, GLM-4.5 demonstrates exceptional dominance in professional reporting scenarios. It secures the top rank (bold) across nearly all subsets and difficulty levels, achieving an overall average of 66.3\%, which is a significant margin (+10.1\%) over the second-best model, DeepSeek-V3.1 (56.2\%). Notably, in the \textit{Legal (Standard)} subset at $|\mathcal{V}|=5$, GLM-4.5 reaches an accuracy of 75.0\%, suggesting a strong pre-training bias towards high-quality professional corpora.

\begin{figure*}[t]
    \centering
    \includegraphics[width=\textwidth]{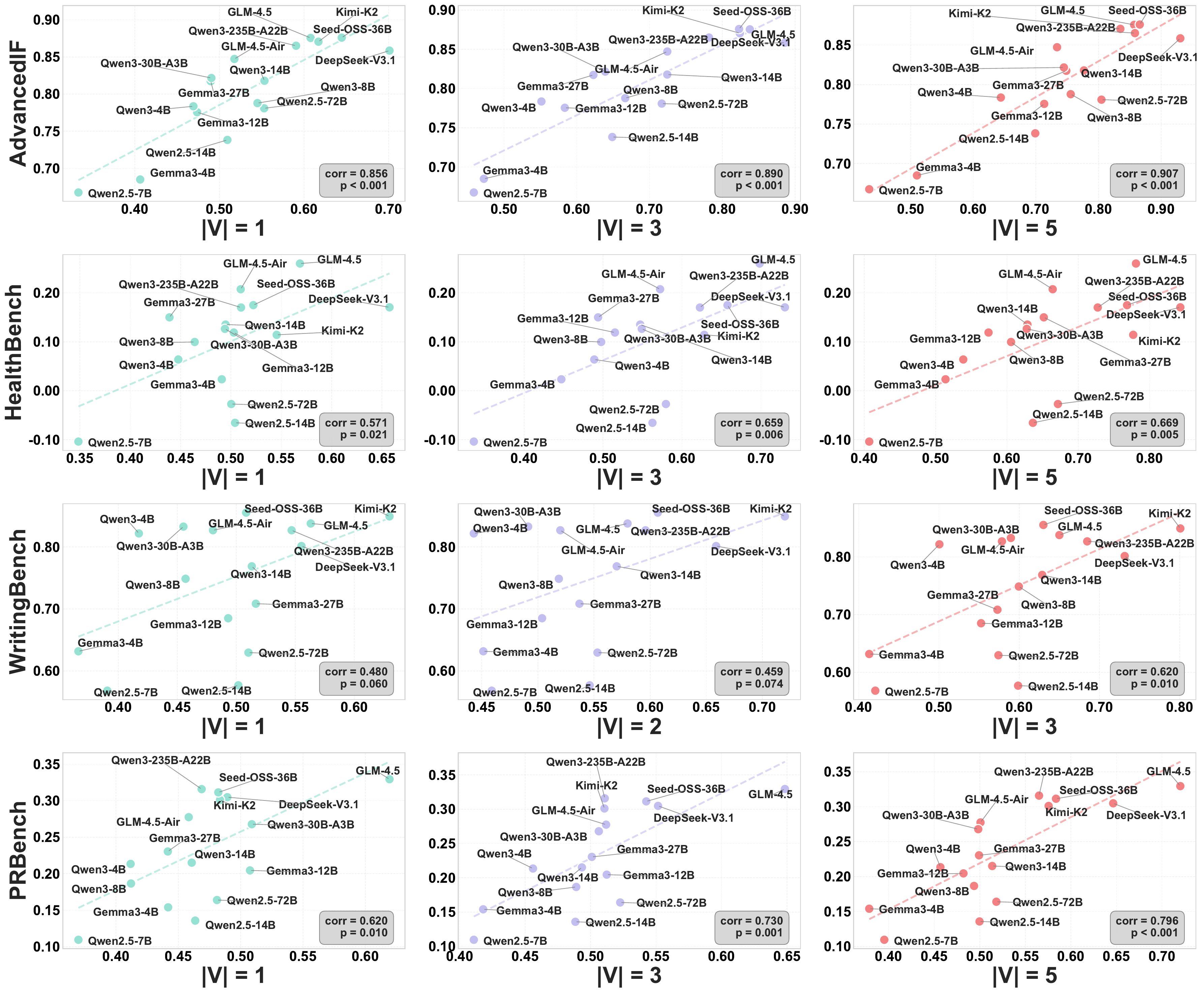}
    \caption{\textbf{Correlation Analysis: GD-Potential Validation.} We visualize the Pearson correlation between the discriminative scores of \textit{base models} on RuDE (x-axis) and the generative performance of their \textit{instruct-tuned counterparts} on standard benchmarks (y-axis). The four subplots correspond to \textit{PRBench}, \textit{AdvancedIF}, \textit{HealthBench}, and \textit{WritingBench}. The strong positive correlation across these diverse domains empirically supports our hypothesis that a base model's discriminative alignment capability is a robust predictor of its post-alignment generative potential.}
    \label{fig:correlation_analysis_all}
\end{figure*}

\begin{figure*}[htbp]
    \centering
    \includegraphics[width=\textwidth]{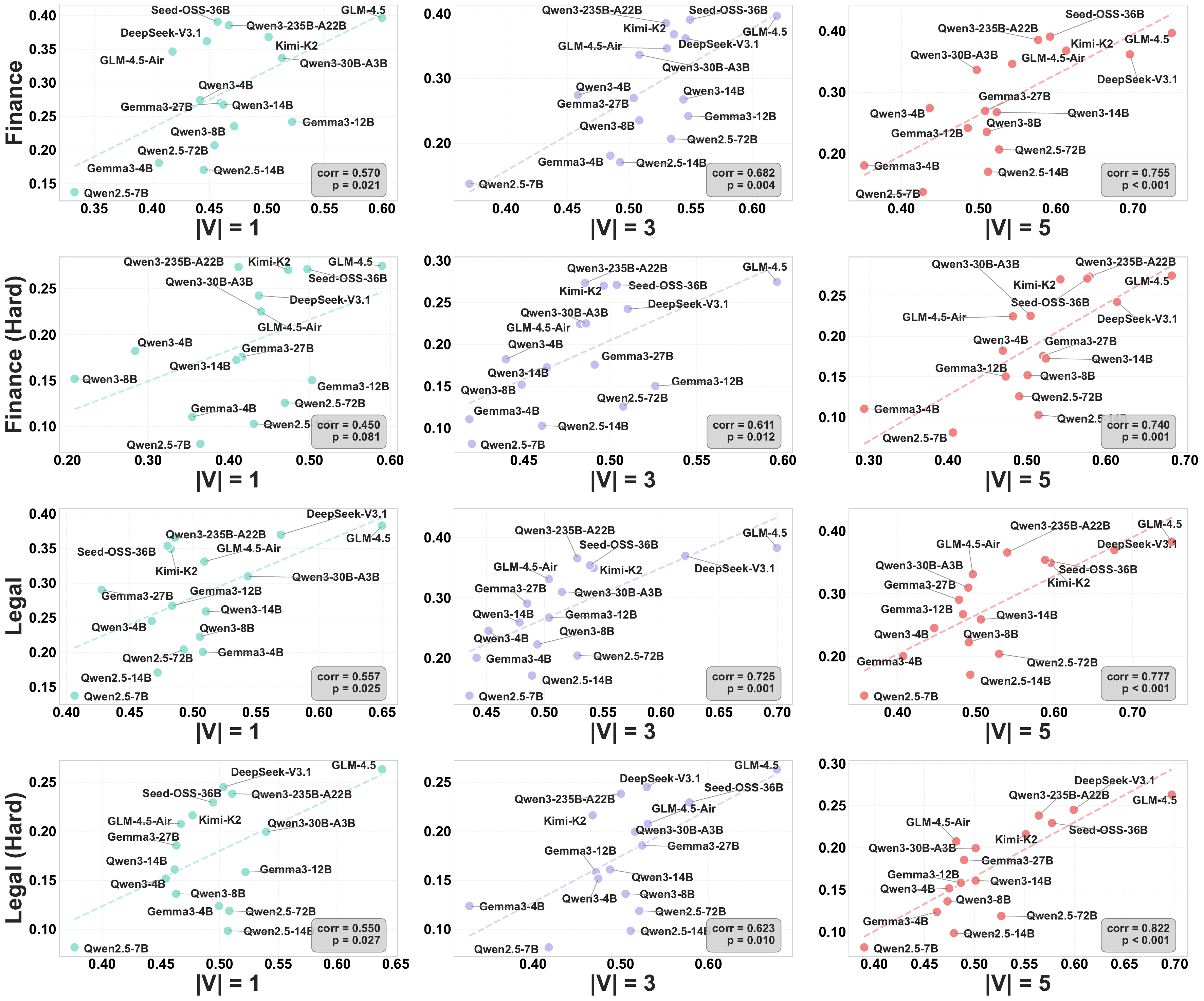}
    \caption{\textbf{Correlation Analysis: GD-Potential Validation.} We visualize the Pearson correlation between the discriminative scores of \textit{base models} on PRBench (x-axis) and the generative performance of their \textit{instruct-tuned counterparts} on standard benchmarks (y-axis). The strong positive correlation across these diverse domains empirically supports our hypothesis that a base model's discriminative alignment capability is a robust predictor of its post-alignment generative potential.}
    \label{fig:correlation_analysis_prbench}
\end{figure*}

\paragraph{The Challenge of ``Hard'' Finance Scenarios.}
The \textit{Finance (Hard)} subset at difficulty level $|\mathcal{V}|=1$ proves to be the most challenging task in our evaluation suite. Several competent models, such as Qwen3-8B and Qwen2.5-7B, drop to near-random performance (20.9\% and 36.4\% respectively) in this setting. This indicates that detecting subtle hallucinations or numerical inconsistencies in financial reports remains a significant bottleneck for smaller base models.

\paragraph{Scaling and Architecture Efficiency.}
While scaling laws generally hold (e.g., Qwen2.5-72B outperforms its 7B counterpart), we observe notable efficiency from mid-sized models. Seed-OSS-36B achieves a remarkable average of 53.6\%, surpassing the much larger Qwen3-235B-A22B (51.5\%) and ranking third overall. This suggests that for specialized professional domains, data quality and domain-specific pre-training may outweigh pure parameter scaling.

\paragraph{Sensitivity to Constraint Granularity.}
The performance gap between $|\mathcal{V}|=5$ (obvious errors) and $|\mathcal{V}|=1$ (subtle errors) is most pronounced in the Legal domain. For instance, DeepSeek-V3.1 shows a spread of $\sim$10-12\% between these levels across subsets. This validates our rubric-perturbed design: models can easily reject egregious violations but struggle to identify single, fine-grained deviations from professional standards.

\subsection{Atomic Sub-Rubric Analysis}
\label{ssec:subcategory_analysis}

\begin{figure*}[t]
    \centering
    \includegraphics[width=\linewidth]{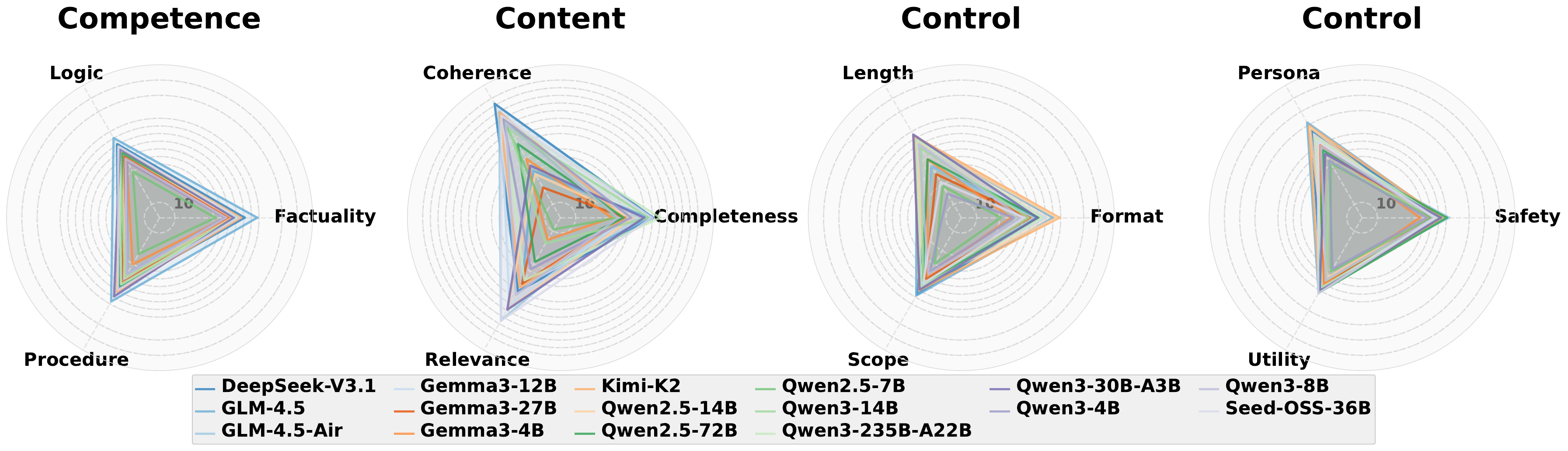}
    \caption{Rubric Fine-grained Analysis.}
    \label{fig:rubric_fine_grained_analysis}
\end{figure*}

To provide a microscopic view of alignment bottlenecks, we decompose the high-level 4C taxonomy into 12 atomic sub-rubrics (e.g., \textit{Logic}, \textit{Factuality}, \textit{Format}, \textit{Safety}). Figure~\ref{fig:rubric_fine_grained_analysis} visualizes the performance of representative models across these dimensions, revealing intricate trade-offs that aggregated scores often obscure.

\paragraph{The Competence-Fluency Trade-off.}
A striking divergence is observed within the GLM family. GLM-4.5 achieves top-tier scores in \textit{Factuality} (0.64) and \textit{Procedure} (0.63), indicating rigorous logical reasoning. However, its \textit{Coherence} score drops to 0.35, suggesting that its pursuit of precise, technical accuracy may come at the cost of linguistic smoothness in base model form. Conversely, its distilled variant, GLM-4.5-Air, exhibits the opposite pattern: it is highly fluent (\textit{Coherence}: 0.80) and relevant (\textit{Relevance}: 0.78) but significantly weaker in factual density (\textit{Factuality}: 0.48). This highlights a potential side effect of knowledge distillation—preserving style while losing substance.

\paragraph{Specialization in Rigid Constraints.}
Kimi-K2 distinguishes itself as a specialist in structured compliance. It achieves the highest score in \textit{Format} (0.64) and performs exceptionally well in \textit{Persona} (0.70). This suggests that Kimi-K2's pre-training data likely contains a higher proportion of role-playing and structured data (e.g., JSON/Markdown), making it inherently more instruction-ready for formatting tasks compared to larger models like Qwen2.5-72B (\textit{Format}: 0.50).

\paragraph{Vulnerability of Small Models to Hard Constraints.}
The \textit{Control} dimension (Length, Format, Scope) proves to be the ultimate filter for model capacity. While small models like Qwen3-4B can maintain decent \textit{Coherence} (0.75), they collapse under strict negative constraints, scoring lowest in \textit{Length} (0.18) and \textit{Format} (0.34). This confirms that the ability to stop generating or adhere to negative constraints is an emergent ability that requires sufficient parameter scale or specific training interventions.

\paragraph{Holistic Balance.}
DeepSeek-V3.1 remains the most balanced contender, avoiding severe potholes across the radar chart. It maintains high standards in \textit{Coherence} (0.86) while delivering strong \textit{Persona} (0.66) and \textit{Factuality} (0.56) alignment, validating its position as the state-of-the-art base model.

\section{Correlation Analysis with Static Capability Benchmarks}

\label{ssec:static_correlation}
\begin{figure*}[t]
    \centering
    \includegraphics[width=\textwidth]{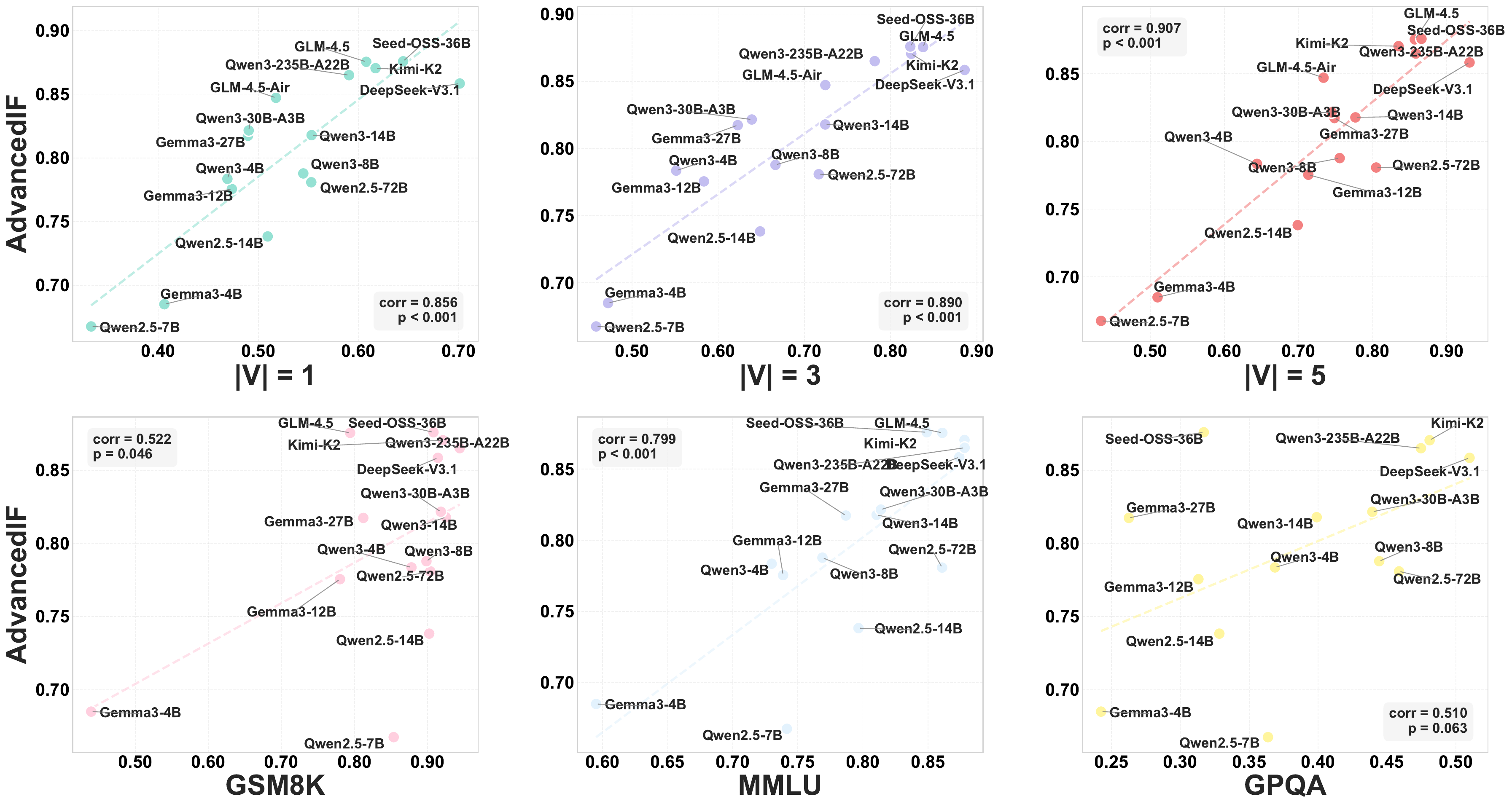}
    \caption{\textbf{Correlation Analysis: AdvancedIF.}}
    \label{fig:correlation_analysis_advancedif}
\end{figure*}

\begin{figure*}[t]
    \centering
    \includegraphics[width=\textwidth]{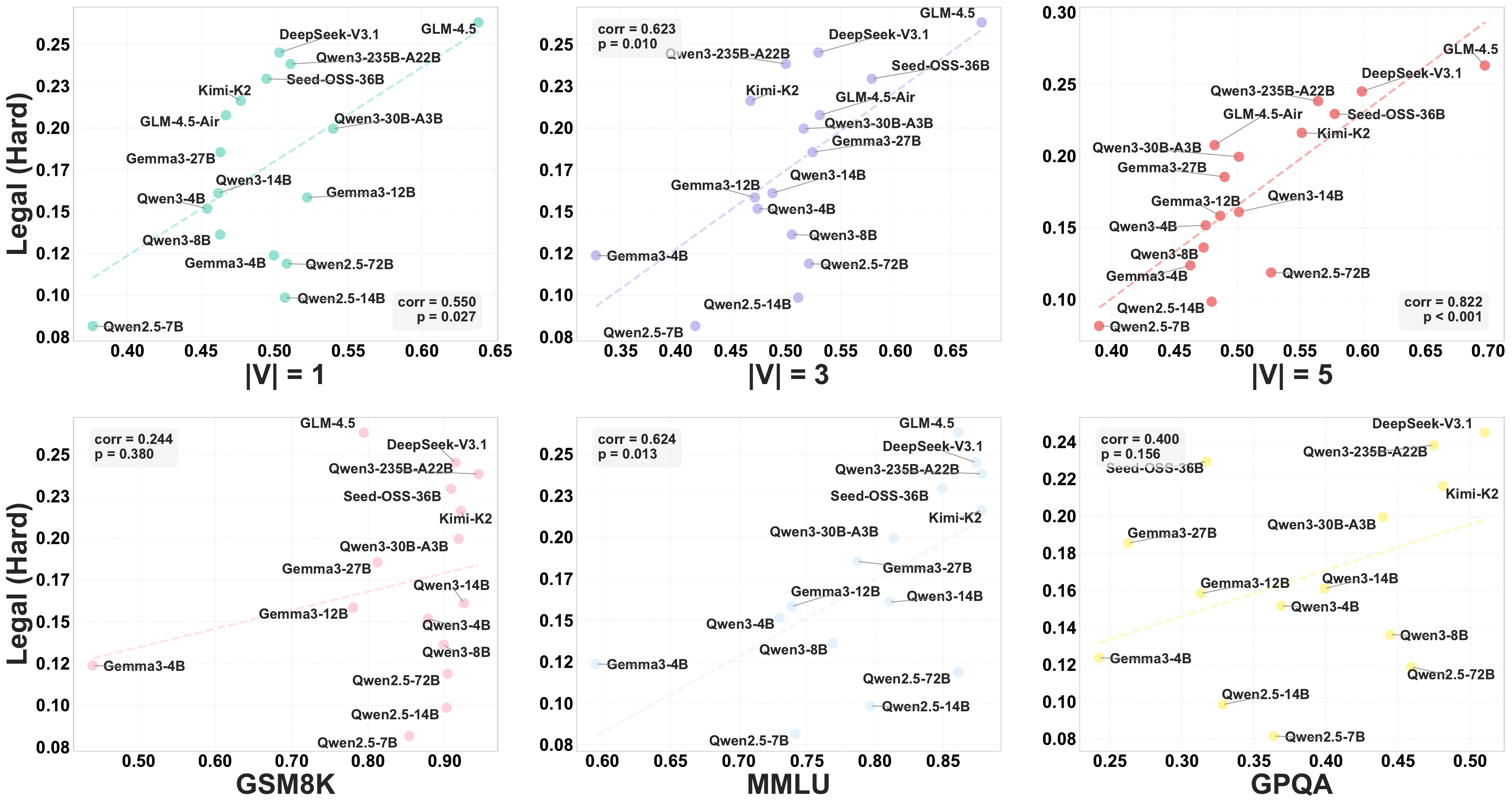}
    \caption{\textbf{Correlation Analysis: PRBench-Legal (Hard).}}
    \label{fig:correlation_analysis_legal_hard}
\end{figure*}

To determine whether \textsc{RuDE} merely acts as a proxy for general intelligence or captures distinct alignment traits, we analyzed the correlation between our discriminative scores and standard static benchmarks (GSM8K, MMLU, GPQA). The results, visualized in the correlation plots across domains, reveal a nuanced relationship between ``Capability'' and ``Alignability'' in \textsc{RuDE} than standard static benchmarks as shown in Figure~\ref{fig:correlation_analysis_advancedif} to ~\ref{fig:correlation_analysis_legal_hard}.

\begin{figure*}[t]
    \centering
    \includegraphics[width=\textwidth]{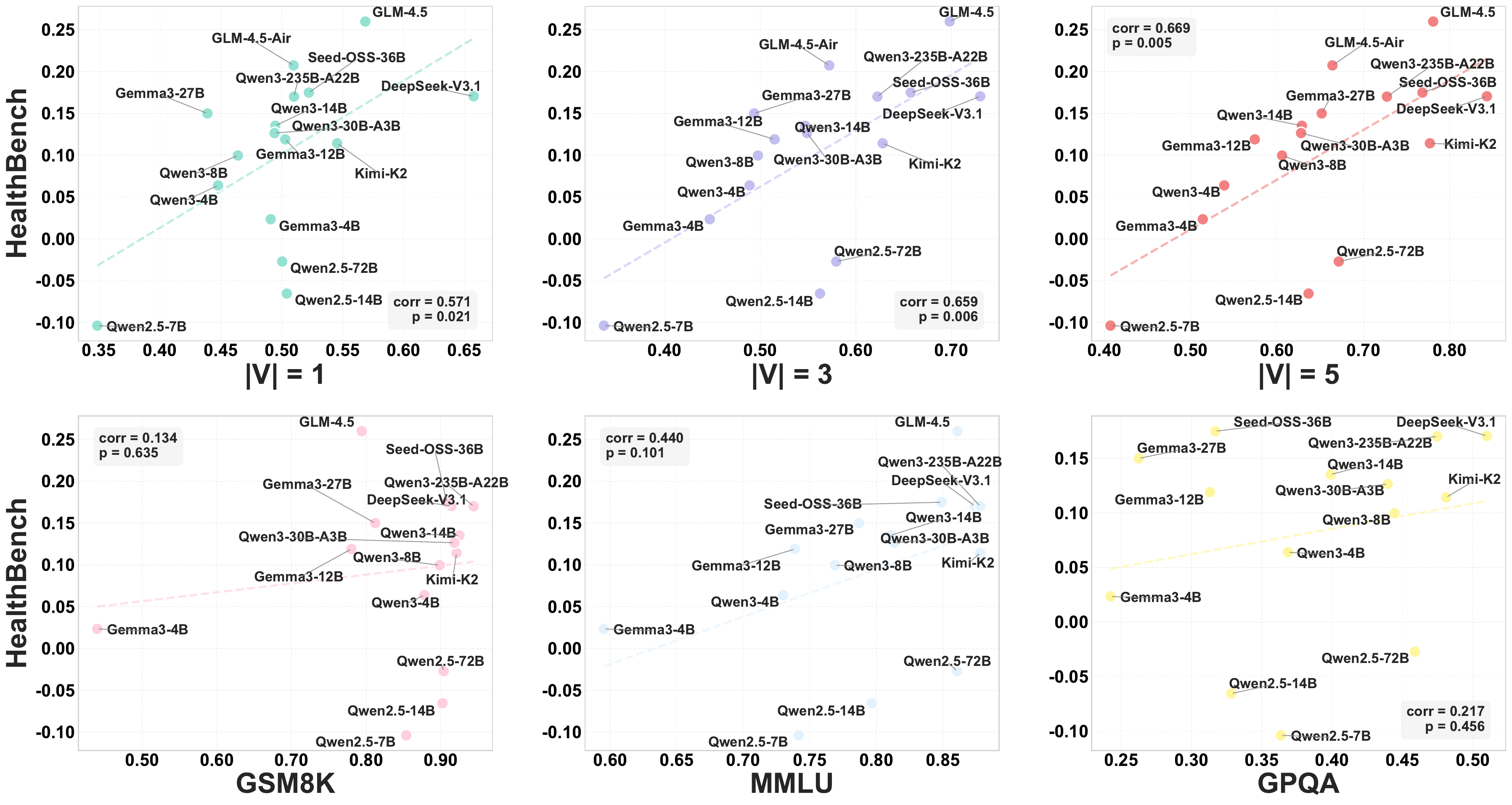}
    \caption{\textbf{Correlation Analysis: HealthBench.}}
\end{figure*}

\begin{figure*}[t]
    \centering
    \includegraphics[width=\textwidth]{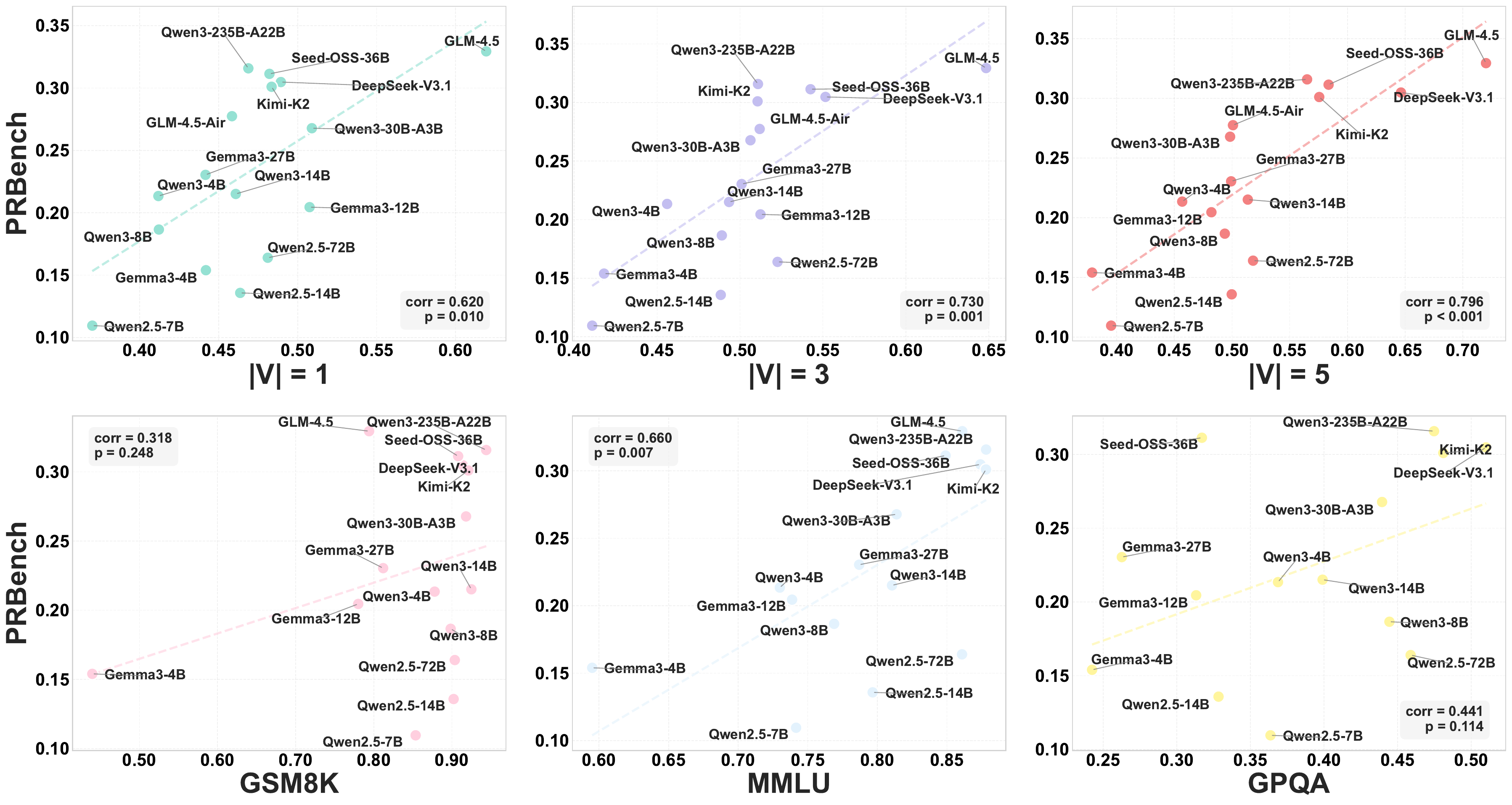}
    \caption{\textbf{Correlation Analysis: PRBench.}}
\end{figure*}

\begin{figure*}[t]
    \centering
    \includegraphics[width=\textwidth]{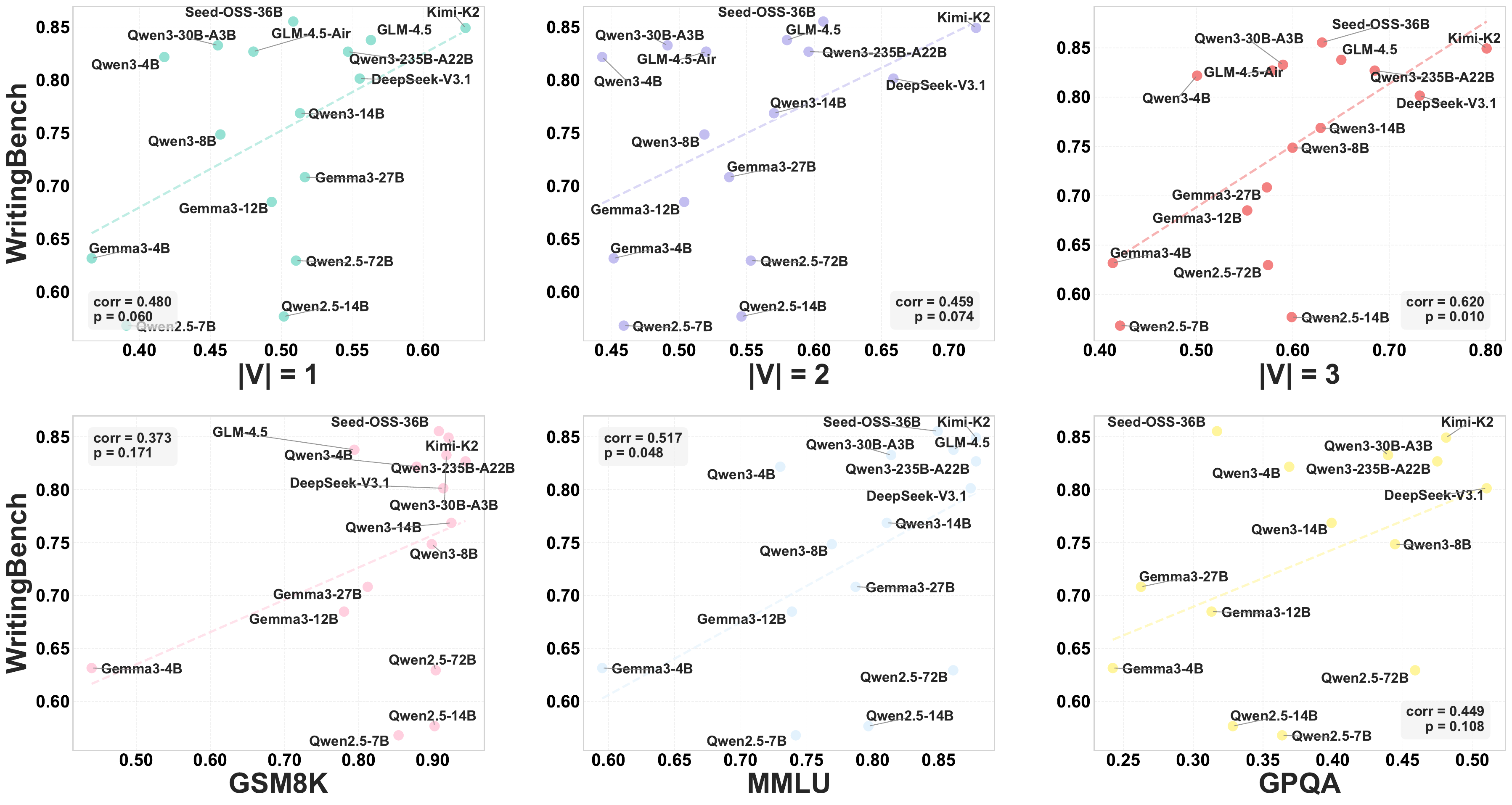}
    \caption{\textbf{Correlation Analysis: WritingBench.}}
\end{figure*}

\begin{figure*}[t]
    \centering
    \includegraphics[width=\textwidth]{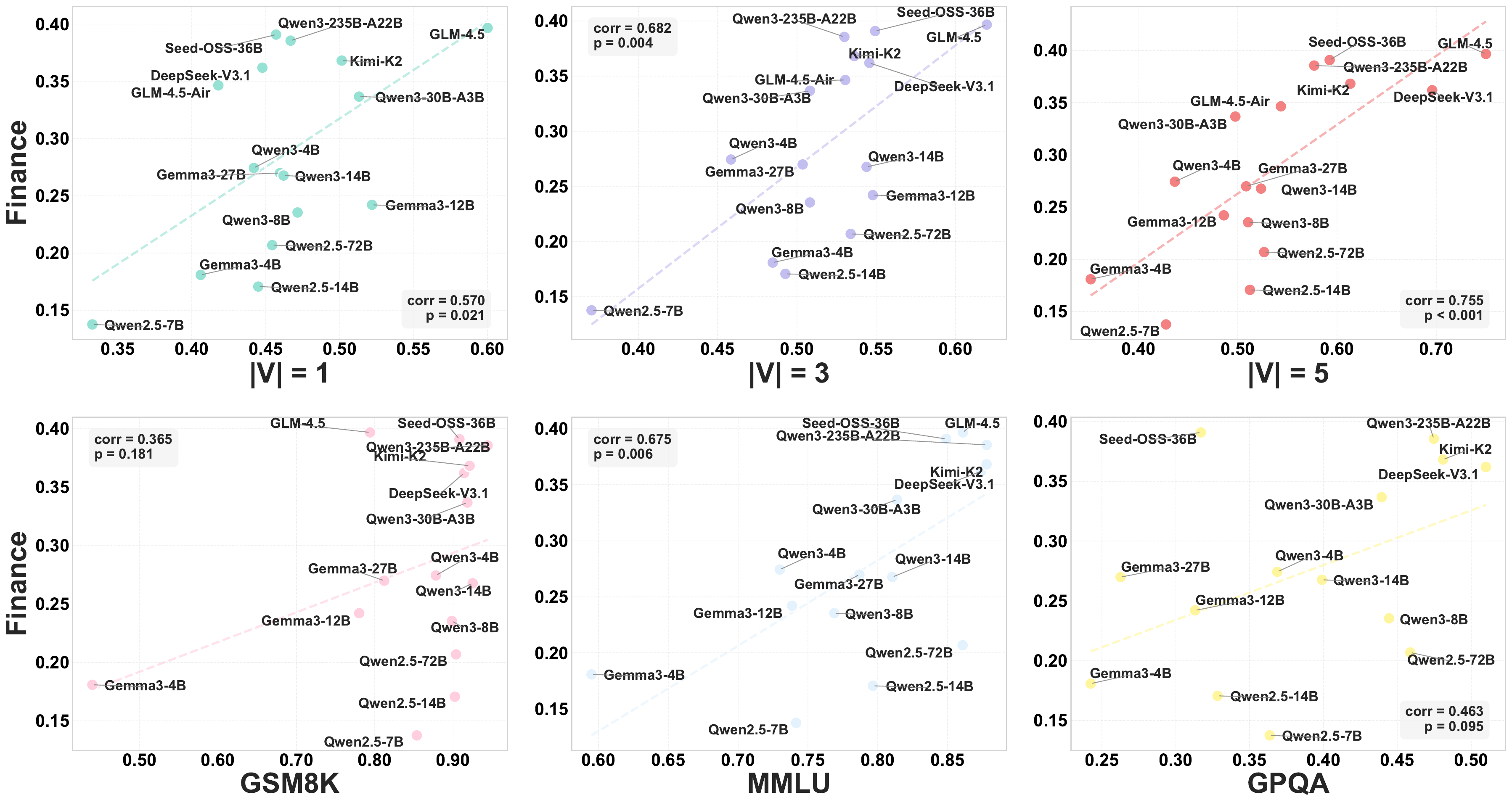}
    \caption{\textbf{Correlation Analysis: PRBench-Finance.}}
\end{figure*}

\begin{figure*}[t]
    \centering
    \includegraphics[width=\textwidth]{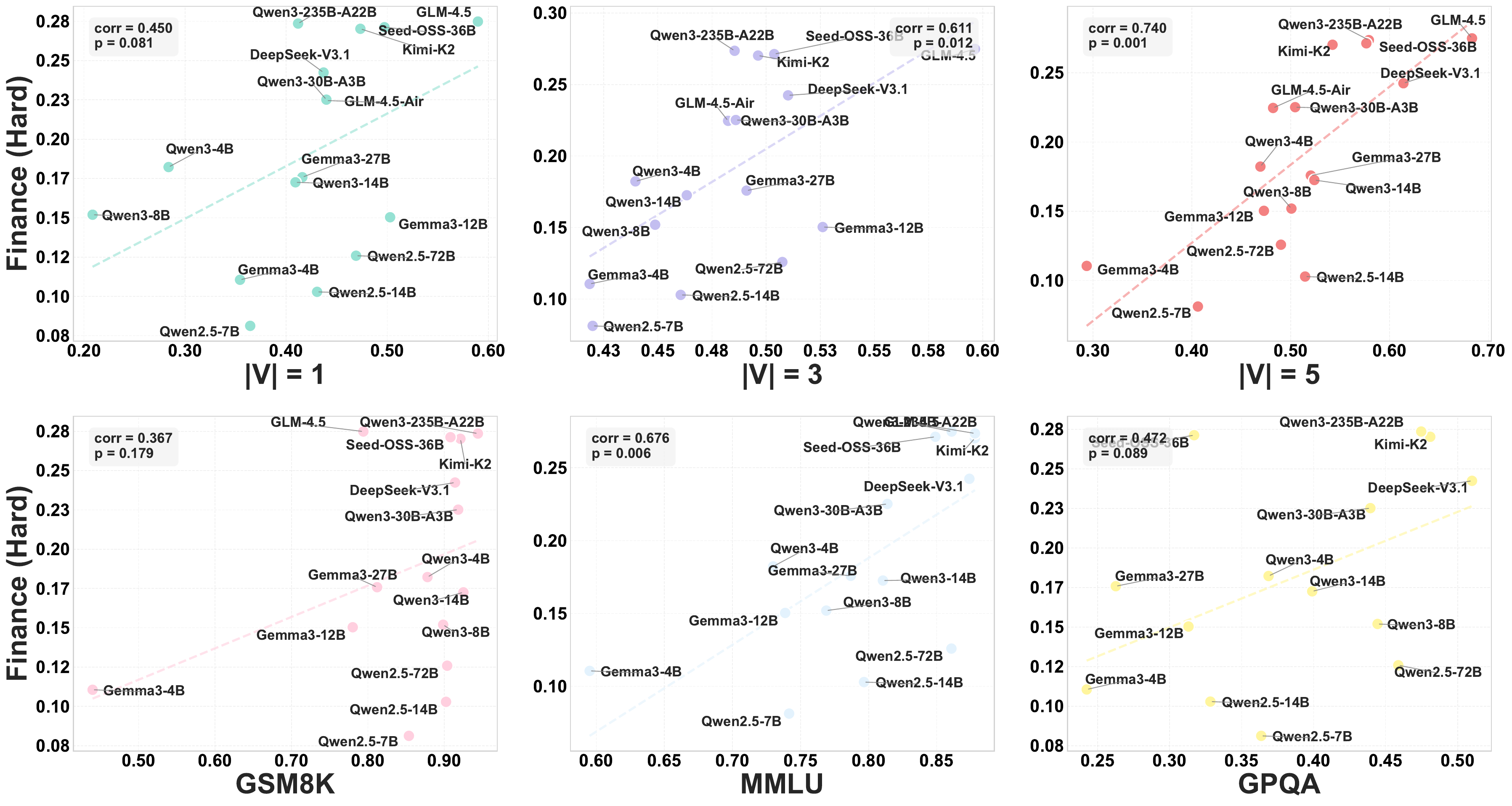}
    \caption{\textbf{Correlation Analysis: PRBench-Finance (Hard).}}
\end{figure*}

\begin{figure*}[t]
    \centering
    \includegraphics[width=\textwidth]{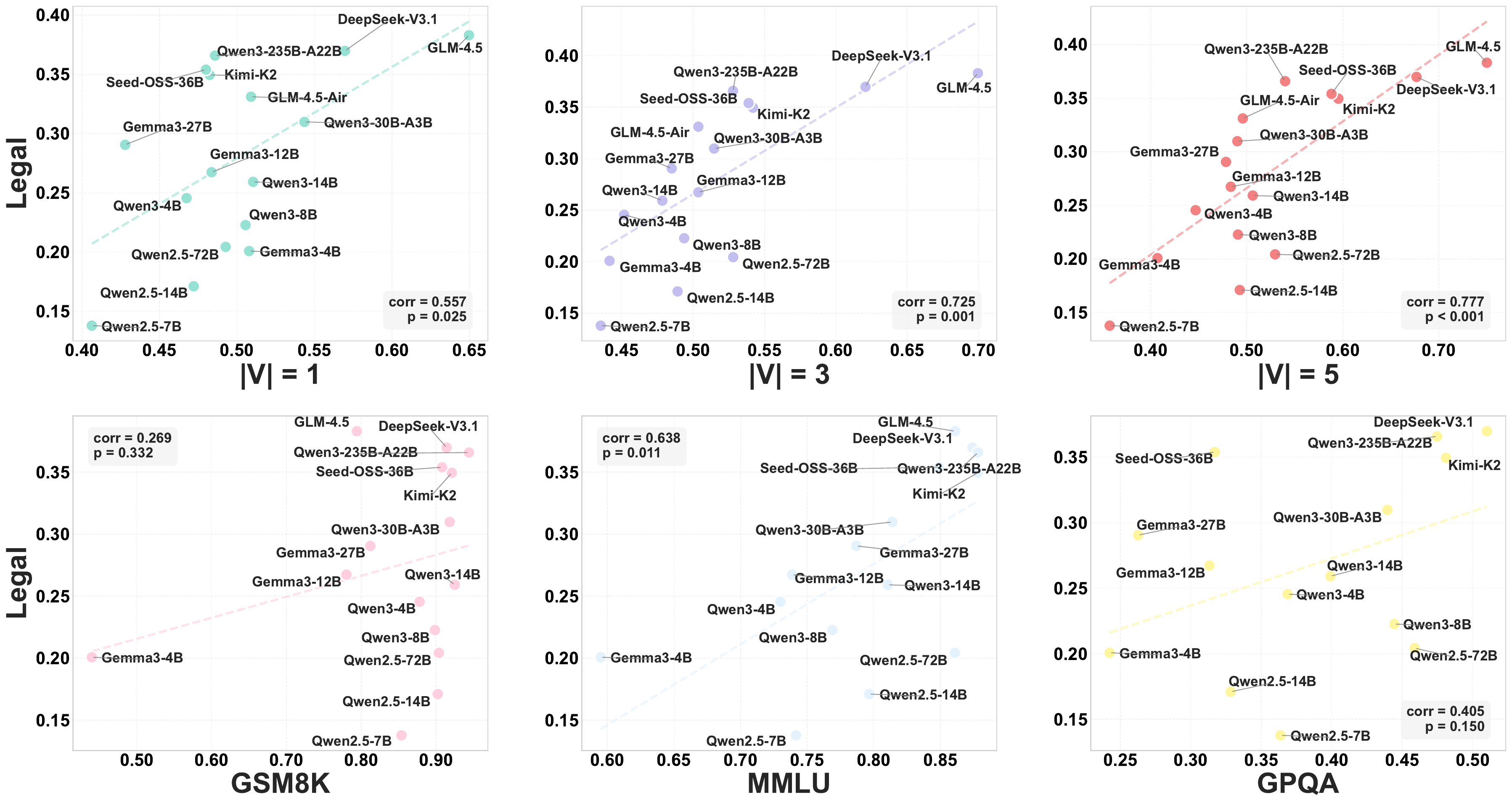}
    \caption{\textbf{Correlation Analysis: PRBench-Legal.}}
\end{figure*}

\begin{figure}[htbp]
    \centering
    \includegraphics[width=\textwidth]{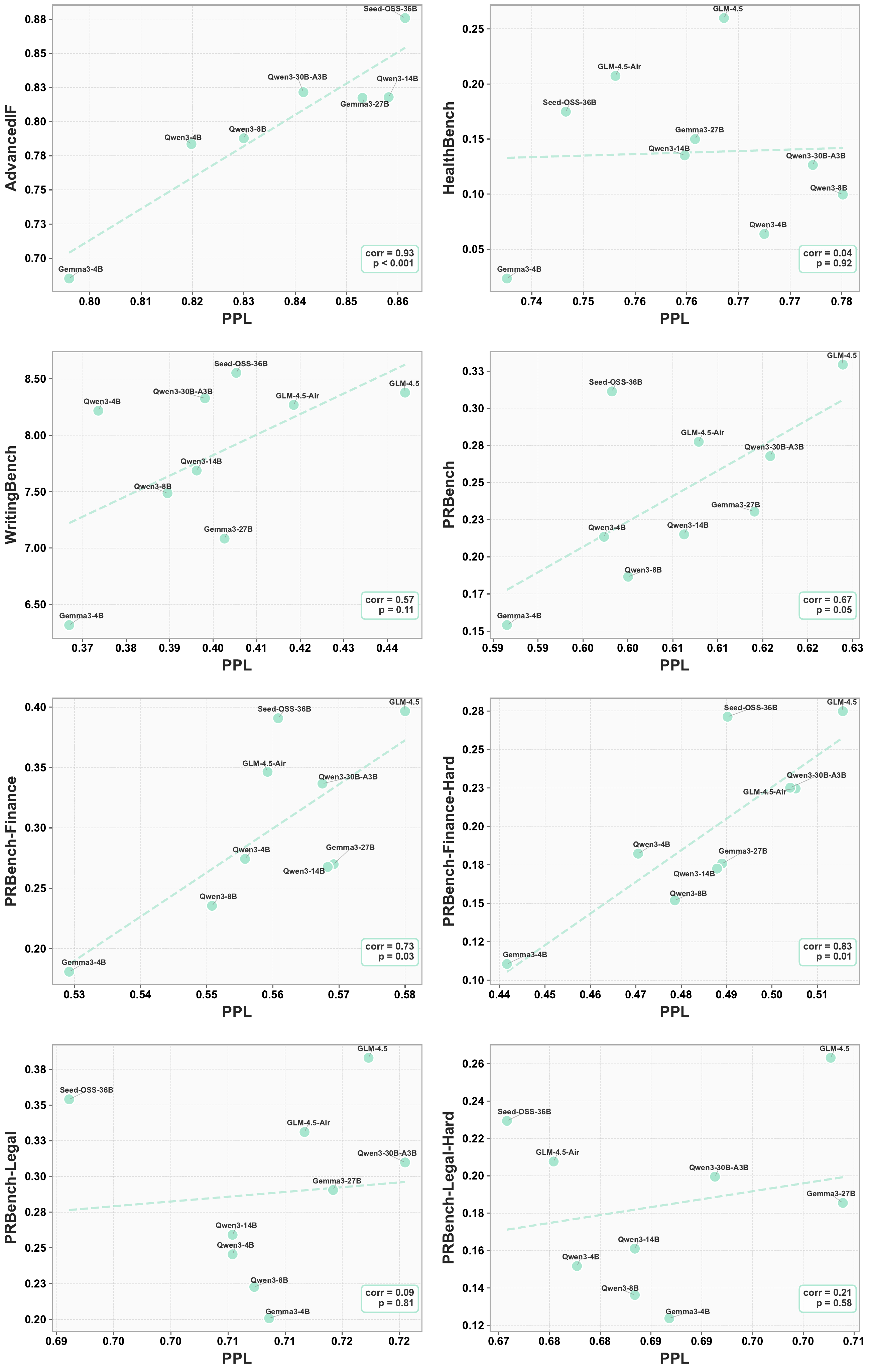}
    \caption{\textbf{Correlation between PPL and post-training generative performance across all evaluation subsets.} Each subplot shows the Pearson correlation for one dataset. PPL yields strong correlations on some domains but fails to generalize consistently.}
    \label{fig:bpbdiff_grid}
\end{figure}

\section{Compute Resources}
\label{sec:compute}
All experiments in this work were conducted on a single machine equipped with 224 CPU cores, 2,048\,GB of RAM, and 8 NVIDIA A100 (80\,GB) GPUs. The discriminative evaluation of all 16 base models across the four \texttt{RuDE} domains and all difficulty levels required approximately 100 GPU-hours in total. The controlled RL training experiments (Section~\ref{sec:rl_verification}) were performed on the same hardware. No preliminary or failed experiments consumed significant additional compute beyond what is reported.

\section{Licenses for Existing Assets}
\label{sec:licenses}
We list the licenses for all major datasets and models used in this work.

\paragraph{Datasets.}
\begin{itemize}
    \item \textbf{HealthBench}~\citep{arora2025healthbench}: MIT License. 
    \item \textbf{PRBench}~\citep{akyurek2025prbench}: CC-BY-4.0 License. 
    \item \textbf{AdvancedIF}~\citep{he2025advancedif}: CC-BY-NC-4.0 License. 
    \item \textbf{WritingBench}~\citep{wu2025writingbench}: Apache-2.0 License.
\end{itemize}

\paragraph{Models.}
\begin{itemize}
    \item \textbf{Qwen2.5 \& Qwen3 Series}~\citep{yang2025qwen3}: Apache-2.0 License.
    \item \textbf{DeepSeek-V3.1}~\citep{liu2025deepseek}: MIT License.
    \item \textbf{Gemma3 Series}~\citep{team2025gemma}: Gemma Terms of Use (permissive with redistribution requirements).
    \item \textbf{GLM-4.5 / GLM-4.5-Air}~\citep{zeng2025glm}: MIT License.
    \item \textbf{Kimi-K2}~\citep{team2025kimi}: Modified MIT License (attribution required for large-scale commercial use).
    \item \textbf{Seed-OSS-36B}~\citep{seed2025seed1}: Apache-2.0 License.
\end{itemize}

All assets are used in compliance with their respective licenses. Our use is limited to inference-based evaluation and does not involve redistribution of model weights or training data.

\section{Model Details}
\label{sec:model_details}
Table~\ref{tab:model_details} summarizes all base and instruct models evaluated in this work, including their Hugging Face identifiers, parameter counts, architectures, and availability.

\begin{table}[h]
    \centering
    \caption{\textbf{Details of all evaluated models.} We list the base and instruct model identifiers, total parameter count, architecture type, and source availability.}
    \label{tab:model_details}
    \small
    \setlength{\tabcolsep}{4pt}
    \renewcommand{\arraystretch}{1.1}
    \resizebox{\textwidth}{!}{
    \begin{tabular}{llllcl}
        \toprule
        \textbf{Short Name} & \textbf{Base Model} & \textbf{Instruct Model} & \textbf{Params} & \textbf{Arch.} & \textbf{Source} \\
        \midrule
        DeepSeek-V3.1 & deepseek-ai/DeepSeek-V3.1-Base & deepseek-ai/DeepSeek-V3.1 & 671B & MoE & Open Weight \\
        Gemma3-4B & google/gemma-3-4b-pt & google/gemma-3-4b-it & 4B & Dense & Open Weight \\
        Gemma3-12B & google/gemma-3-12b-pt & google/gemma-3-12b-it & 12B & Dense & Open Weight \\
        Gemma3-27B & google/gemma-3-27b-pt & google/gemma-3-27b-it & 27B & Dense & Open Weight \\
        GLM-4.5-Air & zai-org/GLM-4.5-Air-Base & zai-org/GLM-4.5-Air & 106B & MoE & Open Weight \\
        GLM-4.5 & zai-org/GLM-4.5-Base & zai-org/GLM-4.5 & 355B & MoE & Open Weight \\
        Kimi-K2 & moonshotai/Kimi-K2-Base & moonshotai/Kimi-K2-Instruct & 1T & MoE & Open Weight \\
        Qwen2.5-7B & Qwen/Qwen2.5-7B & Qwen/Qwen2.5-7B-Instruct & 7B & Dense & Open Weight \\
        Qwen2.5-14B & Qwen/Qwen2.5-14B & Qwen/Qwen2.5-14B-Instruct & 14B & Dense & Open Weight \\
        Qwen2.5-72B & Qwen/Qwen2.5-72B & Qwen/Qwen2.5-72B-Instruct & 72B & Dense & Open Weight \\
        Qwen3-4B & Qwen/Qwen3-4B-Base & Qwen/Qwen3-4B-Instruct-2507 & 4B & Dense & Open Weight \\
        Qwen3-8B & Qwen/Qwen3-8B-Base & Qwen/Qwen3-8B & 8B & Dense & Open Weight \\
        Qwen3-14B & Qwen/Qwen3-14B-Base & Qwen/Qwen3-14B & 14B & Dense & Open Weight \\
        Qwen3-30B-A3B & Qwen/Qwen3-30B-A3B-Base & Qwen/Qwen3-30B-A3B-Instruct-2507 & 30B & MoE & Open Weight \\
        Qwen3-235B-A22B & Qwen3-235B-A22B-Base & Qwen/Qwen3-235B-A22B-Instruct-2507 & 235B & MoE & Partial$^\dagger$ \\
        Seed-OSS-36B & ByteDance-Seed/Seed-OSS-36B-Base & ByteDance-Seed/Seed-OSS-36B-Instruct & 36B & Dense & Open Weight \\
        \bottomrule
    \end{tabular}
    }
    \vspace{2pt}
    \raggedright\scriptsize{$^\dagger$ Instruct model is open weight; base model weights are not publicly released.}
\end{table}

\section{Correlation Analysis with PPL}
\label{sec:bpbdiff}
To complement the discriminative evaluation presented in the main text, we analyze the correlation between PPL and post-training generative performance across all eight evaluation subsets. Figure~\ref{fig:bpbdiff_grid} presents the per-dataset results. While PPL shows strong correlations on certain subsets (e.g., AdvancedIF with $\text{corr}=0.93$ and PRBench-Finance-Hard with $\text{corr}=0.83$), it exhibits weak or insignificant correlations on others (e.g., HealthBench with $\text{corr}=0.04$ and PRBench-Legal with $\text{corr}=0.09$). This inconsistency across domains highlights the advantage of our discriminative scoring approach, which provides more stable and generalizable predictions of post-training potential.

\section{Robustness Analysis}
\label{sec:robustness}
To ensure that our framework reflects true post-training potential rather than superficial features, we conduct a position bias check on the AdvancedIF. By swapping the order of the positive and negative options (i.e., A/B reversal) and re-evaluating all models, we observe highly stable performance, as detailed in Table~\ref{tab:robustness_simplified}. The average deviation ($\Delta$) across all models is $+0.6\%$. This minimal fluctuation confirms that the base models' accuracy is grounded in a semantic understanding of the rubric constraints, rather than an exploitation of positional cues (e.g., a bias towards the first option)~\citep{zheng2023judging,wang2024large}.

\begin{table}[h]
    \centering
    \caption{\textbf{Robustness Analysis on AdvancedIF.} We report the aggregated accuracy (\%) across all difficulty levels ($|\mathcal{V}|$) for both the original (\textit{Forward}) and swapped (\textit{Reverse}) option orders. $\bm{\Delta}$: Forward $-$ Reverse.}
    \label{tab:robustness_simplified}
    \small
    \setlength{\tabcolsep}{8pt}
    \renewcommand{\arraystretch}{1.0}
    \begin{tabular}{l c c c}
        \toprule
        \textbf{Model} & \textbf{Forward} & \textbf{Reverse} & \bm{$\Delta$} \\
        \midrule
        Gemma3-4B       & 46.3 & 47.7 & -1.4 \\
        Gemma3-12B      & 59.0 & 60.6 & -1.6 \\
        Gemma3-27B      & 62.0 & 62.1 & -0.1 \\
        Qwen3-4B        & 55.5 & 52.3 & +3.1 \\
        Qwen3-8B        & 65.6 & 64.4 & +1.1 \\
        Qwen3-14B       & 68.5 & 59.9 & \textbf{+8.6} \\
        Qwen3-30B-A3B   & 62.5 & 63.4 & -1.0 \\
        Qwen3-235B-A22B & 74.3 & 74.3 & +0.1 \\
        Qwen2.5-7B      & 40.8 & 40.6 & +0.3 \\
        Qwen2.5-14B     & 61.9 & 61.2 & +0.7 \\
        Qwen2.5-72B     & 69.1 & 68.4 & +0.7 \\
        Seed-OSS-36B    & 77.8 & 77.2 & +0.6 \\
        GLM-4.5-Air     & 65.9 & 66.3 & -0.5 \\
        GLM-4.5         & 76.7 & 77.1 & -0.4 \\
        Kimi-K2         & 75.9 & 76.4 & -0.5 \\
        DeepSeek-V3.1   & \textbf{83.9} & \textbf{83.3} & +0.6 \\
        \midrule
        \textbf{AVG} & 64.1 & \textbf{63.4} & +0.6 \\
        \bottomrule
    \end{tabular}
\end{table}

\section{Statistical Significance: 95\% Confidence Intervals}
\label{sec:confidence_intervals}

To provide rigorous statistical support for our claims, we compute 95\% confidence intervals (CIs) via bootstrap resampling (10,000 iterations) for all reported metrics. This section presents: (1) the full \textsc{RuDE} performance table with per-domain CIs, and (2) the 4C taxonomy dimension scores with margins of error.

\subsection{RuDE Performance with 95\% Confidence Intervals}

Table~\ref{tab:rude_ci} reports model accuracy (\%) at $|\mathcal{V}|=3$ for WritingBench and $|\mathcal{V}|=5$ for all other domains, accompanied by 95\% bootstrap confidence intervals. The non-overlapping intervals between top-tier models (e.g., DeepSeek-V3.1 vs.\ Qwen3-235B-A22B) and mid-tier models confirm that the observed performance gaps are statistically significant rather than artifacts of sampling variance.

\begin{table}[h]
    \centering
    \caption{\textbf{Model performance on \textsc{RuDE} with 95\% confidence intervals.} Accuracy (\%) at $|\mathcal{V}|=3$ for WritingBench (WR) and $|\mathcal{V}|=5$ for AdvancedIF (AD), HealthBench (HE), and PRBench (PR). 95\% CI computed via bootstrap. \textbf{Bold} = best per domain.}
    \label{tab:rude_ci}
    \small
    \setlength{\tabcolsep}{3pt}
    \renewcommand{\arraystretch}{1.05}
    \resizebox{\textwidth}{!}{
    \begin{tabular}{lccccr}
        \toprule
        \textbf{Model} & \textbf{AD [95\% CI]} & \textbf{HE [95\% CI]} & \textbf{WR [95\% CI]} & \textbf{PR [95\% CI]} & \textbf{AVG} \\
        \midrule
        Gemma3-4B & 51.0 [48.5, 53.5] & 51.4 [48.4, 54.4] & 41.3 [39.5, 43.2] & 37.9 [36.7, 39.6] & 45.4 \\
        Gemma3-12B & 71.3 [69.0, 73.5] & 57.4 [54.4, 60.4] & 55.3 [53.4, 57.1] & 48.2 [46.8, 49.7] & 58.1 \\
        Gemma3-27B & 74.9 [72.6, 76.9] & 65.2 [62.2, 68.0] & 57.3 [55.4, 59.2] & 49.9 [48.2, 51.1] & 61.8 \\
        Qwen3-4B & 64.4 [62.0, 66.7] & 53.9 [50.8, 56.9] & 50.1 [47.7, 52.4] & 45.7 [43.9, 46.8] & 53.5 \\
        Qwen3-8B & 75.6 [73.4, 77.6] & 60.6 [57.6, 63.5] & 60.0 [57.6, 62.2] & 49.4 [48.0, 50.9] & 61.4 \\
        Qwen3-14B & 77.7 [75.5, 79.7] & 62.9 [59.9, 65.7] & 62.9 [60.5, 65.1] & 51.3 [49.8, 52.8] & 63.7 \\
        Qwen3-30B-A3B & 74.5 [72.2, 76.6] & 62.8 [59.8, 65.7] & 59.0 [56.6, 61.3] & 49.7 [48.2, 51.2] & 61.5 \\
        Qwen3-235B-A22B & 85.8 [84.0, 87.5] & 72.7 [69.9, 75.3] & 68.5 [66.2, 70.6] & 56.5 [54.7, 57.6] & 70.9 \\
        Qwen2.5-7B & 43.4 [40.9, 45.9] & 40.7 [37.7, 43.7] & 42.1 [40.2, 44.0] & 39.5 [37.7, 40.6] & 41.4 \\
        Qwen2.5-14B & 69.9 [67.6, 72.1] & 63.6 [60.7, 66.5] & 59.9 [58.0, 61.7] & 50.0 [48.5, 51.4] & 60.9 \\
        Qwen2.5-72B & 80.5 [78.4, 82.4] & 67.1 [64.2, 69.9] & 57.4 [55.5, 59.3] & 51.8 [50.6, 53.5] & 64.2 \\
        Seed-OSS-36B & 86.6 [84.8, 88.2] & 76.8 [74.2, 79.3] & 63.0 [61.2, 64.8] & 58.4 [57.1, 59.9] & 71.2 \\
        GLM-4.5-Air & 73.4 [71.1, 75.5] & 66.4 [63.5, 69.2] & 57.9 [56.0, 59.7] & 50.1 [48.9, 51.8] & 62.0 \\
        GLM-4.5 & 85.7 [83.9, 87.3] & 78.1 [75.5, 80.5] & 65.0 [63.2, 66.8] & 72.0 [71.4, 74.0] & 75.2 \\
        Kimi-K2 & 83.5 [81.6, 85.3] & 77.7 [75.1, 80.1] & 80.1 [78.5, 81.6] & 57.6 [56.7, 59.6] & 74.7 \\
        DeepSeek-V3.1 & \textbf{93.1} [91.7, 94.2] & \textbf{84.3} [82.0, 86.4] & \textbf{73.1} [71.4, 74.8] & \textbf{64.6} [64.1, 66.9] & \textbf{78.8} \\
        \midrule
        AVG & 74.8 & 65.2 & 59.7 & 52.2 & 63.0 \\
        \bottomrule
    \end{tabular}
    }
\end{table}

\subsection{4C Taxonomy Scores with 95\% Confidence Intervals}

Table~\ref{tab:4c_ci} presents the full 4C dimension scores (Overall, Compliance, Competence, Content, Control) with their associated 95\% margins of error ($\pm$ MoE). The tight confidence bands (typically $\pm 0.010$--$0.031$) indicate that our evaluation is highly stable, and the relative rankings among models are robust to resampling variation.

\begin{table}[h]
    \centering
    \caption{\textbf{Full 4C dimension scores (\%) with 95\% confidence intervals.} All scores are averaged across difficulty levels. 95\% CI computed via bootstrap. \textbf{Bold} = best per dimension.}
    \label{tab:4c_ci}
    \small
    \setlength{\tabcolsep}{3pt}
    \renewcommand{\arraystretch}{1.05}
    \resizebox{\textwidth}{!}{
    \begin{tabular}{lccccc}
        \toprule
        \textbf{Model} & \textbf{Overall [95\% CI]} & \textbf{Compliance [95\% CI]} & \textbf{Competence [95\% CI]} & \textbf{Content [95\% CI]} & \textbf{Control [95\% CI]} \\
        \midrule
        DeepSeek-V3.1 & \textbf{61.4} [60.4, 62.4] & \textbf{63.0} [60.7, 65.3] & 59.5 [58.1, 60.9] & \textbf{60.8} [57.7, 63.9] & \textbf{61.4} [59.4, 63.4] \\
        GLM-4.5 & 58.3 [57.2, 59.4] & 61.0 [58.2, 63.8] & 56.9 [55.4, 58.4] & 57.4 [53.5, 61.3] & 58.4 [56.3, 60.5] \\
        Kimi-K2 & 58.1 [57.1, 59.1] & 56.0 [53.7, 58.3] & 50.9 [49.5, 52.3] & 57.4 [54.3, 60.5] & 55.6 [53.6, 57.6] \\
        Seed-OSS-36B & 57.4 [56.4, 58.4] & 54.1 [51.8, 56.4] & 54.1 [52.7, 55.5] & 52.0 [48.9, 55.1] & 54.4 [52.4, 56.4] \\
        Qwen3-235B-A22B & 56.8 [55.8, 57.8] & 51.6 [49.2, 54.0] & 59.0 [57.5, 60.5] & 56.3 [53.2, 59.4] & 55.9 [53.8, 58.0] \\
        Qwen2.5-72B & 52.9 [51.9, 53.9] & 50.8 [48.4, 53.2] & 46.8 [45.4, 48.2] & 52.0 [48.8, 55.2] & 50.6 [48.5, 52.7] \\
        Qwen3-14B & 52.6 [51.6, 53.6] & 51.2 [48.8, 53.6] & 55.7 [54.2, 57.2] & 49.3 [46.2, 52.4] & 52.2 [50.1, 54.3] \\
        Qwen2.5-14B & 52.6 [51.6, 53.6] & 51.2 [48.8, 53.6] & 55.7 [54.3, 57.1] & 49.3 [46.2, 52.4] & 52.2 [50.2, 54.2] \\
        GLM-4.5-Air & 52.2 [51.2, 53.2] & 49.1 [46.8, 51.4] & 44.3 [42.9, 45.7] & 52.6 [49.5, 55.7] & 49.6 [47.6, 51.6] \\
        Gemma3-27B & 52.2 [51.2, 53.2] & 47.9 [45.5, 50.3] & 43.1 [41.7, 44.5] & 46.2 [43.1, 49.3] & 47.3 [45.3, 49.3] \\
        Gemma3-12B & 51.4 [50.4, 52.4] & 49.8 [47.4, 52.2] & 47.1 [45.7, 48.5] & 46.4 [43.3, 49.5] & 48.7 [46.7, 50.7] \\
        Qwen3-8B & 50.2 [49.2, 51.2] & 46.8 [44.4, 49.2] & 45.2 [43.7, 46.7] & 48.1 [45.0, 51.2] & 47.6 [45.5, 49.7] \\
        Qwen3-30B-A3B & 49.5 [48.5, 50.5] & 50.1 [47.7, 52.5] & 50.8 [49.3, 52.3] & 52.0 [48.9, 55.1] & 50.6 [48.5, 52.7] \\
        Gemma3-4B & 43.5 [42.5, 44.5] & 42.5 [40.2, 44.8] & 38.8 [37.4, 40.2] & 43.0 [39.9, 46.1] & 41.7 [39.7, 43.7] \\
        Qwen3-4B & 43.1 [42.1, 44.1] & 42.5 [40.1, 44.9] & 47.0 [45.5, 48.5] & 39.3 [36.2, 42.4] & 43.0 [40.9, 45.1] \\
        Qwen2.5-7B & 39.6 [38.6, 40.6] & 35.2 [32.9, 37.5] & 37.4 [36.0, 38.8] & 34.7 [31.7, 37.7] & 36.7 [34.7, 38.7] \\
        \midrule
        Overall AVG & 52.0 [51.8, 52.2] & 49.9 [49.3, 50.5] & 49.3 [49.0, 49.6] & 49.8 [49.1, 50.5] & 50.3 [49.8, 50.8] \\
        \bottomrule
    \end{tabular}
    }
\end{table}

\paragraph{Key Observations.}
\begin{itemize}
    \item The top-3 models (DeepSeek-V3.1, GLM-4.5, Kimi-K2) have non-overlapping overall CIs with mid-tier models, confirming statistically significant separation.
    \item The per-dimension margins of error are consistently narrow ($\leq 0.039$), indicating that the 4C diagnostic profiles are reliable and not dominated by noise.
    \item The overall average margin of error across all 16 models is merely $\pm 0.002$, reflecting high aggregate stability of the benchmark.
\end{itemize}

\section{Generalizability of RuDE across Heterogeneous Post-Training Pipelines}
\label{sec:generalizability}

A natural concern is whether \textsc{RuDE}'s predictive power is contingent on the downstream post-training procedure adopting a rubric-style reward signal. In this section, we clarify that \textsc{RuDE} generalizes broadly across heterogeneous post-training paradigms.

\paragraph{Main Validation (Section~\ref{sec:experiments}): Correlation with Officially Released Instruct Models.}
The primary evidence for \textsc{RuDE}'s predictive validity comes from Section~\ref{sec:experiments}, where the Y-axis represents the generative performance of each organization's \textit{officially released} instruct model (e.g., Qwen3-235B-A22B-Instruct, DeepSeek-V3.1, GLM-4.5, Kimi-K2-Instruct). Crucially, these models were post-trained using diverse, proprietary pipelines that typically involve multi-stage combinations of SFT, RLHF, DPO, and other alignment techniques—\textit{none of which are specifically designed around rubric-based reward signals}. The training recipes, data mixtures, reward model architectures, and optimization hyperparameters differ substantially across organizations.

Despite this extreme heterogeneity in post-training methodology, \textsc{RuDE} achieves strong and statistically significant correlations, e.g. AdvancedIF: $r = 0.91$, $p < 0.001$.
This demonstrates that \textsc{RuDE} scores capture a \textit{fundamental property} of the base model—its intrinsic discriminative sensitivity to fine-grained quality distinctions—rather than a narrow aptitude that only manifests under rubric-aligned training. The high correlations hold regardless of whether the downstream pipeline uses human preference data, AI feedback, rule-based rewards, or hybrid approaches.

\paragraph{Controlled RL Experiment (Section~\ref{sec:rl_verification}): Additional Causal Evidence.}
The controlled RL experiment in Section~\ref{sec:rl_verification}, which uses rubric-based rewards, provides \textit{complementary} but not exclusive validation. Its purpose is to establish a tighter causal link by controlling confounding variables (identical training data, identical reward, identical hyperparameters), isolating the base model as the sole independent variable. This controlled setting confirms that the discriminative gap measured by \textsc{RuDE} translates directly into differential learning speed and final performance under identical optimization pressure.

\paragraph{Interpretation.}
Together, these two validation paths establish \textsc{RuDE}'s generalizability from complementary angles:
\begin{enumerate}
    \item \textbf{Ecological validity} (Section~\ref{sec:experiments}): \textsc{RuDE} predicts real-world post-training outcomes across organizations with unknown, heterogeneous pipelines.
    \item \textbf{Causal validity} (Section~\ref{sec:rl_verification}): Under controlled conditions, the discriminative gap causally determines post-training trajectory.
\end{enumerate}
The convergence of both results strongly supports our core thesis: a base model's ability to \textit{discriminate} rubric-compliant from rubric-violating responses is a generalizable proxy for its post-training potential, independent of the specific alignment algorithm applied downstream.

\section{Iteration Cap and Failure Rates in the Construction Pipeline}
\label{sec:iteration_stats}

To provide full transparency on the engineering details of our data synthesis pipeline, we report the iteration statistics and failure rates for both phases of the \textit{construct-verify-refine} cycle.

\paragraph{Phase 1: Positive Sample Generation.}
We set a maximum iteration cap of 5 rounds for generating the golden reference ($y^+$). In practice, Phase 1 converges efficiently: the average number of iterations is \textbf{2.3}, with the vast majority of samples satisfying all rubric criteria within 2--3 rounds. The failure rate (samples that reach the iteration cap without passing verification) is below \textbf{3\%}. These failures are primarily attributable to queries with inherently complex constraint sets where multiple rubrics interact or conflict. All samples that fail to pass verification within the cap are \textit{discarded} and excluded from the final dataset.

\paragraph{Phase 2: Negative Sample Generation via Controlled Degradation.}
Precise degradation is inherently more challenging than generation. Phase 2 uses the same 5-iteration cap but requires more refinement: the average number of iterations is \textbf{3.1}, with a failure rate of approximately \textbf{8\%}. We identify two primary failure modes:
\begin{enumerate}
    \item \textit{Insufficient violation}: The generator fails to effectively violate the target rubric while maintaining response quality—the degradation is too subtle to be detected even by the verifier.
    \item \textit{Collateral violation}: The generator successfully violates the target rubric but inadvertently violates non-target rubrics as well, breaking the controlled single-constraint design.
\end{enumerate}
The second failure mode is more prevalent, underscoring that precisely injecting a defect while preserving all other quality dimensions is a non-trivial generation challenge. As with Phase 1, all failed samples are strictly discarded.

\paragraph{Final Data Quality Assurance.}
After this filtering process, all \textbf{28,683} samples in the final dataset have passed the verifier's rigorous dual-check: confirmation that $y^+$ satisfies all rubric criteria \textit{and} that $y^-$ violates exactly the intended target constraints. Furthermore, our human verification on 100 randomly sampled instances (stratified across domains) yields a \textbf{92\%} pipeline-human agreement rate (Section~\ref{ssec:human_verification}), providing additional confidence in the reliability of samples that pass the automated quality control.

\begin{table}[h]
    \centering
    \caption{\textbf{Construction pipeline statistics.} Iteration and failure rate summary for both phases of the data synthesis process.}
    \label{tab:pipeline_stats}
    \small
    \begin{tabular}{lccc}
        \toprule
        \textbf{Phase} & \textbf{Iteration Cap} & \textbf{Avg. Iterations} & \textbf{Failure Rate} \\
        \midrule
        Phase 1 (Positive $y^+$) & 5 & 2.3 & $<$3\% \\
        Phase 2 (Negative $y^-$) & 5 & 3.1 & $\sim$8\% \\
        \bottomrule
    \end{tabular}
\end{table}

\section{Spearman Rank Correlation Analysis}
\label{sec:spearman}

To verify that our conclusions do not rely on the linearity assumption inherent in Pearson's $r$, we supplement all correlation analyses with Spearman's rank correlation coefficient ($\rho$). Unlike Pearson's $r$, which measures linear association, Spearman's $\rho$ captures monotonic relationships by operating solely on rank orderings, making it robust to non-linear patterns and outlier influence.

\begin{table}[h]
    \centering
    \caption{\textbf{Pearson vs.\ Spearman correlation between \textsc{RuDE} scores and post-training generative performance.} Both metrics achieve statistical significance across all domains, confirming that the positive association is robust and does not depend on linearity assumptions.}
    \label{tab:spearman}
    \small
    \begin{tabular}{lcccc}
        \toprule
        \textbf{Domain} & \textbf{Pearson $r$} & \textbf{$p$-value} & \textbf{Spearman $\rho$} & \textbf{$p$-value} \\
        \midrule
        AdvancedIF & 0.91 & $<$0.001 & 0.84 & $<$0.001 \\
        HealthBench & 0.67 & $<$0.001 & 0.66 & $=$0.01 \\
        WritingBench & 0.62 & $=$0.01 & 0.63 & $=$0.01 \\
        PRBench & 0.80 & $<$0.001 & 0.78 & $<$0.001 \\
        \bottomrule
    \end{tabular}
\end{table}

\paragraph{Consistency between Pearson and Spearman.}
As shown in Table~\ref{tab:spearman}, both correlation measures achieve statistical significance across all four domains (all $p \leq 0.01$), and their magnitudes are highly consistent. The largest discrepancy occurs on AdvancedIF (Pearson $r = 0.91$ vs.\ Spearman $\rho = 0.84$), while the remaining three domains show near-identical values. This consistency indicates that: (1) the positive correlation between \textsc{RuDE} scores and post-training performance is robust and does not depend on the linearity assumption; and (2) the data does not contain severe non-linear patterns or influential outliers that would inflate Pearson's $r$ while suppressing rank-based measures.

\paragraph{Practical Implications for Model Selection.}
From a practitioner's perspective, the primary use case of \textsc{RuDE} is to \textit{rank} candidate base models and select the one with the highest predicted post-training potential. Spearman's $\rho$ directly quantifies exactly this ranking consistency. Achieving $\rho \geq 0.63$ ($p \leq 0.01$) across all domains demonstrates that model rankings derived from \textsc{RuDE} scores are highly concordant with rankings derived from actual post-training generative performance. In other words, if \textsc{RuDE} ranks Model A above Model B, there is strong statistical evidence that Model A will indeed outperform Model B after post-training—providing reliable guidance for compute-efficient model selection decisions.

\section{Use of Large Language Models}
\label{sec:llm_usage}

We disclose all uses of LLMs in this work:

\paragraph{Use in Experiments.}
As described in Section~\ref{sec:pipeline}, our data construction pipeline employs Gemini-3-Pro as the generator and GPT-4.1 as the verifier within the iterative \textit{construct-verify-refine} cycle. These LLMs are integral, non-standard components of our core methodology. The specific prompt templates used with these models are documented in Appendix~\ref{sec:prompt_templates}.

\paragraph{Use in Writing Assistance.}
LLMs were used to assist with polishing and editing the prose of this manuscript, including grammar correction and improving the clarity of exposition.
The authors have carefully reviewed all LLM-assisted outputs and take full responsibility for the correctness and integrity of all statements, claims, and scientific conclusions presented in this paper.


\end{document}